\documentclass[10pt,twocolumn,letterpaper]{article}

\usepackage{cvpr}
\usepackage{times}
\usepackage{epsfig}
\usepackage{graphicx}
\usepackage{amsmath}
\usepackage{amssymb}

\usepackage{caption}
\usepackage{subcaption}
\usepackage{booktabs}
\usepackage{multirow}
\usepackage{bm}
\usepackage[inline]{enumitem}
\usepackage{amsthm}
\usepackage{array}
\usepackage{dirtytalk}
\usepackage{appendix}
\usepackage[dvipsnames]{xcolor}

\newcommand{\myparagraph}[1]{
\vspace{0.1cm}\noindent
\textbf{#1}
\hspace{0.1cm}
}

\newcommand{\attr}[1]{\texttt{#1}}

\renewcommand{\eg}[0]{e.g. }

\newcommand{\green}[1]{{\color{ForestGreen}#1}}
\newcommand{\red}[1]{{\color{BrickRed}#1}}

\definecolor{textual}{HTML}{4BB193}
\definecolor{visual}{HTML}{D28B3E}
\definecolor{multimodal}{HTML}{688ACC}

\usepackage[pagebackref=true,breaklinks=true,letterpaper=true,bookmarks=false]{hyperref}

\cvprfinalcopy 

\def\cvprPaperID{xxxx} 

\ifcvprfinal\fi
\begin{document}

\title{Connecting Pixels to Privacy and Utility:\\ Automatic Redaction of Private Information in Images}


\author{Tribhuvanesh Orekondy \qquad   Mario Fritz \qquad Bernt Schiele\vspace{0.5cm} \\
Max Planck Institute for Informatics\\
Saarland Informatics Campus\\
Saabr\"ucken, Germany\\
{\tt\small \{orekondy,mfritz,schiele\}@mpi-inf.mpg.de}
}

\maketitle

\begin{abstract}
Images convey a broad spectrum of personal information. 
If such images are shared on social media platforms, this personal information is leaked which conflicts with the privacy of depicted persons.
Therefore, we aim for automated approaches to redact such private information and thereby protect privacy of the individual.

By conducting a user study we find that obfuscating the image regions related to the private information leads to privacy while retaining utility of the images.
Moreover, by varying the size of the regions different privacy-utility trade-offs can be achieved. 
Our findings argue for a ``redaction by segmentation'' paradigm. 

Hence, we propose the first sizable dataset of private images ``in the wild'' annotated with pixel and instance level labels across a broad range of privacy classes. 
We present the first model for automatic redaction of diverse private information. 
It is effective at achieving various privacy-utility trade-offs within 83\% of the performance of redactions based on ground-truth annotation.
\end{abstract}

\section{Introduction}

\begin{figure}
\begin{center}
   \includegraphics[width=\linewidth]{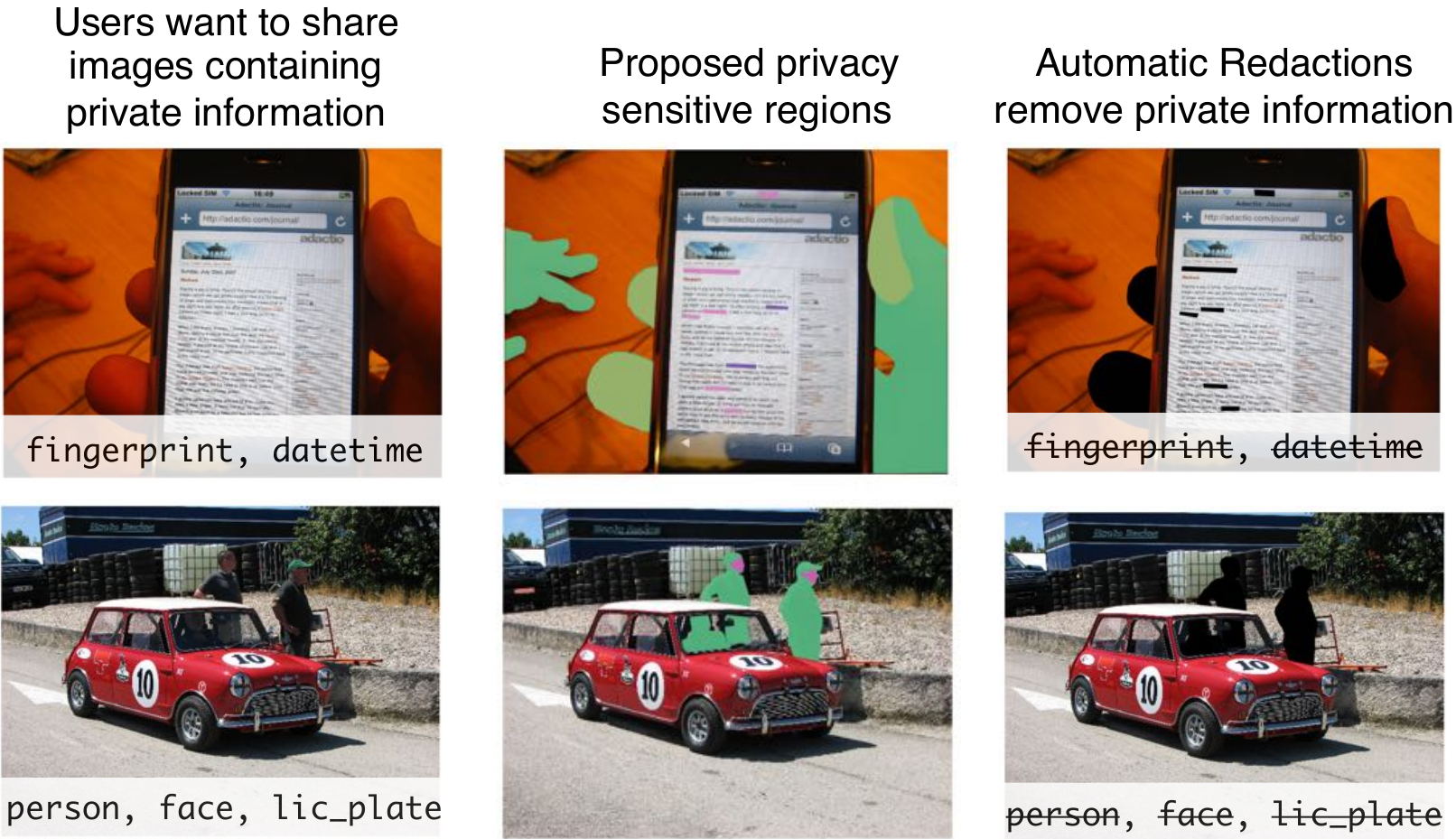}
\end{center}
  \vspace{-1em}
   \caption{
   Users often share images containing private information on the Internet, which poses a privacy risk.
   For example, in the top row, user might unintentionally leak their fingerprint.
   We present methods to aid users automatically redact such content by proposing privacy sensitive regions in images.}
\label{fig:teaser}
\end{figure}

More and more visual data is captured and shared on the Internet. Images and video contain a wide range of private information that may be shared unintentionally such as e.g. email-address, picture-id or finger-print (see \autoref{fig:teaser}). 
Consequently, there is a growing interest  within the computer vision community \cite{brkic2017know,hassan2017cartooning,joon16eccv,Oh_2017_ICCV,orekondy17iccv,raval2017protecting} to  assess the amount  of leaked information, understand implications on privacy and ultimately control and enforce privacy again. 
Yet, we are missing an understanding how image content relates to private information and how automated redaction can be approached.

Therefore, we address two important questions in this context. 
First, how can private information be redacted while maintaining an intelligible image?
We investigate this question in a user study with highly encouraging results: 
we can redact private information in images while preserving its utility.
Furthermore, varying the amount of pixels redacted results in different privacy vs.\ utility trade-offs.
We conclude that redaction by segmentation is a valid approach to perform visual redactions.

We ask a second question in this paper: What kind of privacy-utility trade-offs can be achieved by automatic redaction schemes?
Based on our first finding, we approach this as a pixel labeling task on multiple privacy classes (which we refer to as \textit{privacy attributes}).
Segmenting privacy attributes in images presents a new challenge of reasoning about regions including multiple modalities.
For instance, in \autoref{fig:teaser}, identifying the \attr{name} and \attr{datetime} requires mapping the relevant pixels to the text domain for understanding, while identifying the \attr{student\_id} requires reasoning over both visual and text domains.
Our automated methods address these challenges and localize these privacy attributes for redaction via segmentation.
By performing both quantitative and human evaluation, we find these automated methods to be effective in segmentation as well as privacy-utility metrics.

Our model and evaluation for automatic redaction is facilitated by a new dataset that extends the Visual Privacy (VISPR) dataset \cite{orekondy17iccv} to include high-quality pixel and instance-level annotations.
To this end, we propose a dataset containing 8.5k images annotated with 47.6k instances over 24 privacy attributes.
We will make the dataset publicly available for future research.

\section{Related Work}
\label{sec:related_work}
\myparagraph{Text Sanitation}
Redaction techniques are primarily studied in the context of confidential text documents, wherein certain sensitive entities need to be removed.
Studies focus on identification of such entities \cite{Chakaravarthy2008EfficientTF,Chow2008DetectingPL, Chow2009SanitizationsSS, Snchez2017TowardSD,Snchez2012DetectingSI, Snchez2013AutomaticGS} and methods to prevent over-sanitation \cite{Chakaravarthy2008EfficientTF, Snchez2017TowardSD}.
However, unlike these works which have access to dense structured text data (\eg documents), we deal with unstructured pixel-level representations of such entities.

\myparagraph{Image Perturbations for Privacy}
Adversarial perturbations \cite{43405,Metzen_2017_ICCV,Moosavi-Dezfooli_2017_CVPR} are suggested to evade person identification \cite{Oh_2017_ICCV,Sharif2016AccessorizeTA}.
However, these methods typically assume a white-box CNN-based adversary for the specific task of face recognition.
In contrast, we propose redacting content at the expense of some utility to achieve better privacy (measured against humans) across a broad range of privacy classes.
\cite{brkic2017know} proposes de-identifying people by generating an alternate appearance for the person.
We study a more fundamental problem of identifying such regions where such methods could be directly applicable.
\cite{Korshunov2012SubjectiveSO, Korshunov2013UsingWF} study obfuscation of private content, but are limited to constrained surveillance videos and non-automated methods.

\myparagraph{Private Information Recognition}
Many existing studies focus on either detecting faces \cite{Sun2017FaceDU, Viola2001RobustRF}, license plates \cite{Chang2004AutomaticLP,Zhang2006LearningBasedLP,Zhou2012PrincipalVW}, relationships \cite{Sun_2017_CVPR,Wang2010SeeingPI}, age \cite{Bauckhage2010AgeRI} or occupations \cite{shao2013you}.
Research in determining privacy risk across a broad range of privacy classes are typically treated as a classification problem \cite{orekondy17iccv,tonge2015privacy,Xioufis2016PersonalizedPI}.
However, many studies \cite{acquisti2006imagined,debatin2009facebook} demonstrate a ``privacy paradox'' -- users share such images in spite of knowing the privacy risks.
Hence in this work, we propose a middle ground for reducing privacy leakage, such that users can still share images by redacting private content while preserving its utility.

\myparagraph{Visual Privacy Datasets}
PicAlert \cite{zerr2012know} and YourAlert \cite{Xioufis2016PersonalizedPI} propose datasets with user-classified privacy labels.
VISPR \cite{orekondy17iccv} provides a more exhaustive dataset of 22k images annotated with a broad range of image-level privacy labels.
The PEViD video dataset \cite{korshunov2013pevid} provides person-centric bounding box annotation over 20 video sequences in a constrained setting.
In contrast, our dataset based on VISPR images provides pixel level annotation from a diverse set of privacy classes.

\myparagraph{Segmentation}
Identifying pixel-level labels from images is a well-studied problem in computer vision.
However, most methods \cite{li2016fully,long2015fully} and datasets \cite{Cordts2016Cityscapes,Everingham10,lin2014microsoft} focus on segmenting common objects in visual scenes.
We however focus on identifying private regions in a privacy-utility framework, which introduces many new challenges.

\section{The Visual Redactions Dataset}
\label{sec:dataset}

\begin{figure*}
  \begin{center}
  	\includegraphics[width=\textwidth]{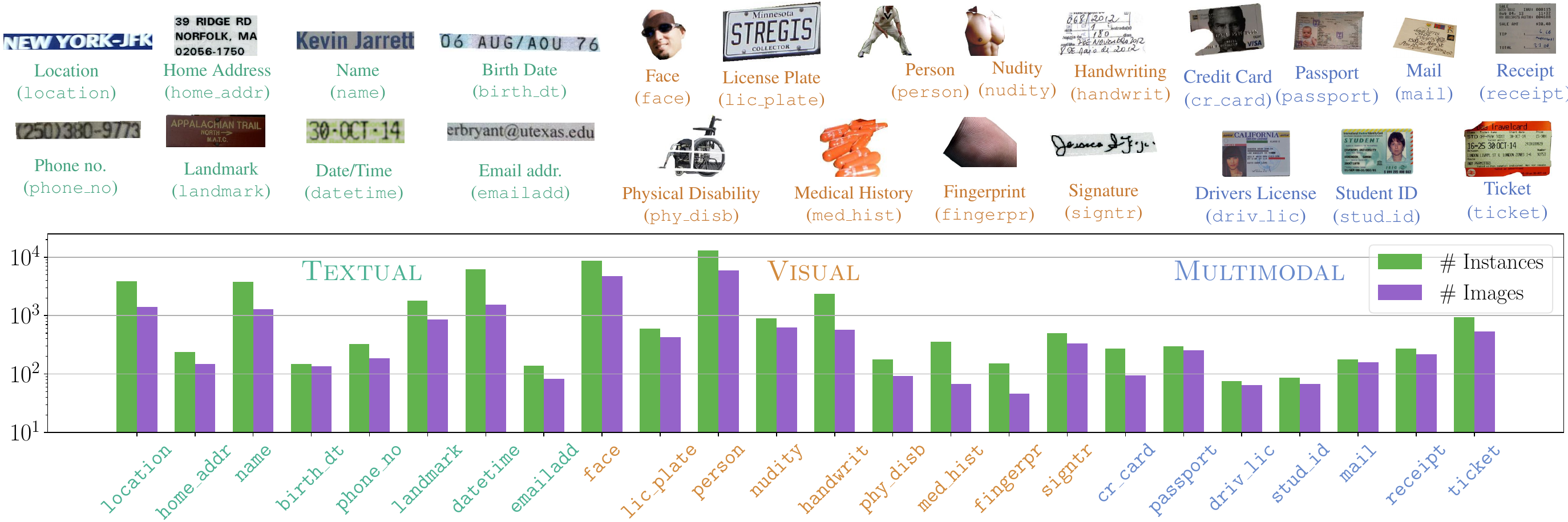}
  \end{center}
  \vspace{-1.5em}
  \caption{Examples and distribution of privacy attributes in the dataset. 
  }
  \label{fig:label_distribution}
\end{figure*}

In this section we present our pixel-label visual privacy dataset as an extension to the VISPR dataset \cite{orekondy17iccv}.
We begin with a discussion on how images (Section \ref{sec:dataset_images}) and attributes (Section \ref{sec:dataset_attributes}) were selected for the task.
This is followed by the annotation procedure (Section \ref{sec:dataset_creating}) and a brief analysis (Section \ref{sec:dataset_analysis}) of the dataset.

\subsection{Selecting Images for Pixel-level Annotation}
\label{sec:dataset_images}
The VISPR dataset contains 22k real-world user-uploaded publicly available Flickr images
which makes this a great starting point for addressing the visual redaction problem ``in the wild''.
10k of these images are annotated as safe. From the remaining 12k images we pixel-annotate the subset of 8,473 images that contain at most 5 people. The main reason to focus on this subset was to reduce the annotation cost while maximizing the amount of non-person pixels.
We preserve the identical 45-20-35 train-val-test split of these images as in the VISPR dataset.

\subsection{Shortlisting Privacy Attributes}
\label{sec:dataset_attributes}
The 22k images in the multilabel VISPR dataset are annotated using 68 image-level privacy attributes ($\sim$5.2 attributes per image).
These privacy attributes are compiled from multiple privacy-relevant sources -- the US Privacy Act of 1974,  EU Data Protection Directive 95/46/EC and various social network website rules.
Additionally, they cover a diverse range of private information that can be leaked in images (\eg face, tattoo, physical disability, personal relationships, passport, occupation).
Therefore, we use these as a starting point for redactions in images.
We select 42 out of 67 privacy attributes (excluding attribute safe) for three reasons.
First, for 11 attributes (\eg religion, occupation, sports) typically the entire image is linked to the attribute (\eg scene with church or sport stadium). 
In such cases, the solution to keeping the information private is to not share such images (as proposed in \cite{orekondy17iccv}).
We instead focus on attributes which can be localized for redaction, such that the image might still be useful.
Second, 8 attributes were extremely tedious to annotate, because of their strong co-occurrence with crowd-scenes (\eg political and general opinion, occupation) or the effort required to outline them (\eg hair color).
Third, 6 attributes (\eg place of birth, email content, national id) contained under 30 examples for training.
In spite of filtering such attributes, we still cover a broad spectrum of information to help de-identify people in images (such as by obfuscating faces or names).
We further merge few groups among these 42 attributes:
\begin{enumerate*}[label=(\roman*)]
    \item when they occur as a complete and partial version (\eg (complete face, partial face) merged into \attr{face})
    \item when they localize to the same region (\eg (race, skin color, gender, relationships) merged into \attr{person}).
\end{enumerate*}
As a result, we work with 24 localizable privacy attributes in our dataset representative of 42 of the original 67 VISPR privacy attributes (see Figure \ref{fig:label_distribution} for the complete list).

\subsection{Dataset Annotation}
\label{sec:dataset_creating}

In this section, we discuss the annotation procedure.

\myparagraph{Annotation Tool and Instructions}
We use a customized version of the VGG Image Annotator tool \cite{dutta2016via}.
Five expert annotators draw polygons around instances based on an instruction manual.
A summary of instructions, definitions of attributes and examples are provided in the supplementary material.

\myparagraph{Consensus and Agreement Measure}
Agreement is calculated \wrt images annotated by one of the authors.
We measure agreement using Mean Intersection Over Union (mIoU): $ \sum \frac{tp}{tp + fp + fn}$ averaged over images.

\myparagraph{Consensus Experiment and Annotating \attr{person}}
We observed 93.8\% agreement in consensus task of annotating instances of \attr{person} in 272 images.
Annotators separately annotated \attr{person} in remaining images.
With an annotation effort of $\sim$240 hours, we obtain 13,171 \attr{person} instances annotated over 5,920 images.

\myparagraph{Annotating \attr{face}}
We observed an agreement of 86.2\% (lower due to small sizes of instances) in the consensus task for annotating \attr{face} in 100 images.
Using the 5,920 images of people as a starting point, annotators annotated faces in separate sets of images.
In $\sim$60 hours, we gather 8,996 instances of faces.

\myparagraph{Annotating Remaining Attributes}
Images for each of the remaining 22 attributes are annotated together successively by at most a single annotator.
8 of the text-based attributes (\eg \attr{name}, \attr{phone\_no}) are annotated using 4-sided polygons or bounding boxes.
Over $\sim$220 hours, we gather annotation of 26,676 instances.

\myparagraph{Text Annotations}
We augment all images in the dataset with text detections obtained using the Google Cloud Vision API to aid localization of text-based attributes.
This is provided as OCR and bounding box annotation in structured hierarchy of text elements in the order: characters, words, paragraphs, blocks and pages.
In addition, we also gather face and landmark bounding box detections using the same API.

\myparagraph{Summary}
With an annotation effort of $\sim$800 hours concentrated over four months with five annotators (excluding the authors),  we propose the first sizable pixel-labeled privacy dataset of 8,473 images annotated with $\sim$47.6k instances using 24 privacy attributes. 

\subsection{Dataset Analysis and Challenges}
\label{sec:dataset_analysis}
We now present a brief analysis of the dataset and the new challenges it presents for segmentation tasks.
Examples of the proposed attributes and their distribution among the 8k images in the dataset are presented in Figure \ref{fig:label_distribution}.

Popular datasets \cite{Cordts2016Cityscapes,Everingham10,lin2014microsoft} provide pixel-level annotation of various common visual objects.
These objects are common in visual scenes, such as vehicles (car, bicycle), animals (dog, sheep) or household items (chair, table).
Common to all these objects are their distinctive visual cues.
Looking at the examples of attributes in Figure \ref{fig:label_distribution}, one can notice similar cues among the \textsc{Visual} attributes, but it is not evident in the others.
Recognizing \textsc{Textual} attributes (such as names or phone numbers) in images instead require detecting and parsing text information and additionally associating it with prior knowledge.
While some of the \textsc{Multimodal} attributes can be associated with visual cues, often the text content greatly helps disambiguate instances (a card-like object could be a \attr{student\_id} or \attr{driv\_lic}).
We also observe a strong correlation between modalities and sizes of instances.
We find \textsc{Textual} instances to occupy on average less than 1\% of pixels in images, while the \textsc{Multimodal} attributes predominantly occur as close-up photographs occupying 45\% of the image area on average.
Consequently, the privacy attributes pose challenges from multiple modalities and require specialized methods to individually address them.
Moreover, they provide different insights due to the variance in sizes.
Hence, going forward, we treat the modes \textsc{Textual}, \textsc{Visual} and \textsc{Multimodal} as categories to aid analysis and addressing challenges presented by them.

\myparagraph{Applicability to other problems}
We believe the proposed dataset could be beneficial to many other problems apart from visual redactions.
In visual privacy, it complements datasets to perform tasks such as person de-identification \cite{brkic2017know,hassan2017cartooning}.
Outside of the privacy domain, we also provide a sizable face segmentation dataset with 9k \attr{face} instances, compared to 2.9k in Labeled Faces in the Wild \cite{GLOC_CVPR13} and 200 in FASSEG \cite{khan2015multi}.

\section{Understanding Privacy and Utility \wrt Redacted Pixels}
\label{sec:privacy_utility_analysis}

In this section, we study how redacting ground-truth pixels of attributes influences privacy and utility of the image by conducting a user study on Amazon Mechanical Turk (AMT).
We will also use the results from this study as a reference point for evaluating our proposed automated methods in Section \ref{sec:experiments_human}.

\subsection{Generating Redactions}
Given an image $I_a$ containing attribute $a$, we generate a ground-truth redacted version of the image $I_{\bar{a}}$ by simply blacking-out pixels corresponding to $a$ in the ground-truth.

\begin{figure}
  \begin{center}
  	\includegraphics[width=\linewidth]{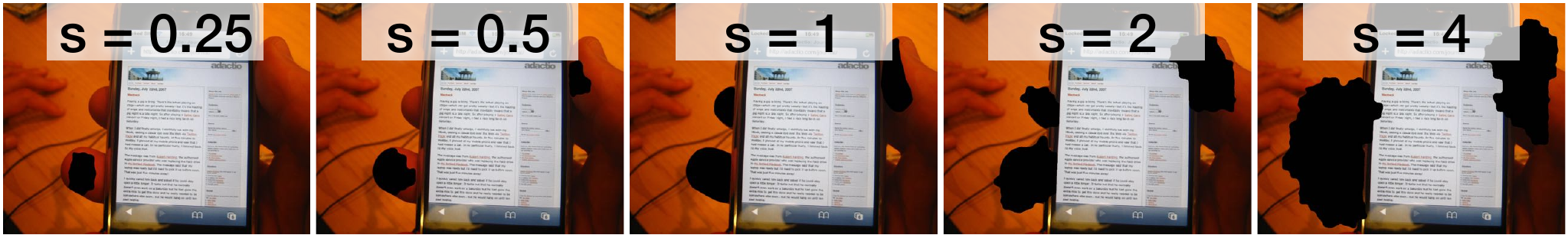}
  \end{center}
  \vspace{-1em}
  \caption{Dilation/Erosion of attribute \attr{fingerprint}}
  \label{fig:dilation}
\end{figure}

\begin{figure*}
  \begin{center}
  	\includegraphics[width=\textwidth]{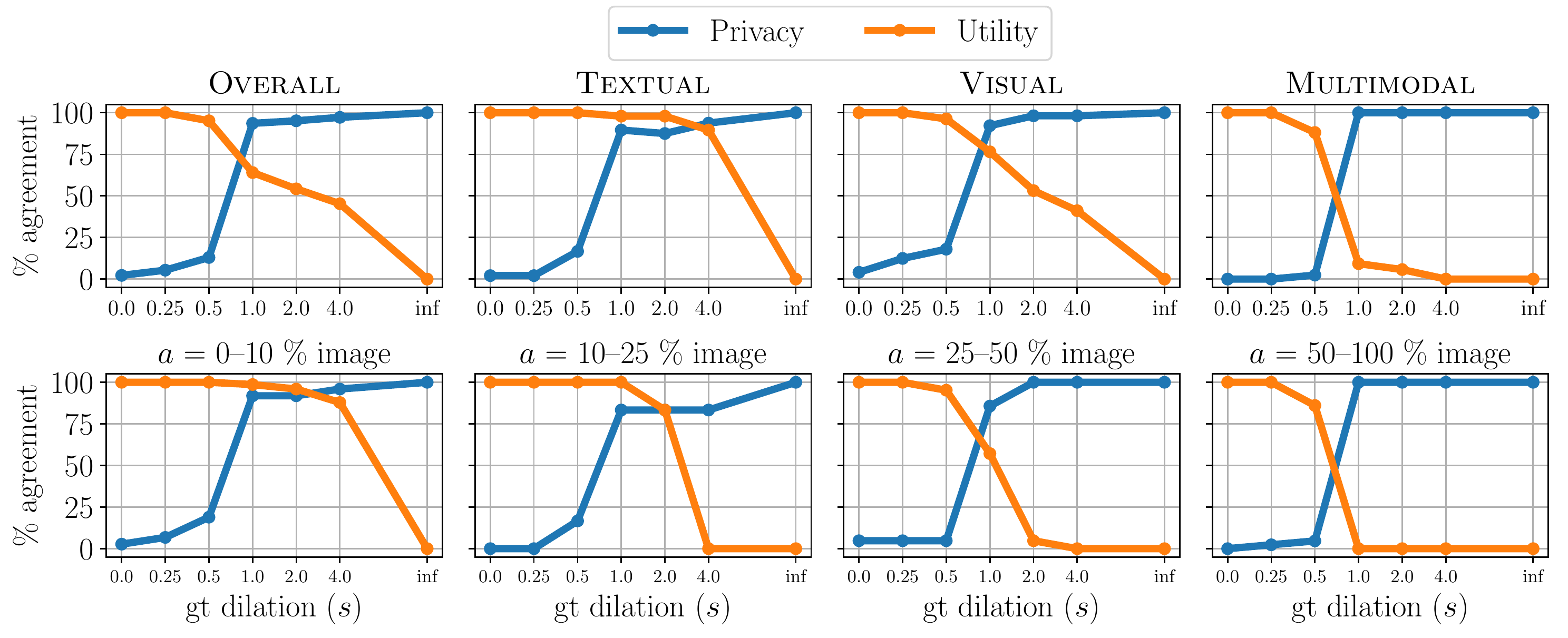}
  \end{center}
  \vspace{-2em}
  \caption{Privacy and Utility using various scales of ground-truth redaction over
  (Top row) modes
  (Bottom row) sizes}
  \label{fig:p_vs_u_modes_sizes}
\end{figure*}

\myparagraph{Spatially extending $a$}
We now want to redact fewer or more pixels in image $I_{\bar{a}}$ to understand how this influences the image's privacy and utility.
We generate multiple versions of the ground-truth redacted image $\left\{I^{s}_{\bar{a}}: s \in S\right\}$ at different scales of redaction, such that $I^{ns}_{\bar{a}}$ contains $n$ times as many blacked-out pixels of $I^{s}_{\bar{a}}$.
We achieve different scales of redactions by dilating/eroding the ground-truth binary mask of $a$, as shown in \autoref{fig:dilation}.
We use seven scales $S = \{0.0, 0.25, 0.5, 1.0, 2.0, 4.0, \inf\}$, where $I^{0}_{\bar{a}}$ is the unredacted image, $I^{1}_{\bar{a}} \left(= I_{\bar{a}}\right)$ is the GT redacted image and $I^{\inf}_{\bar{a}}$ is a completely blacked-out image.

\subsection{User Study}
\label{sec:privacy_utility_analysis_study}
We create an AMT project of 1,008 tasks (24 attributes $\times$ 6 images $\times$ 7 scales), each to be responded by 5 unique workers from a pool of 29 qualified workers.
Each task contains 2 yes/no questions based on an image $I^{s}_{\bar{a}}$, one each for Privacy and Utility.
We consider \textit{privacy} and \textit{utility} \wrt 
\begin{enumerate*}[label=(\roman*)]
	\item two versions of the same image: $(I_a, I^{s}_{\bar{a}})$, and
    \item users (AMT workers in our case).
\end{enumerate*}

\myparagraph{Defining Privacy}
To understand if attribute $a$ has been successfully redacted in $I^{s}_{\bar{a}}$, we pose the privacy question in the form: ``Is $a$ visible in the image?''.
We also provide a brief description of the attribute $a$ along with examples.
We consider $I^{s}_{\bar{a}}$ to be \textit{private}, if a majority of the users respond \textit{no}.

\myparagraph{Defining Utility}
To understand utility of an image, we pose the question: ``Is the image intelligible, so that it can be shared on social networking websites? \ie does this image convey the main content of the original image (i.e., the image without the black patch)''.
As a result, we define the utility of an image independent to its aesthetic value and instead associate it with the semantic information.
We consider $I^{s}_{\bar{a}}$ to have \textit{utility}, if a majority of the users respond \textit{yes}.

\myparagraph{Measuring Privacy and Utility}
We label each of the 1,008 images with varying redacted scales their privacy and utility as discussed above.
For any given redaction scale $s$, we aggregate privacy/utility scores simply as the percentage of images considered private/useful.
Consequently, an ideal visual redaction has both high privacy and utility.

\subsection{Analysis}
We now discuss results based on the privacy-utility scores obtained over modes and various sizes (\ie relative size of $a$ in $I_a$) based on \autoref{fig:p_vs_u_modes_sizes}.

\myparagraph{Privacy is a Step Function}
We observe in Figure \ref{fig:p_vs_u_modes_sizes} across all plots, that a minimum number of pixels of attribute $a$ need to be removed to effectively redact it from the image.
This minimum number corresponds to exactly the ground-truth redaction ($s=1$) -- redacting fewer pixels than this makes the image non-private and redacting more pixels achieves marginal privacy gains.
More specifically, we achieve 94\% privacy with ground-truth redactions.
The imperfect privacy score is predominantly (5/9 failure cases) due to turkers overlooking important details in the question.
Apart from this, other cases involve contextual cues revealing the attribute (\eg shadow of a wheelchair) and regions that were not annotated (\eg outline of a \attr{person} at a distance).

\myparagraph{Gradual Loss in Utility}
From Figure \ref{fig:p_vs_u_modes_sizes} \textsc{Overall}, we find utility to decrease gradually as the size of redacted region increases.
Another interesting observation is that utility strongly depends on the size of $a$ in the image.
In the bottom row of Figure \ref{fig:p_vs_u_modes_sizes}, we see that for smaller GT regions ($a = 0-10$\%), we still obtain high utility at larger dilations.
However, as the area of the GT regions increases beyond 50\% of the image, redaction entails blacking-out the majority of the image pixels and hence zero utility.

\myparagraph{Privacy and Utility}
What can we take away from this while proposing automated methods to preserve privacy while retaining utility?
Due to the correlation between modes and sizes, we can predict more pixels for smaller attributes with minimal loss to utility.
For instance, for \textsc{Textual} attributes, we can predict 4x as many ground-truth pixels for redaction.
However, for larger ground-truth regions ($>$50\% of image) both privacy and utility are step functions and hence making redaction a choice between privacy and utility.

\myparagraph{GT Segmentations are a Good Proxy}
In general, for images over all attributes and sizes (Figure \ref{fig:p_vs_u_modes_sizes} \textsc{Overall}), we see that we can already achieve high privacy \textit{while} retaining considerable utility of the image.
Moreover, we obtain near-perfect privacy with the highest utility in all cases at $s=1$, the ground-truth redactions.
This justifies to address privacy attribute redaction as a segmentation task. 

\section{Pixel-Labeling of Private Regions}
\label{sec:methods}
In Section \ref{sec:dataset} we discussed the challenges of attributes occurring across multiple modalities (\textsc{Textual, Visual, Multimodal}).
In Section \ref{sec:privacy_utility_analysis}, we motivated how ground-truth segmentations in our dataset make a good proxy for visual redactions.
In this section we propose automated methods to perform pixel-level labeling (semantic segmentation) of privacy attributes in images, with an emphasis on methods tackling each modality.

We begin with a simple baseline \textbf{Nearest Neighbor (NN)}: A 2048-dim feature is extracted using ResNet-50 for each image.
At test time, we predict the segmentation mask of the closest training image in terms of $L_2$ distance.

\subsection{Methods for \textsc{Textual}-centric attributes}
\label{sec:methods_textual}
To facilitate segmenting textual attributes, for each image we first obtain an ordered sequence of bounding box detections of words and their OCR using the Google Cloud Vision API (as discussed in Section \ref{sec:dataset_creating}).

\myparagraph{Proxy GT}
We represent $n$ words in an image as a sequence $[(w_i, b_i, y_i)]_{i=1}^n$, where $w_i$ is the word text, $b_i$ is the bounding box and $y_i$ is the label.
We use 9 labels (8 \textsc{Textual} attributes + safe).
We assign each $y_i$ in the sequence the ground-truth attribute that maximally overlaps with $b_i$, or a \textit{safe} label in case of zero overlap.
At test-time, we segment pixels in region $b_i$ if a non-safe label is predicted for word $w_i$.
For the test set, we refer to predictions from this proxy dataset as \textbf{PROXY} to obtain an upper-bound for our methods on these text detections.

\myparagraph{Rule-based Classification (RULES)}
We use the following rules to label words in the sequence:
\begin{enumerate*}[label=(\roman*)]
	\item \attr{name}: if it exists in a set of 241k names obtained from the US Census Bureau website.
    \item \attr{location, landmark, home\_address}: if it exists in a set of 3.7k names of cities and countries from Wikipedia's list of locations with a population of more than 110k.
    \item \attr{datetime, phone\_no, birth\_dt}: if the word contains a digit
    \item \attr{emailadd}: if the word contains the symbol @, we predict this word and adjacent words assuming a format $\Box@\Box.\Box$
\end{enumerate*}

\myparagraph{Named Entity Recognition (NER)}
We use the popular Stanford NER CRFClassifier \cite{Finkel2005IncorporatingNI} to label each word of the sequence as from a set of recognized entity classes (\eg person, organiziation, \etc).
We use the model which is trained on case-invariant text to predict one of seven entity classes.

\begin{figure}
  \begin{center}
  	\includegraphics[width=\linewidth]{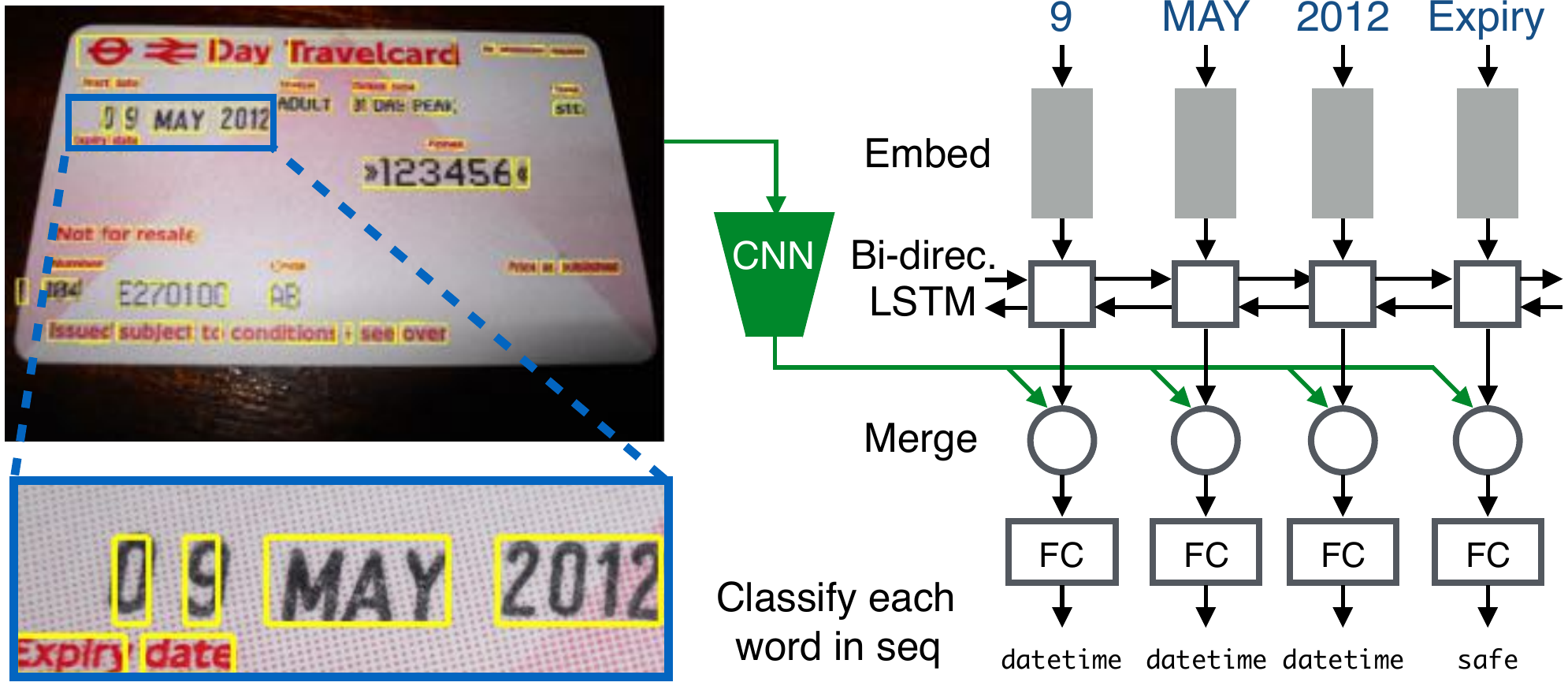}
  \end{center}
  \vspace{-1.5em}
  \caption{Architecture to perform Sequence Labeling}
  \label{fig:seqlbl}
\end{figure}

\myparagraph{Sequence Labeling (SEQ)}
We train a sequence labeler similar to \cite{huang2015bidirectional,LampleBSKD16,ma2016end} as shown in \autoref{fig:seqlbl}.
We preprocess by replacing all digits with 0s and stem each word to reduce the size of the vocabulary.
We tokenize the words in the training sequences using a vocabulary of size 4,149 (number of words with at least 4 occurrences).
We embed the words using 100-d GloVe embeddings \cite{pennington2014glove}.
To capture the temporal nature, we use two-level Bidirectional LSTMs.
At each time-step, we obtain a joint embedding by element-wise multiplication of: the \textit{text} embedding (256-d output of the LSTM) and the \textit{image} embedding (2048-d ResNet-50 \cite{He2015} feature reduced to 256-d using an FC layer).
We classify this joint embedding into 9 labels using an FC layer followed by softmax activation. 

\subsection{Methods for \textsc{Visual}-centric attributes}
\label{sec:methods_visual}
Recent deep-learning segmentation methods have proven to be effective in localizing objects based on their visual cues.
We propose using a state of the art method in addition to few pretrained methods for \textsc{Visual} attributes.

\myparagraph{Pretrained Models (PTM)}
We use pretrained methods to classify three classes typically encountered in popular visual scene datasets.
\begin{enumerate*}[label=(\roman*)]
	\item \attr{face}: We use bounding box face detections obtained using the Google Cloud Vision API.
    \item \attr{person}: We use the state-of-the-art segmentation method FCIS \cite{li2016fully} to predict pixels of COCO class ``person''
    \item \attr{lic\_plate}: We use OpenALPR \cite{openalpr} to detect license plates in images.
\end{enumerate*}

\begin{figure*}
  \begin{center}
  	\includegraphics[width=\textwidth]{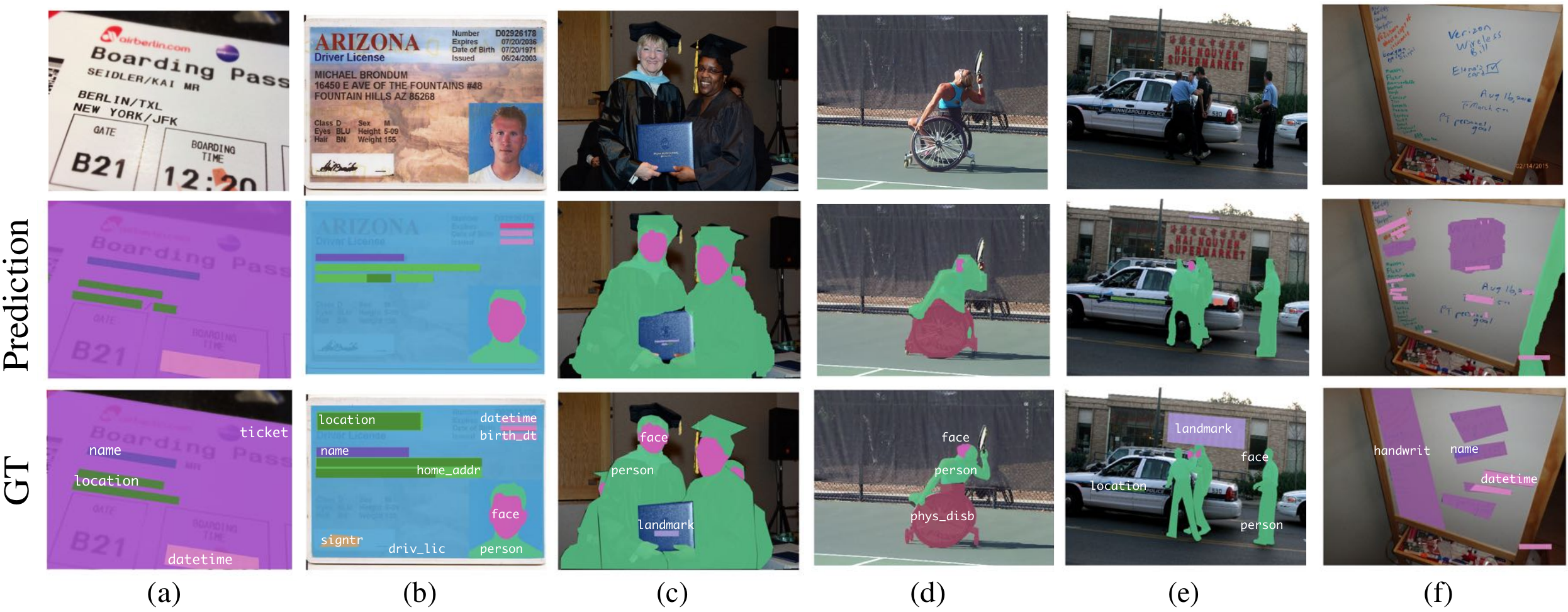}
  \end{center}
  \vspace{-1.5em}
  \caption{Qualitative examples from our method}
  \label{fig:qual}
\end{figure*}

\myparagraph{FCIS}
We retrain all layers of the FCIS model \cite{li2016fully} for our task and dataset.
We train it for 30 epochs with learning rate 0.0005 over trainval examples and their horizontally mirrored versions.
We fine tune it from the model provided by the authors trained for segmentation on MS-COCO \cite{lin2014microsoft}.
We obtained best results using default hyper-parameters.

\subsection{Methods for \textsc{Multimodal}-centric attributes}
\label{sec:methods_multimodal}
Recognizing Multimodal attributes (e.g. \attr{driv\_lic}, \attr{receipt}) require reasoning over both visual and textual domains.
We treat this as a classification problem due to:
\begin{enumerate*}[label=(\roman*)]
    \item limited training examples ($\sim$125 per multimodal attribute)
    \item large region of these attributes ($\sim$45\% image area), which provides only $\sim$10\% utility even after GT-based redaction (Section \ref{sec:privacy_utility_analysis_study}).
\end{enumerate*}

\myparagraph{Weakly Supervised Labeling (WSL)}
We propose learning a multilabel classifier based on visual-only (\textbf{WSL:I}) and visual+text content (\textbf{WSL:I+T}).
If the class probability of an attribute is beyond a certain threshold, we predict all pixels in the image for the attribute.
WSL:I is the same approach used in \cite{orekondy17iccv} -- a multilabel ResNet-50 \cite{He2015} classifier.
In the case of WSL:I+T, we obtain a multimodal embedding by concatenating visual and text representations.
We obtain visual representation (identical to WSL:I) with a ResNet-50 architecture.
We obtain text representation by encoding all words in the image.
We tried three such variants:
\begin{enumerate*}[label=(\roman*)]
	\item \textit{Bag-of-Words (BOW) encoding}: Words in the image are represented as a one-hot vector with vocabulary of size 1,751.
    \item \textit{LSTM encoding}: Identical to SEQ, we encode the word sequence using an LSTM with 128-hidden units.
    We use output from the last cell as the text representation.
    \item \textit{Conv1D encoding}: We use 1D convolutions to encode the word sequence (typically used for sentence classification tasks \cite{Kim2014ConvolutionalNN}) followed by max pooling to obtain a fixed-size text representation
\end{enumerate*}
In all three cases, we reduce the text-representation to 512-d using an FC+ReLU layer.
We report BOW encoding results  for \textbf{WSL:I+T} in the rest of the paper since this provided the best results.

\myparagraph{Salient Object Prediction (SAL)}
Using WSL:I+T as the base classifier, we use the salient object as an approximation of the attribute's location.
We obtain class-agnostic saliency obtained using DeepLab-v2 ResNet \cite{chen2016deeplab,joon17cvpr}.

\myparagraph{Weakly Supervised Iterative Refinement (IR)}
For document-like objects, the text regions tend to be densely clustered in images.
Hence, after classification using WSL:I+T, we refine the convex hull of the text regions using DenseCRF \cite{krahenbuhl2011efficient} to ``spill into'' the document region.

\section{Experiments and Discussion}
\label{sec:experiments}
In this section, we discuss segmentation performance (Section \ref{sec:experiments_segmentation}) and privacy-vs-utility performance (Section \ref{sec:experiments_human}) of our proposed methods.

\subsection{Evaluating Segmentation Performance}
\label{sec:experiments_segmentation}
We now evaluate methods proposed in Section \ref{sec:methods} in terms of its segmentation performance using Mean Average Precision, suggested in Pascal VOC \cite{Everingham10}.
This is calculated by averaging area under precision-recall curves over the privacy attributes.
We use 50 thresholds uniformly spaced between 0 and 1 to obtain this curve.
At each threshold $t$, we:
\begin{enumerate*}[label=(\roman*)]
	\item binarize the prediction score masks per image by thresholding pixel-level scores at $t$
	\item aggregate pixel-level TP, FP, FN  counts (normalized by image size) per attribute over all images to obtain attribute-level precision and recall.
\end{enumerate*}
We ignore GT masks containing under $25^2$ pixels during evaluation ($<$1\% GT masks).

\autoref{tab:quant} presents the quantitative results of the proposed methods on the test set.
Qualitative results in Figure \ref{fig:qual} are based on an \textbf{ENSEMBLE}, using predictions of SEQ for \textsc{Textual}, FCIS for \textsc{Visual}, WCS:I+T for \textsc{Multimodal} attributes.
We generally observe that NN underperforms simple baselines across all modalities, highlighting the difficulty and diversity presented by the dataset.

\myparagraph{\textsc{Textual}}
We observe:
\begin{enumerate*}[label=(\roman*)]
	\item \textit{Patterns, frequency and context}: SEQ achieves the best overall score, justifying the need for special methods to tackle text attributes.
	It is reasonably effective in detecting \attr{datetime} (Fig. \ref{fig:qual}a), \attr{emailadd} and \attr{phone\_no} due to patterns they often display.
	We additionally find SEQ detect attributes which often require prior knowledge (\eg \attr{name}, \attr{location}).
    The common success modes in such cases are when the words are popular entities (\eg ``Berlin'' in Fig. \ref{fig:qual}a) or have discriminative visual/textual context (\eg detecting \attr{home\_addr} in Fig. \ref{fig:qual}b).
    
    \item \textit{Challenges imposed by text detections}: PROXY represents an upper bound to our textual methods.
    The low scores highlights the difficulty of text detection and this is especially severe for scene and handwritten text detection, a frequent case in our dataset (\eg Fig. \ref{fig:qual}e,f).
    Moreover, our text detections do not perfectly overlap with ground-truth annotations.
    Since text regions are small, we additionally pay a high performance penalty even for correct detections (\eg IoU=0.42 for \attr{home\_addr} in Fig. \ref{fig:qual}b).
    Moreover, even in the case of correct text detections, we observe failures in OCR which affects the quality of input for dependent methods.
    This can be observed by the under-performance of NER, which is typically very effective on clean sanitized text.
\end{enumerate*}

\myparagraph{\textsc{Visual}}
We observe:
\begin{enumerate*}[label=(\roman*)]
	\item \textit{The unreasonable effectiveness of FCIS}:
    We obtain the highest score in the \textsc{Visual} category using FCIS.
    We find FCIS to be highly effective localizing visual objects commonly encountered in other datasets (\eg \attr{person}, \attr{face}).
    Moreover, we find it achieves reasonable performance even when there is a lack of training data (\eg only $<$60 examples  of \attr{fingerpr}, \attr{phys\_disb}, see Fig. \ref{fig:qual}d).
    The common failure modes are either difficult examples (\eg \attr{face} in Fig. \ref{fig:qual}e) or uncommon visual objects (\eg \attr{signtr} in Fig. \ref{fig:qual}b).
    \item \textit{Comparison with Baselines}: PTM achieves comparable results for \attr{person}, due to Flickr images used to train both models.
	However, it underperforms for \attr{face} (detections are not precise enough) and \attr{lic\_plate} (poor performance in the wild).
\end{enumerate*}

\myparagraph{\textsc{Multimodal}}
We observe:
\begin{enumerate*}[label=(\roman*)]
	\item \textit{WSL:I is a good simple baseline}:
    WSL:I achieves reasonable performance (45.4) for multimodal attributes, compared to other modes (1.5 in text and 20.8 in visual) although the prediction spans the entire image.
    This is attributed to large size of \textsc{Multimodal} instances found in images.
    \item \textit{Multimodal reasoning helps}:
    We find WSL:I+T improves performance over WCS:I by 20\%, justifying the need for methods to perform multimodal reasoning to detect these attributes.
    This is particularly necessary to disambiguate similar looking visual objects (\eg card-like objects \attr{driv\_lic} and \attr{stud\_id}, Fig. \ref{fig:qual}b).
    \item \textit{Precision-Recall trade-off}:
    We find precision for WSL:I+T for this method can be improved for some attributes (\eg \attr{cr\_card}, \attr{ticket}) by IR, which instead of the entire image, predicts only the smoothened hull of text regions.
    We observe FCIS achieve the best overall score due to higher precision.
\end{enumerate*}

\begin{table}[]
\centering
\footnotesize
	\textsc{Textual} \\ \vspace{0.1em}
  \begin{tabular}{@{}p{0.8cm}p{0.4cm}p{0.4cm}p{0.4cm}p{0.4cm}p{0.4cm}p{0.4cm}p{0.4cm}p{0.4cm}p{0.4cm}@{}}
  \toprule
Method      & mAP           & \multicolumn{1}{p{0.4cm}}{\centering \texttt{loca} \\ \vspace{-0.3em} \texttt{tion}}      
							& \multicolumn{1}{p{0.4cm}}{\centering \texttt{home} \\ \vspace{-0.3em} \texttt{addr}}    
                            & \attr{name}          
                            & \multicolumn{1}{p{0.4cm}}{\centering \texttt{birth} \\ \vspace{-0.3em} \texttt{dt}}     
                            & \multicolumn{1}{p{0.4cm}}{\centering \texttt{phone} \\ \vspace{-0.3em} \texttt{no}}     
                            & \multicolumn{1}{p{0.4cm}}{\centering \texttt{land} \\ \vspace{-0.3em} \texttt{mark}}      
                            & \multicolumn{1}{p{0.4cm}}{\centering \texttt{date} \\ \vspace{-0.3em} \texttt{time}}      
                            & \multicolumn{1}{p{0.4cm}}{\centering \texttt{email} \\ \vspace{-0.3em} \texttt{add}}      \\ \midrule
PROXY & 45.0          & 31.7          & 37.8          & 48.7          & 52.5          & 52.6          & 33.6          & 52.4          & 50.8          \\
NN          & 0.5           & 0.2           & 0.4           & 0.1           & \textit{0.6}  & 0.0           & 2.0           & 0.5           & 0.0           \\
NER         & 3.0           & \textit{6.0}  & 1.7           & 4.4           & 0.5           & 0.0           & 0.5           & 10.9          & 0.0           \\
RULES  & 4.2           & 3.1           & 0.5           & 2.8           & 0.6           & 1.4           & 1.2           & 6.4           & \textit{17.5} \\
FCIS        & \textit{7.2}  & 4.3           & 0.2           & \textit{9.8}  & 0.1           & \textit{2.5}  & \textbf{27.6} & \textit{12.9} & 0.0           \\
SEQ     & \textbf{26.8} & \textbf{18.4} & \textbf{19.4} & \textbf{19.1} & \textbf{25.1} & \textbf{45.8} & \textit{13.9} & \textbf{33.4} & \textbf{38.9} \\ \bottomrule
  \end{tabular} \\ \vspace{0.5em}
  \textsc{Visual} \\ \vspace{0.2em}
   \begin{tabular}{@{}p{0.6cm}p{0.3cm}p{0.3cm}p{0.3cm}p{0.3cm}p{0.3cm}p{0.4cm}p{0.4cm}p{0.3cm}p{0.4cm}p{0.3cm}@{}}
  \toprule
Method & mAP           & \attr{face}          
						& \multicolumn{1}{p{0.4cm}}{\centering \texttt{licp} \\ \vspace{-0.3em} \texttt{late}}    
                        & \multicolumn{1}{p{0.3cm}}{\centering \texttt{per} \\ \vspace{-0.3em} \texttt{son}}                 
                        & \multicolumn{1}{p{0.3cm}}{\centering \texttt{nud} \\ \vspace{-0.3em} \texttt{ity}}       
                        & \multicolumn{1}{p{0.4cm}}{\centering \texttt{hand} \\ \vspace{-0.3em} \texttt{writ}}      
                        & \multicolumn{1}{p{0.4cm}}{\centering \texttt{phy} \\ \vspace{-0.3em} \texttt{disb}}     
                        & \multicolumn{1}{p{0.3cm}}{\centering \texttt{med} \\ \vspace{-0.3em} \texttt{hist}}     
                        & \multicolumn{1}{p{0.4cm}}{\centering \texttt{fing} \\ \vspace{-0.3em} \texttt{erpr}}   
                        & \multicolumn{1}{p{0.3cm}}{\centering \texttt{sig} \\ \vspace{-0.3em} \texttt{ntr}}        \\ \midrule
NN     & 13.5          & 8.4           & 11.4          & 33.1                   & 6.0           & 32.1          & 11.4          & 7.4           & 11.7          & 0.1           \\
WSL:I  & \textit{20.8} & 5.0           & 4.3           & 30.3                   & \textit{16.4} & \textit{49.9} & \textit{13.7} & \textit{37.7} & \textit{28.8} & \textit{1.3}  \\
PTM    & 16.4          & \textit{47.6} & \textit{11.6} & \textbf{88.3}          & 0.0           & 0.0           & 0.0           & 0.0           & 0.0           & 0.0           \\
FCIS   & \textbf{68.3} & \textbf{83.8} & \textbf{77.9} & \textit{87.0} & \textbf{69.7} & \textbf{80.7} & \textbf{59.0} & \textbf{45.8} & \textbf{68.1} & \textbf{42.6} \\ \bottomrule
  \end{tabular} \\ \vspace{0.5em}
  \textsc{Multimodal} \\ \vspace{0.2em}
  \begin{tabular}{@{}p{0.8cm}p{0.5cm}p{0.5cm}p{0.5cm}p{0.5cm}p{0.5cm}p{0.5cm}p{0.5cm}p{0.5cm}p{0.5cm}@{}}
  \toprule
Method  & mAP           & \multicolumn{1}{p{0.4cm}}{\centering \texttt{cr} \\ \vspace{-0.3em} \texttt{card}}      
						& \multicolumn{1}{p{0.4cm}}{\centering \texttt{pass} \\ \vspace{-0.3em} \texttt{port}}      
                        & \multicolumn{1}{p{0.4cm}}{\centering \texttt{driv} \\ \vspace{-0.3em} \texttt{lic}}     
                        & \multicolumn{1}{p{0.4cm}}{\centering \texttt{stud} \\ \vspace{-0.3em} \texttt{id}}      
                        & \attr{mail}          
                        & \multicolumn{1}{p{0.4cm}}{\centering \texttt{rece} \\ \vspace{-0.3em} \texttt{ipt}}       
                        & \multicolumn{1}{p{0.4cm}}{\centering \texttt{tic} \\ \vspace{-0.3em} \texttt{ket}}        \\ \midrule
NN      & 21.5          & 9.8           & 44.2          & 17.9          & 13.0          & 15.7          & 16.8          & 33.3          \\
WSL:I+T & \textit{56.2} & 29.7          & \textit{67.4} & \textbf{82.4} & \textbf{58.4} & \textbf{43.3} & 54.5          & 57.8          \\
SAL & 36.2          & \textbf{55.9} & 37.2          & 23.8          & 30.4          & 8.1           & 42.5          & 55.1          \\
IR      & 53.6          & 41.7          & 51.2          & 67.8          & 48.1          & 36.9          & \textit{57.2} & \textit{72.5} \\ 
FCIS    & \textbf{59.2} & \textit{53.2} & \textbf{76.3} & 66.5          & \textit{50.3} & 33.1          & \textbf{59.4} & \textbf{75.4} \\ \bottomrule
  \end{tabular}
\caption{Quantitative results of our methods for segmenting privacy regions. 
	\textbf{Bold} numbers denote highest and \textit{italicized} numbers second highest scores in the columns.}
\label{tab:quant}
\vspace{-1em}
\end{table}

\subsection{Privacy vs. Utility Trade-off by Automatic Redaction}
\label{sec:experiments_human}

\begin{figure}
  \begin{center}
  	\includegraphics[width=\linewidth]{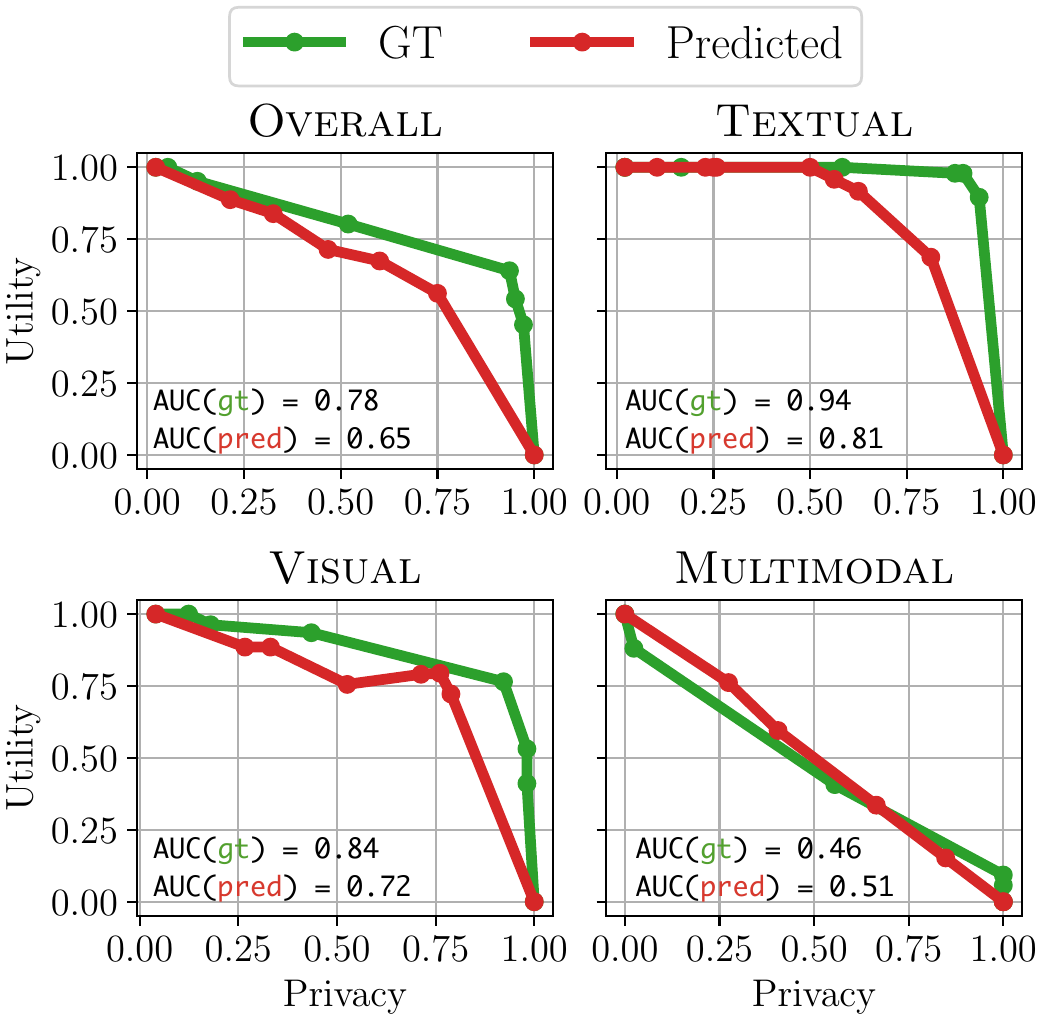}
  \end{center}
  \vspace{-1.5em}
  \caption{Comparing redactions using predicted and ground-truth segmentations}
  \label{fig:closeloop}
\vspace{-1em}
\end{figure}

In the previous section, we evaluated our approaches \wrt segmentation quality.
Now, we ask how effective are redactions based on our proposed methods in terms of privacy and utility?

To answer this, we once again run the user study in Section \ref{sec:privacy_utility_analysis_study} on AMT, but now by redacting proposed pixels of our automated method over those exact images.
To vary the number of predicted pixels, we vary the threshold to binarize the predicted score masks over attributes.
As a result, we obtain 6-8 redacted versions for each of the 144 images (24 attributes $\times$ 6 images).
Each image is labeled by 5 unique qualified AMT workers.

\myparagraph{Results}
We obtain privacy-utility scores for each threshold and plot it as a curve in Figure \ref{fig:closeloop}.
We also plot the scores obtained for different dilations of redacted ground-truth annotated region.
It should be noted that perfect redactions are unavailable to us and we use these ground-truth based redactions (or manual redactions) only to serve as a reference.
We evaluate performance by calculating area under the curve (AUC).
We observe:
\begin{enumerate*}[label=(\roman*)]
	\item Overall, we find our method obtain a privacy-utility score of 65\% -- a relative performance of 83\% compared to redactions using ground-truth annotation from the dataset.
    \item \textsc{Multimodal} attributes present a hard choice between privacy and utility, as these regions are often large.
    We find the slightly lower AUC(gt) to be an artifact of sampling.
    \item Although we obtain a low mAP for \textsc{Textual} attributes, we observe an 81\% privacy-utility score.
    This occurs as we can now over-predict regions, exhibiting low precision and high recall \wrt segmentation, but yet retaining high utility due to their small size.
    Consequently, we can predict more text pixels ``for free''.
\end{enumerate*}

Based on these observations, we find the automatic redactions of our models trained on the proposed dataset show highly promising results -- they closely mimic performance achieved by redacting ground-truth regions across a broad range of private information.

\section{Conclusion}
\label{sec:conclusion}
We proposed a redaction by segmentation approach to aid users selectively sanitize images of private content.
To learn automated approaches for this task, we proposed the first sizable visual redactions dataset containing images with pixel-level annotations of 24 privacy attributes.
By conducting a user study, we showed that redacting ground-truth regions in this dataset provides near-perfect privacy while preserving the image's utility.
We then presented automated approaches to segment privacy attributes in images and observed that we can already reasonably segment these attributes.
By performing a privacy-vs-utility evaluation of our automated approach, we achieved a highly encouraging 83\% performance \wrt GT-based redactions.

\vspace{0.5cm}
\myparagraph{Acknowledgement}
This research was partially supported by the German Research Foundation (DFG CRC 1223).
We thank Anna Khoreva and Alina Dima for feedback on the paper.

{\small
\bibliographystyle{ieee}
\bibliography{paper}

\begin{thebibliography}{10}\itemsep=-1pt

\bibitem{achanta2012slic}
R.~Achanta, A.~Shaji, K.~Smith, A.~Lucchi, P.~Fua, and S.~S{\"u}sstrunk.
\newblock Slic superpixels compared to state-of-the-art superpixel methods.
\newblock {\em TPAMI}, 2012.

\bibitem{acquisti2006imagined}
A.~Acquisti and R.~Gross.
\newblock Imagined communities: Awareness, information sharing, and privacy on
  the facebook.
\newblock In {\em PET}, 2006.

\bibitem{Bauckhage2010AgeRI}
C.~Bauckhage, A.~Jahanbekam, and C.~Thurau.
\newblock Age recognition in the wild.
\newblock In {\em ICPR}, 2010.

\bibitem{brkic2017know}
K.~Brkic, I.~Sikiric, T.~Hrkac, and Z.~Kalafatic.
\newblock I know that person: Generative full body and face de-identification
  of people in images.
\newblock In {\em CVPRW}, 2017.

\bibitem{Chakaravarthy2008EfficientTF}
V.~T. Chakaravarthy, H.~Gupta, P.~Roy, and M.~K. Mohania.
\newblock Efficient techniques for document sanitization.
\newblock In {\em CIKM}, 2008.

\bibitem{Chang2004AutomaticLP}
S.-L. Chang, L.-S. Chen, Y.-C. Chung, and S.-W. Chen.
\newblock Automatic license plate recognition.
\newblock {\em IEEE Trans. Intelligent Transportation Systems}, 2004.

\bibitem{chen2016deeplab}
L.-C. Chen, G.~Papandreou, I.~Kokkinos, K.~Murphy, and A.~L. Yuille.
\newblock Semantic image segmentation with deep convolutional nets and fully
  connected crfs.
\newblock In {\em ICLR}, 2015.

\bibitem{Chow2008DetectingPL}
R.~Chow, P.~Golle, and J.~Staddon.
\newblock Detecting privacy leaks using corpus-based association rules.
\newblock In {\em KDD}, 2008.

\bibitem{Chow2009SanitizationsSS}
R.~Chow, I.~Oberst, and J.~Staddon.
\newblock Sanitization's slippery slope: the design and study of a text
  revision assistant.
\newblock In {\em SOUPS}, 2009.

\bibitem{Cordts2016Cityscapes}
M.~Cordts, M.~Omran, S.~Ramos, T.~Rehfeld, M.~Enzweiler, R.~Benenson,
  U.~Franke, S.~Roth, and B.~Schiele.
\newblock The cityscapes dataset for semantic urban scene understanding.
\newblock In {\em CVPR}, 2016.

\bibitem{debatin2009facebook}
B.~Debatin, J.~P. Lovejoy, A.-K. Horn, and B.~N. Hughes.
\newblock Facebook and online privacy: Attitudes, behaviors, and unintended
  consequences.
\newblock {\em Journal of Computer-Mediated Communication}, 2009.

\bibitem{dutta2016via}
A.~Dutta, A.~Gupta, and A.~Zissermann.
\newblock Vgg image annotator (via), 2016.
\newblock \url{http://www.robots.ox.ac.uk/~vgg/software/via/} Accessed:
  2017-11-08.

\bibitem{Everingham10}
M.~Everingham, L.~Van~Gool, C.~K.~I. Williams, J.~Winn, and A.~Zisserman.
\newblock The pascal visual object classes (voc) challenge.
\newblock {\em IJCV}, 2010.

\bibitem{Finkel2005IncorporatingNI}
J.~R. Finkel, T.~Grenager, and C.~D. Manning.
\newblock Incorporating non-local information into information extraction
  systems by gibbs sampling.
\newblock In {\em ACL}, 2005.

\bibitem{43405}
I.~Goodfellow, J.~Shlens, and C.~Szegedy.
\newblock Explaining and harnessing adversarial examples.
\newblock In {\em ICLR}, 2015.

\bibitem{hassan2017cartooning}
E.~T. Hassan, R.~Hasan, P.~Shaffer, D.~Crandall, and A.~Kapadia.
\newblock Cartooning for enhanced privacy in lifelogging and streaming videos.
\newblock In {\em CVPRW}, 2017.

\bibitem{He2015}
K.~He, X.~Zhang, S.~Ren, and J.~Sun.
\newblock Deep residual learning for image recognition.
\newblock In {\em CVPR}, 2016.

\bibitem{Metzen_2017_ICCV}
J.~Hendrik~Metzen, M.~Chaithanya~Kumar, T.~Brox, and V.~Fischer.
\newblock Universal adversarial perturbations against semantic image
  segmentation.
\newblock In {\em ICCV}, 2017.

\bibitem{huang2015bidirectional}
Z.~Huang, W.~Xu, and K.~Yu.
\newblock Bidirectional lstm-crf models for sequence tagging.
\newblock {\em arXiv preprint arXiv:1508.01991}, 2015.

\bibitem{joon16eccv}
S.~Joon~Oh, R.~Benenson, M.~Fritz, and B.~Schiele.
\newblock Faceless person recognition; privacy implications in social media.
\newblock In {\em ECCV}, 2016.

\bibitem{joon17cvpr}
S.~Joon~Oh, R.~Benenson, A.~Khoreva, Z.~Akata, M.~Fritz, and B.~Schiele.
\newblock Exploiting saliency for object segmentation from image level labels.
\newblock In {\em CVPR}, 2017.

\bibitem{Oh_2017_ICCV}
S.~Joon~Oh, M.~Fritz, and B.~Schiele.
\newblock Adversarial image perturbation for privacy protection -- a game
  theory perspective.
\newblock In {\em ICCV}, 2017.

\bibitem{GLOC_CVPR13}
A.~Kae, K.~Sohn, H.~Lee, and E.~Learned-Miller.
\newblock Augmenting {CRF}s with {B}oltzmann machine shape priors for image
  labeling.
\newblock In {\em CVPR}, 2013.

\bibitem{khan2015multi}
K.~Khan, M.~Mauro, and R.~Leonardi.
\newblock Multi-class semantic segmentation of faces.
\newblock In {\em ICIP}, 2015.

\bibitem{Kim2014ConvolutionalNN}
Y.~Kim.
\newblock Convolutional neural networks for sentence classification.
\newblock In {\em EMNLP}, 2014.

\bibitem{koestinger2011annotated}
M.~Koestinger, P.~Wohlhart, P.~M. Roth, and H.~Bischof.
\newblock Annotated facial landmarks in the wild: A large-scale, real-world
  database for facial landmark localization.
\newblock In {\em ICCVW}, 2011.

\bibitem{Korshunov2012SubjectiveSO}
P.~Korshunov, C.~Araimo, F.~D. Simone, C.~Velardo, J.-L. Dugelay, and
  T.~Ebrahimi.
\newblock Subjective study of privacy filters in video surveillance.
\newblock {\em MMSP}, 2012.

\bibitem{korshunov2013pevid}
P.~Korshunov and T.~Ebrahimi.
\newblock Pevid: privacy evaluation video dataset.
\newblock In {\em SPIE}, 2013.

\bibitem{Korshunov2013UsingWF}
P.~Korshunov and T.~Ebrahimi.
\newblock Using warping for privacy protection in video surveillance.
\newblock {\em DSP}, 2013.

\bibitem{krahenbuhl2011efficient}
P.~Kr{\"a}henb{\"u}hl and V.~Koltun.
\newblock Efficient inference in fully connected crfs with gaussian edge
  potentials.
\newblock In {\em NIPS}, 2011.

\bibitem{LampleBSKD16}
G.~Lample, M.~Ballesteros, S.~Subramanian, K.~Kawakami, and C.~Dyer.
\newblock Neural architectures for named entity recognition.
\newblock In {\em NAACL}, 2016.

\bibitem{li2016fully}
Y.~Li, H.~Qi, J.~Dai, X.~Ji, and Y.~Wei.
\newblock Fully convolutional instance-aware semantic segmentation.
\newblock In {\em CVPR}, 2017.

\bibitem{lin2014microsoft}
T.-Y. Lin, M.~Maire, S.~Belongie, J.~Hays, P.~Perona, D.~Ramanan,
  P.~Doll{\'a}r, and C.~L. Zitnick.
\newblock Microsoft coco: Common objects in context.
\newblock In {\em ECCV}, 2014.

\bibitem{long2015fully}
J.~Long, E.~Shelhamer, and T.~Darrell.
\newblock Fully convolutional networks for semantic segmentation.
\newblock In {\em CVPR}, 2015.

\bibitem{ma2016end}
X.~Ma and E.~Hovy.
\newblock End-to-end sequence labeling via bi-directional lstm-cnns-crf.
\newblock In {\em ACL}, 2016.

\bibitem{Moosavi-Dezfooli_2017_CVPR}
S.-M. Moosavi-Dezfooli, A.~Fawzi, O.~Fawzi, and P.~Frossard.
\newblock Universal adversarial perturbations.
\newblock In {\em CVPR}, 2017.

\bibitem{openalpr}
Openalpr.
\newblock \url{https://github.com/openalpr/openalpr} Accessed: 2017-11-08.

\bibitem{orekondy17iccv}
T.~Orekondy, B.~Schiele, and M.~Fritz.
\newblock Towards a visual privacy advisor: Understanding and predicting
  privacy risks in images.
\newblock In {\em ICCV}, 2017.

\bibitem{pennington2014glove}
J.~Pennington, R.~Socher, and C.~D. Manning.
\newblock Glove: Global vectors for word representation.
\newblock In {\em EMNLP}, 2014.

\bibitem{raval2017protecting}
N.~Raval, A.~Machanavajjhala, and L.~P. Cox.
\newblock Protecting visual secrets using adversarial nets.
\newblock In {\em CVPRW}, 2017.

\bibitem{Snchez2017TowardSD}
D.~S{\'a}nchez and M.~Batet.
\newblock Toward sensitive document release with privacy guarantees.
\newblock {\em Engineering Applications of AI}, 2017.

\bibitem{Snchez2012DetectingSI}
D.~S{\'a}nchez, M.~Batet, and A.~Viejo.
\newblock Detecting sensitive information from textual documents: An
  information-theoretic approach.
\newblock In {\em MDAI}, 2012.

\bibitem{Snchez2013AutomaticGS}
D.~S{\'a}nchez, M.~Batet, and A.~Viejo.
\newblock Automatic general-purpose sanitization of textual documents.
\newblock {\em IEEE Transactions on Information Forensics and Security}, 2013.

\bibitem{shao2013you}
M.~Shao, L.~Li, and Y.~Fu.
\newblock What do you do? occupation recognition in a photo via social context.
\newblock In {\em ICCV}, 2013.

\bibitem{Sharif2016AccessorizeTA}
M.~Sharif, S.~Bhagavatula, L.~Bauer, and M.~K. Reiter.
\newblock Accessorize to a crime: Real and stealthy attacks on state-of-the-art
  face recognition.
\newblock In {\em ACM CCS}, 2016.

\bibitem{Sun_2017_CVPR}
Q.~Sun, B.~Schiele, and M.~Fritz.
\newblock A domain based approach to social relation recognition.
\newblock In {\em CVPR}, 2017.

\bibitem{Sun2017FaceDU}
X.~Sun, P.~Wu, and S.~C.~H. Hoi.
\newblock Face detection using deep learning: An improved faster rcnn approach.
\newblock {\em CoRR}, 2017.

\bibitem{tonge2015privacy}
A.~Tonge and C.~Caragea.
\newblock Privacy prediction of images shared on social media sites using deep
  features.
\newblock {\em arXiv preprint arXiv:1510.08583}, 2015.

\bibitem{Viola2001RobustRF}
P.~A. Viola and M.~J. Jones.
\newblock Robust real-time face detection.
\newblock {\em IJCV}, 2001.

\bibitem{Wang2010SeeingPI}
G.~Wang, A.~C. Gallagher, J.~Luo, and D.~A. Forsyth.
\newblock Seeing people in social context: Recognizing people and social
  relationships.
\newblock In {\em ECCV}, 2010.

\bibitem{Xioufis2016PersonalizedPI}
E.~S. Xioufis, S.~Papadopoulos, A.~Popescu, and Y.~Kompatsiaris.
\newblock Personalized privacy-aware image classification.
\newblock In {\em ICMR}, 2016.

\bibitem{zerr2012know}
S.~Zerr, S.~Siersdorfer, J.~Hare, and E.~Demidova.
\newblock I know what you did last summer!: Privacy-aware image classification
  and search.
\newblock In {\em ACM SIGIR}, 2012.

\bibitem{Zhang2006LearningBasedLP}
H.~Zhang, W.~Jia, X.~He, and Q.~Wu.
\newblock Learning-based license plate detection using global and local
  features.
\newblock In {\em International Conference on Pattern Recognition (ICPR)},
  2006.

\bibitem{Zhou2012PrincipalVW}
W.~Zhou, H.~Li, Y.~Lu, and Q.~Tian.
\newblock Principal visual word discovery for automatic license plate
  detection.
\newblock {\em IEEE Transactions on Image Processing}, 2012.

\end{thebibliography}
}


\cleardoublepage
\appendix
\appendixpage

\section{Contents}
The appendix contains:
\begin{itemize}[noitemsep,topsep=3pt,parsep=3pt,partopsep=3pt]
	\item Detailed descriptions, examples and auxiliary analysis of the 24 privacy attributes discussed in Section 3.2
    \item Precision-Recall curves for the methods discussed in Table 1
    \item Qualitative results to supplement Figure 6
    \item Implementation details and qualitative results to supplement Section 4 and Section 6.2
\end{itemize}

\section{Privacy Attributes}
In this section, we provide detailed descriptions and examples of the 24 Privacy Attributes used in the proposed dataset.
We also present a brief supplementary analysis of the conditional co-occurrence of these attributes in the dataset.

\myparagraph{Detailed Descriptions and Instructions}
In Figures \ref{fig:definitions_t}-\ref{fig:definitions_m}, we provide detailed descriptions and examples of the 24 privacy attributes grouped by category, which was discussed in Section 3.2.
The descriptions briefly summarize the instructions provided to the annotators.
The figures displays instance-agnostic ground-truth annotations of respective attributes.
Ground-truth annotations are stored in a format similar to MS-COCO \cite{lin2014microsoft}.

\textsc{Textual}, \attr{signtr} and \attr{handwrit} attributes are annotated using 4-sided polygons or bounding-boxes.
For \textsc{Textual} attributes, only Latin-based words understandable by English-speakers are annotated.
For remaining attributes, the objects are enclosed in a polygon.
In case of severe occlusion, the object is enclosed using multiple polygons.

\begin{figure}
  \begin{center}
  	\includegraphics[width=0.9\linewidth]{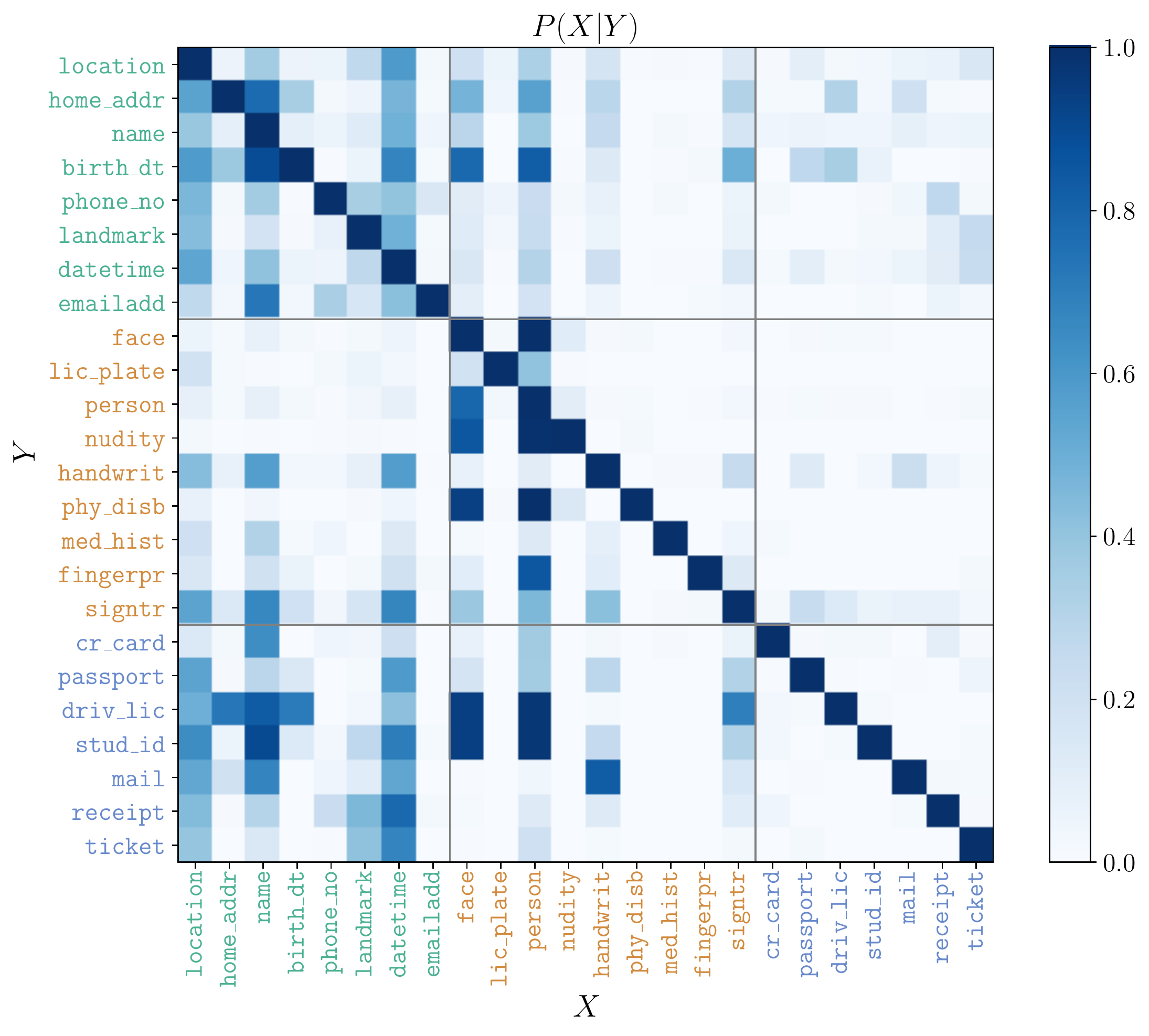}
  \end{center}
  \vspace{-1.75em}
  \caption{Conditional co-occurrence matrix. Groups of attributes are sorted by categories. Color codes used for attribute categories: {\color{textual} \textsc{Textual}}, {\color{visual} \textsc{Visual}}, {\color{multimodal} \textsc{Multimodal}}}
  \label{fig:co_occur}
\end{figure}

\myparagraph{Auxiliary Privacy Attribute Analysis}
Figure \ref{fig:co_occur} represents the conditional co-occurrence matrix (\ie probability that attribute $X$ occurs in an image containing attribute $Y$) of the 24 privacy attributes in images.
The privacy attributes along rows and columns are sorted by category.
From this plot, we find:
\begin{enumerate*}[label=(\roman*)]
    \item Images of \textsc{Multimodal} attributes often appear alongside a variety of \textsc{Textual} attributes (bottom-left block of matrix).
    \item However, the contrary is not true -- \textsc{Textual} attributes do not frequently occur only in the presence of \textsc{Multimodal} attributes (top-right block of matrix).
    \item \attr{person} and \attr{face} occur frequently alongside other \textsc{Visual} attributes as they are central to many common visual scenes (central block of matrix).
\end{enumerate*}

\begin{figure*}
  \begin{center}
  	\includegraphics[width=\textwidth]{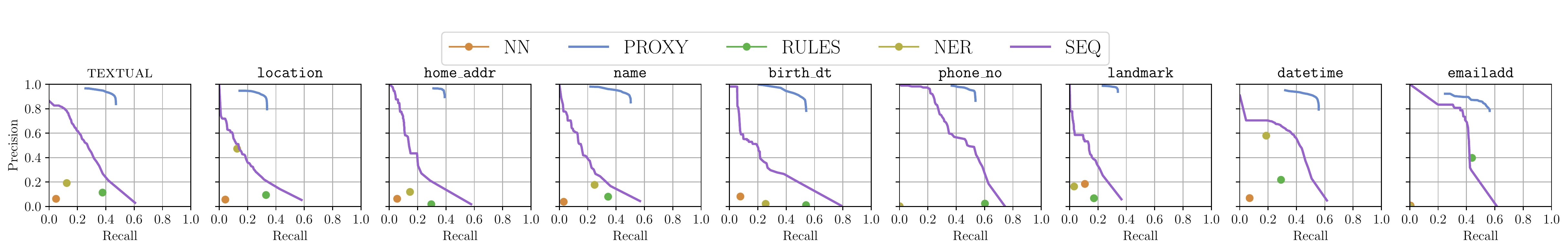}
    \includegraphics[width=\textwidth]{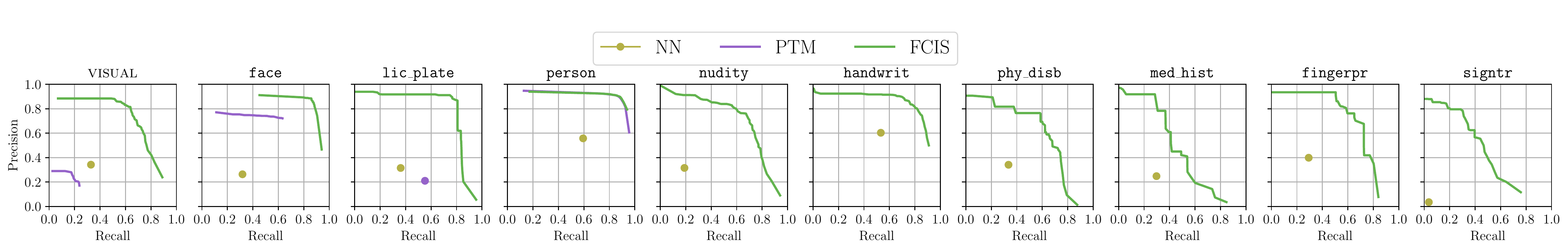}
    \includegraphics[width=\textwidth]{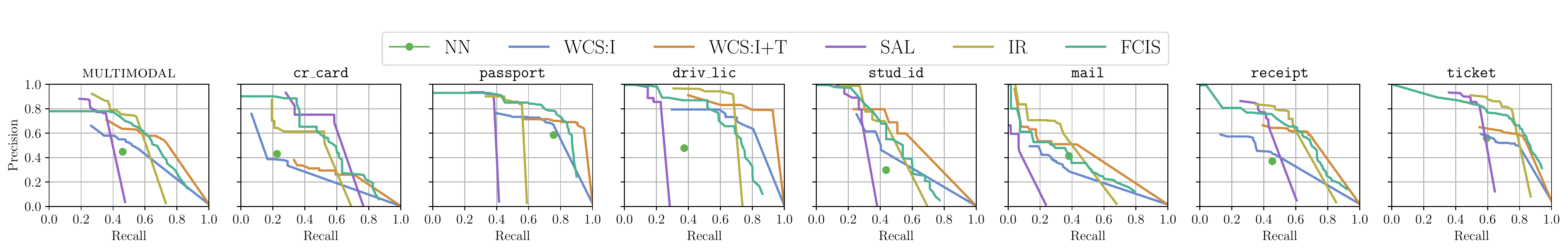}
  \end{center}
  \caption{Precision-Recall curves for methods in Table 1}
  \label{fig:pr_curves}
\end{figure*}

\section{Precision Recall Curves}
The Precision-Recall curves of methods proposed in Table 1 are presented in Figure \ref{fig:pr_curves}.
The first column represents averaged category performance.
We plot these curves by thresholding our methods at 50 uniform intervals in the range $[0, 1]$.
Similar to Pascal VOC \cite{Everingham10}, we correct the curves to have monotonically decreasing precision by setting precision at $r$ to be the highest precision at $r' \ge r$.
Moreover, precision at $r=0$ is extrapolated as highest precision at $r' \ge 0$.
We calculate Average Precision as area under this curve using trapezoidal rule.

\myparagraph{Auxiliary Discussion}
From PR curves in Figure \ref{fig:pr_curves}, we observe:
\begin{enumerate*}[label=(\roman*)]
    \item The under-performance NN indicates diversity and difficulty of the dataset.
    \item \textsc{Textual}: We find the best performance using SEQ. PROXY denotes a rough upper bound. We find SEQ obtain slightly higher recall as it predicts overlapping masks.
    \item \textsc{Visual}: We find FCIS achieve the best performance. For \attr{person}, we find a similar curve with PTM since both have the same architecture and images from the same domain (Flickr) used for training.
    \item \textsc{Multimodal}: FCIS achieves slightly higher category performance compared to others. WCS:I+T generally achieves better recall across all attributes. IR/SAL improves precision of WCS:I+T by trading off recall.
\end{enumerate*}

\section{Qualitative Results for Segmentation}
We present qualitative results in Figure \ref{fig:qual1} to supplement results in Figure 6 and discussion in Section 6.1.
We present the qualitative results per attribute, sorted by their Intersection Over Union (IoU) Scores.
Hence, figures on top represent common success modes and figures at the bottom represent common failure modes.
These results were obtained using ENSEMBLE by choosing the operating point with the highest IoU score per mode.

\section{Privacy vs. Utility Trade-off}
In this section, we provide implementation details on the redaction scaling strategy used for ground-truth redactions (Section 4.1) and predicted redactions (Section 6.1).
In both cases, we perform a black-out of relevant pixels.
For \attr{phy\_disb}, we black-out \wrt a bounding-box region since we observed the silhouette is a strong visual indicator of the attribute.
In addition, we provide qualitative results for these strategies in Figures \ref{fig:redactions_success} and \ref{fig:redactions_failure} to supplement Figure 3.

\myparagraph{Scaling Ground-truth Redactions}
We scale ground-truth redactions using super-pixels to roughly adhere to edges and object boundaries.
The downscaled image is first represented using 3000-5000 superpixels generated using SLIC0 \cite{achanta2012slic}.
We represent the ground-truth binary mask per attribute using a 0-1 labeling over the graph of super-pixels, where 1 represents the node (superpixel) belongs in the redaction.
To \textit{dilate}, we iteratively add 0-nodes with most number of adjacent 1-nodes.
To \textit{erode}, we perform the same operation with an inverted ground-truth binary-mask.
We parameterize the scaling using $s \in S$ (where $S=\{0.0, 0.25, 0.5, 1.0, 2.0, 4.0, \inf\}$), representing the dilation/erosion factor of the ground-truth mask.

\myparagraph{Scaling Predicted Redactions}
From the ENSEMBLE method, we obtain softmax probability score masks $\mathbb{R}^{w \times h \times k}$ for $k$ attributes per image.
We compute multiple thresholds per attribute to binarize the score masks, such that at threshold $t \in T$, $t$ times the number of ground-truth attribute pixels are redacted over the entire test-set of images.
We use $T = \{0.25, 0.5, 1.0, 2.0, 4.0, 8.0\}$.
For \textsc{Textual} attributes, we use an additional threshold such that all detected text is redacted.

\myparagraph{Qualitative Results Auxiliary Discussion}
Figure \ref{fig:redactions_success} and \ref{fig:redactions_failure}  displays examples of common success and failure modes \wrt to the attribute mentioned.
All images in these figures are from the test set.
$P$ and $U$ indicate privacy and utility score, which is simply the percentage of $\sim$5 AMT workers who agree to the privacy and utility questions.
High $P$ indicates the image is private \wrt to attribute $a$ and high $U$ indicates the image is intelligible.
In these figures, we find:
\begin{enumerate*}[label=(\roman*)]
	\item For small private regions, we can redact more pixels without affecting utility (Figure \ref{fig:redactions_success} \attr{location} and \attr{face})
    \item \textsc{Multimodal} attributes often display a hard choice between privacy and utility (Figure \ref{fig:redactions_success} \attr{mail})
    \item Text detections or OCR is a common failure mode with handwritten text for automatic redactions (Figure \ref{fig:redactions_failure} \attr{home\_addr})
    \item Some difficult \textsc{Multimodal} attributes (Figure \ref{fig:redactions_failure} \attr{stud\_id}) can be detected only at high thresholds, entailing complete redactions of many FP images too
    \item Figure \ref{fig:redactions_failure} \attr{fingerpr} represents one of the failure cases for ground-truth redaction discussed in Section 4.3, where AMT turkers overlook details in the question. In this particular case, the workers were asked to only consider fingerprints from fingertips. However, even at $s=1$ where the finger-tips are redacted, many workers incorrectly answer fingerprints as being visible.
\end{enumerate*}

\begin{figure*}[tbp]
  \centering
  \begin{tabular}{@{}m{2cm}m{3.0cm}m{11cm}@{}}
  \toprule
  Attribute & Example & Description  \\ \midrule
  Location (\attr{location})  & \includegraphics[width=0.17\textwidth]{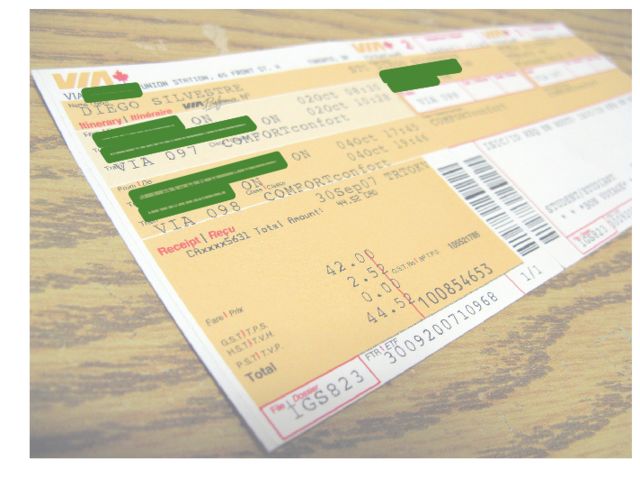}        &    Region of the image depicting where the photographer might have visited. Includes the following cases: Street signs, addresses, GPS co-ordinates, flags.          \\ \midrule
  Home Address (\attr{home\_addr})  & \includegraphics[width=0.17\textwidth]{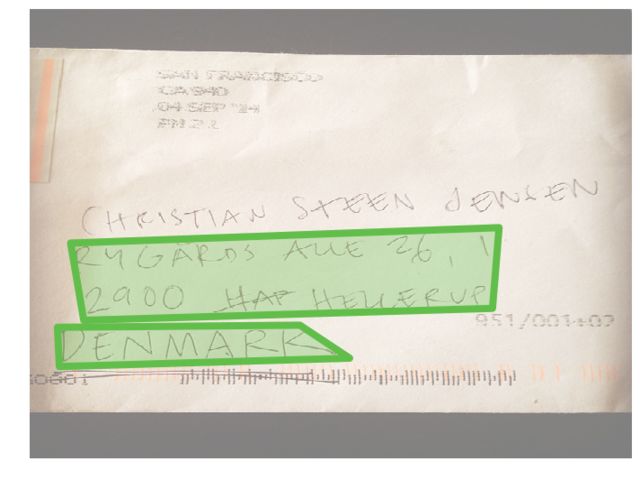}        &    Someone's home address based on the context, such as on an identity card or mail.          \\ \midrule
  Name (\attr{name})  & \includegraphics[width=0.17\textwidth]{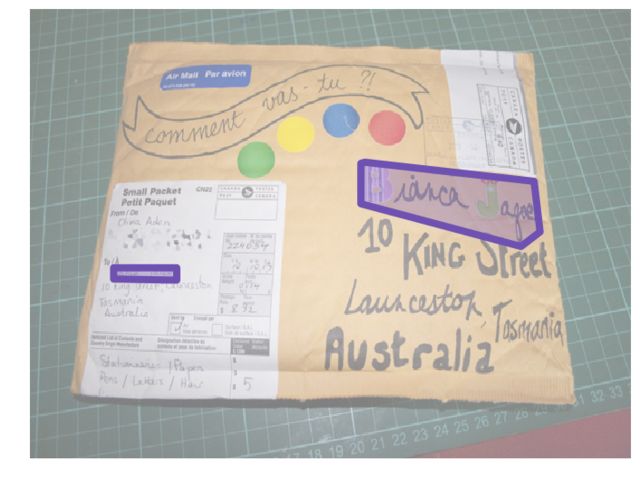}        &    Someone's name such as on a name-tag or identity card. Any recognizable name in Latin-based text is included, including that of popular figures.          \\ \midrule
  Birth Date (\attr{birth\_dt})  & \includegraphics[width=0.17\textwidth]{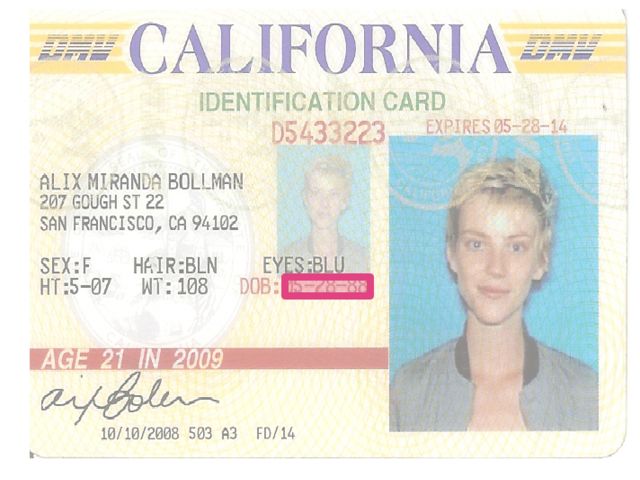}        &    Someone's date of birth (day, month and/or year) determined based on context, such as on identity cards or passports.          \\ \midrule
  Phone no. (\attr{phone\_no})  & \includegraphics[width=0.17\textwidth]{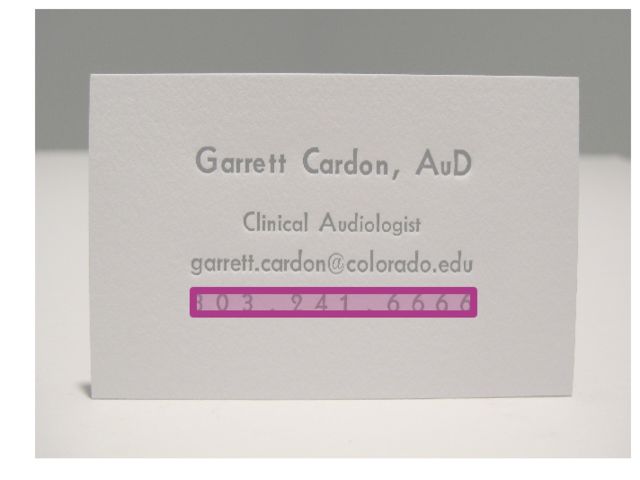}        &    A syntactically-correct phone number (either personal or business), determined either based on context or pattern.          \\ \midrule
  Landmark (\attr{landmark})  & \includegraphics[width=0.17\textwidth]{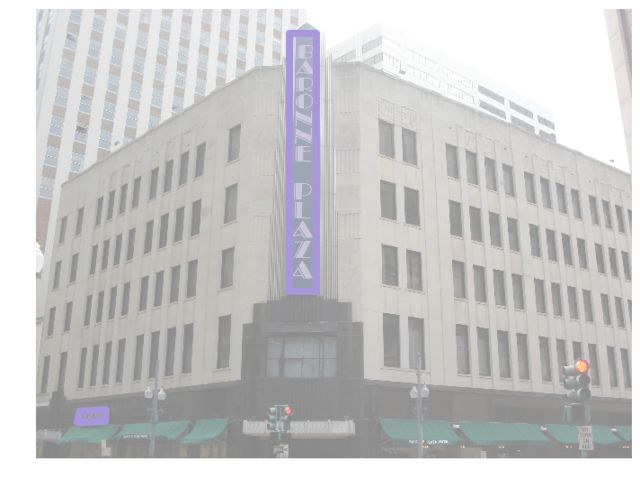}        &    Name of a store, restaurant or a business such as on a store front or a receipt.          \\ \midrule
  Date/Time (\attr{datetime})  & \includegraphics[width=0.17\textwidth]{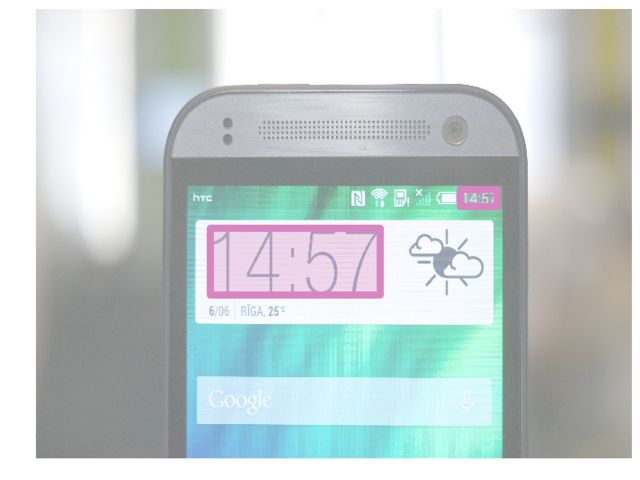}        &    A date or time, such as revealing a time-frame when the photograph might have been captured.          \\ \midrule
  Email address (\attr{emailadd})  & \includegraphics[width=0.17\textwidth]{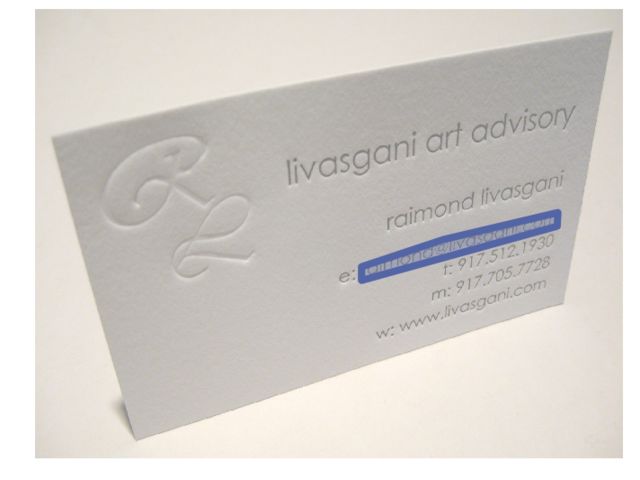}        &    A syntactically-correct email address          \\
            \bottomrule
  \end{tabular}
  \caption{Descriptions and examples of \textsc{Textual} attributes. privacy attributes. For readability, we display images where attributes are salient. 
  }
  \label{fig:definitions_t}
\end{figure*}

\begin{figure*}[tbp]
  \centering
  \begin{tabular}{@{}m{2cm}m{3.0cm}m{11cm}@{}}
  \toprule
  Attribute & Example & Description  \\ \midrule
  Face (\attr{face})  & \includegraphics[width=0.15\textwidth]{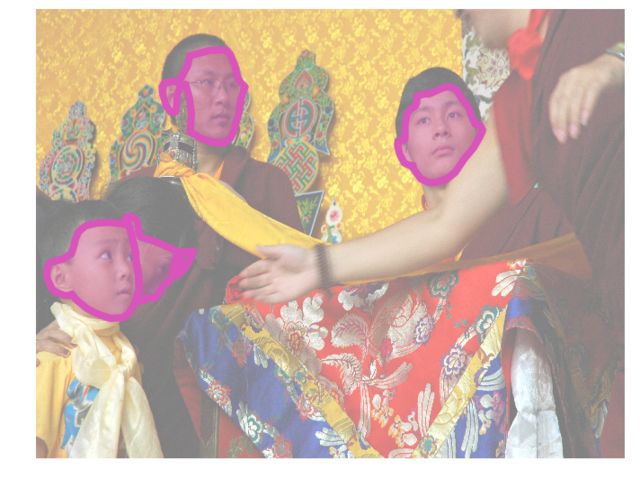}        &    Region indicating a person's face, containing all visible facial landmarks discussed in \cite{koestinger2011annotated}. Regions occluded by hair or masks are excluded.          \\ \midrule
  License Plate (\attr{lic\_plate})  & \includegraphics[width=0.15\textwidth]{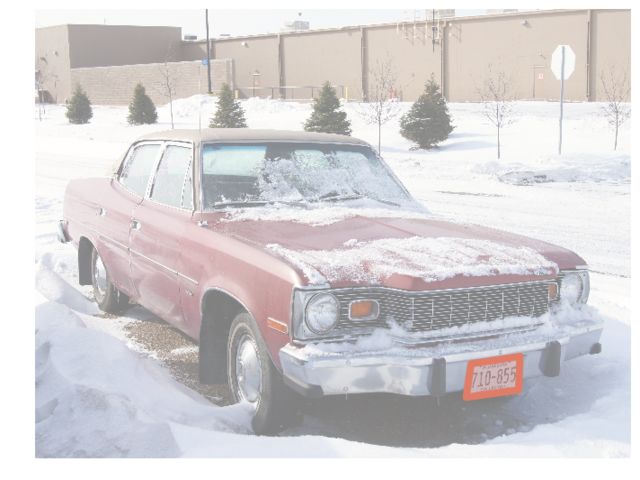}        &    Region containing a license plate or vehicle registration or identification number in any language/country. We consider any motorized vehicle (\eg cars, motorbike, train).          \\ \midrule
  Person (\attr{person})  & \includegraphics[width=0.15\textwidth]{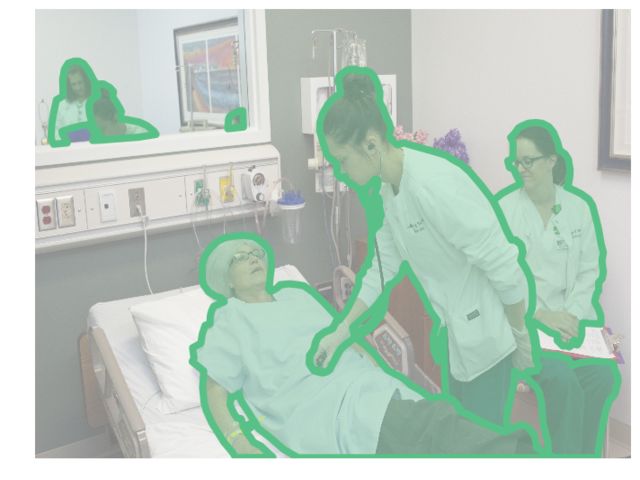}        &    Region indicating any part of a person or their reflections. Includes person's body along with wearables (\eg hats, goggles, backpacks). Excludes objects the person is holding (\eg shopping bag, guitar).          \\ \midrule
  Nudity (\attr{nudity})  & \includegraphics[width=0.15\textwidth]{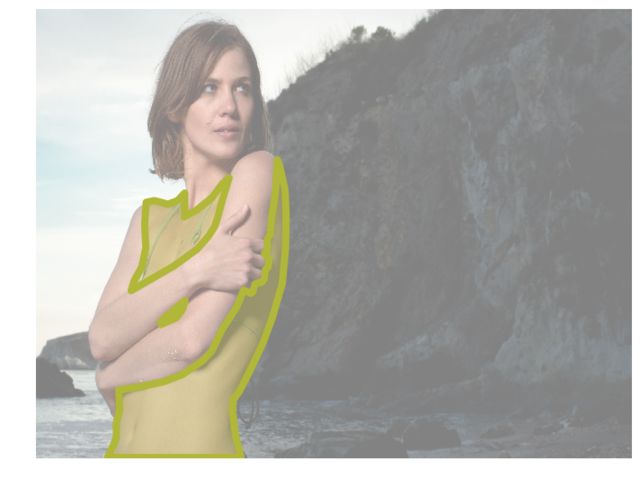}        &    Torso and thigh region of a person, if skin is completely/partially visible in this region.          \\ \midrule
  Handwriting (\attr{handwrit})  & \includegraphics[width=0.15\textwidth]{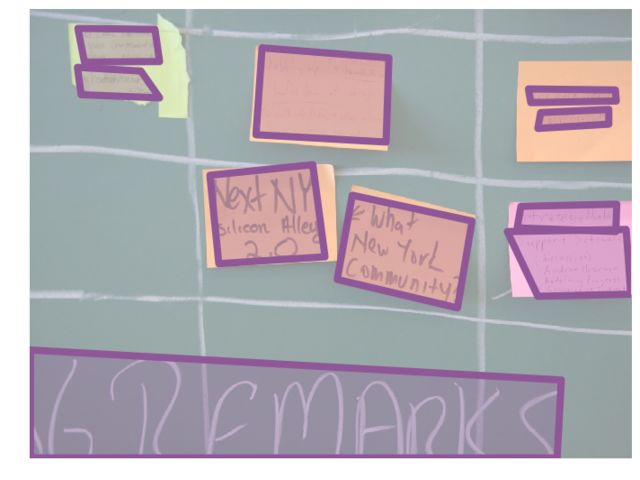}        &    Someone's handwritten text in any language.          \\ \midrule
  Physical Disability (\attr{phy\_disb})  & \includegraphics[width=0.15\textwidth]{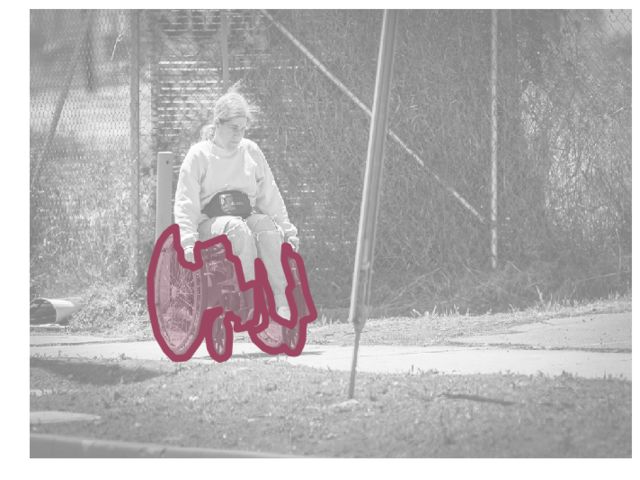}        &    Region indicating either a) special equipment used by a physically disabled person (\eg wheelchair) or b) region around limbs, if limbs are absent.           \\ \midrule
  Medical History (\attr{med\_hist})  & \includegraphics[width=0.15\textwidth]{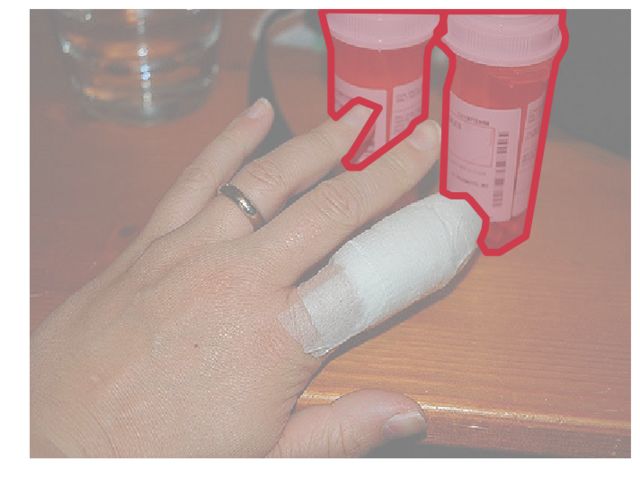}        &    Any pharmaceutical consumable such as pills, capsules or syrups (including their containers and packaging).        \\ \midrule
  Fingerprint (\attr{fingerpr})  & \includegraphics[width=0.15\textwidth]{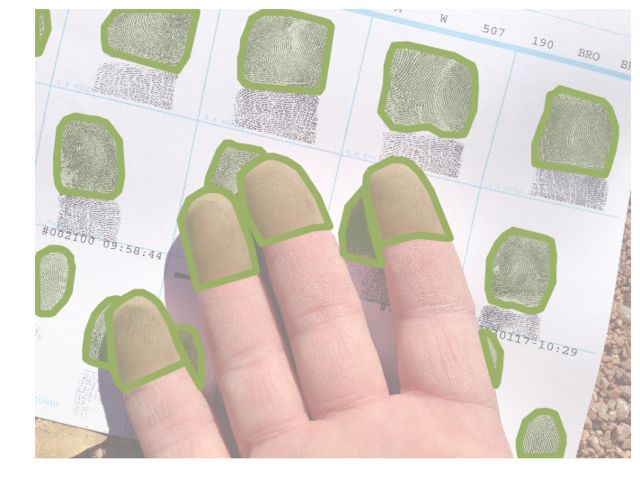}        &    Someone's finger-tips if ridges are clearly visible upon zooming-in or finger-print impressions on any surface.          \\ \midrule
  Signature (\attr{signtr})  & \includegraphics[width=0.15\textwidth]{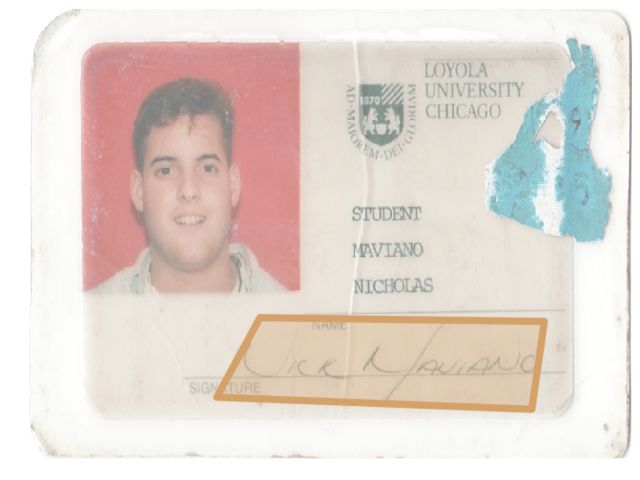}        &    Region indicating someone's signature          \\
  \bottomrule
  \end{tabular}
  \caption{Descriptions and examples of \textsc{Visual} attributes. For readability, we display images where attributes are salient.}
  \label{fig:definitions_v}
\end{figure*}

\begin{figure*}[tbp]
  \centering
  \begin{tabular}{@{}m{2cm}m{3.0cm}m{11cm}@{}}
  \toprule
  Attribute & Example & Description  \\ \midrule
  Credit Card (\attr{cr\_card})  & \includegraphics[width=0.17\textwidth]{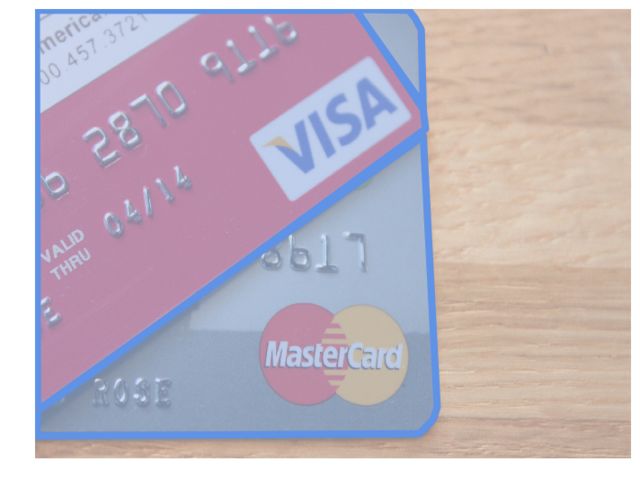}        &    Either front, rear or any details of a credit card or similar monetary instrument          \\ \midrule
  Passport (\attr{passport})  & \includegraphics[width=0.17\textwidth]{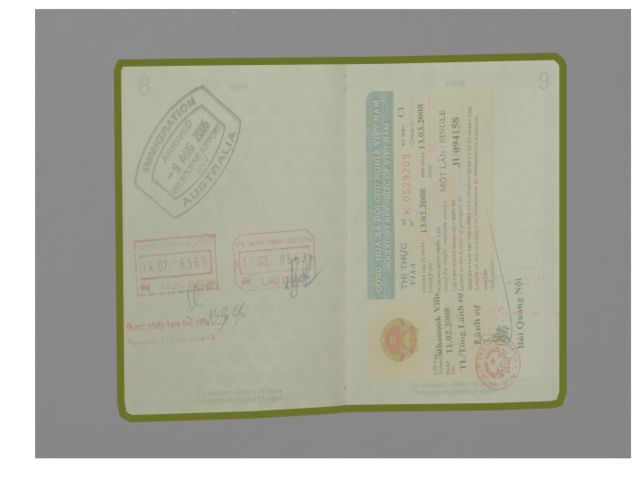}        &    Any page (including cover) of a Passport          \\ \midrule
  Drivers License (\attr{driv\_lic})  & \includegraphics[width=0.17\textwidth]{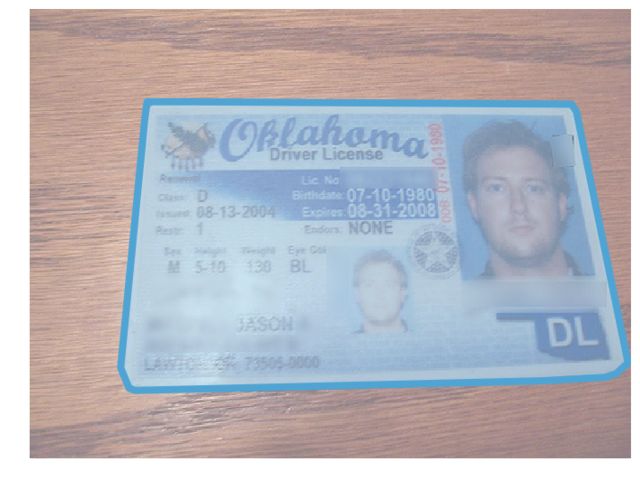}        &    Front, rear or written details of a Drivers License or driving permit          \\ \midrule
  Student ID (\attr{stud\_id})  & \includegraphics[width=0.17\textwidth]{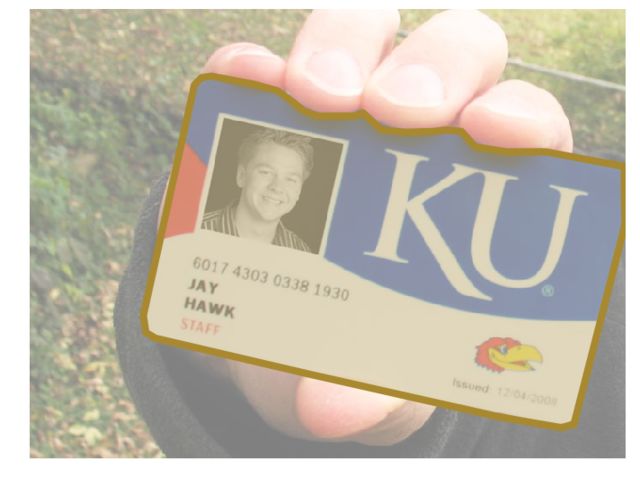}        &    Front or rear of a student identity card          \\ \midrule
  Mail (\attr{mail})  & \includegraphics[width=0.17\textwidth]{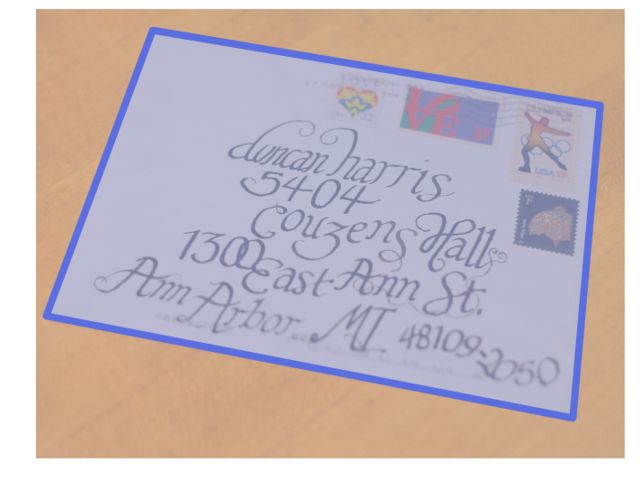}        &    Mail including hand-written letters, post-cards or packages          \\ \midrule
  Receipt (\attr{receipt})  & \includegraphics[width=0.17\textwidth]{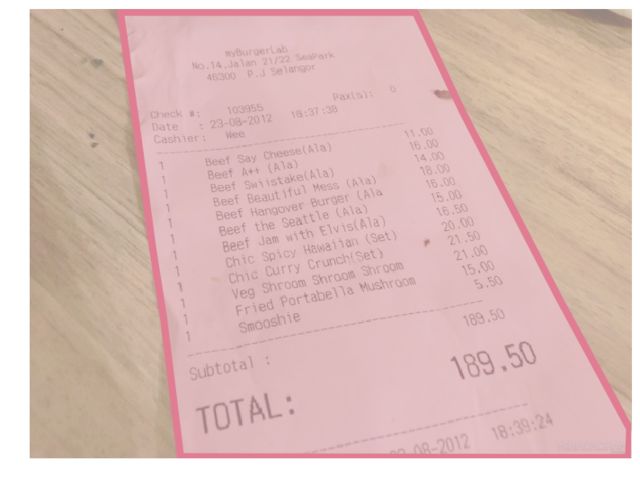}        &    A document indicating a financial transaction, such as receipts or checks          \\ \midrule
  Ticket (\attr{ticket})  & \includegraphics[width=0.17\textwidth]{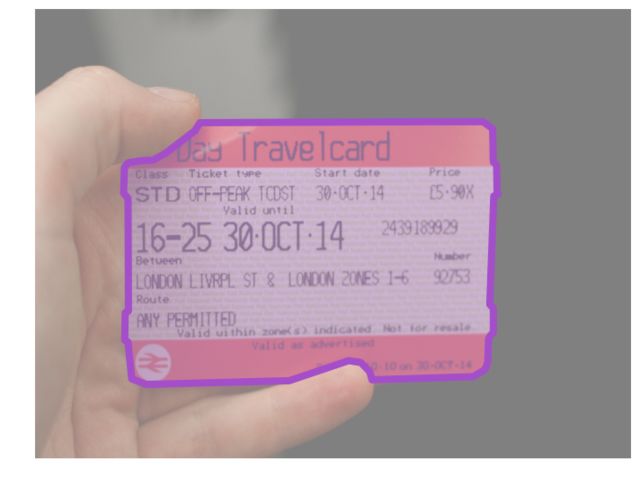}        &    A ticket, such as for travel, concert or sports match          \\ \bottomrule
  \end{tabular}
  \caption{Descriptions and examples of \textsc{Multimodal} attributes. For readability, we display images where  attributes are salient.}
  \label{fig:definitions_m}
\end{figure*}

\begin{figure*}[tbp]
  \centering
  \begin{tabular}{p{0.01cm}cccccc}
  & \attr{location} & \attr{home\_addr} & \attr{name} & \attr{birth\_dt} & \attr{phone\_no} & \attr{landmark} \\
  &         &            &      &           &           &          \\
  &         &            &  \multicolumn{2}{c}{good (iou $\ge$ 0.75)}           &           &          \\
  &         &            &      &           &           &          \\
  \rotatebox[origin=l]{90}{\quad \: GT} &
  \includegraphics[width=0.14\textwidth]{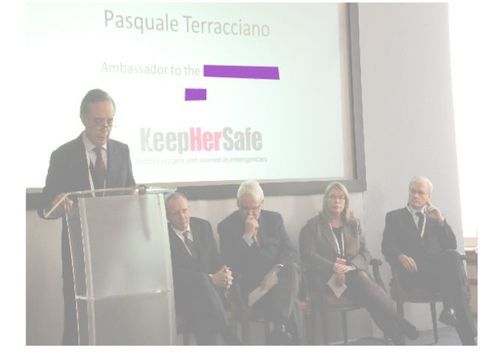} &
  \includegraphics[width=0.14\textwidth]{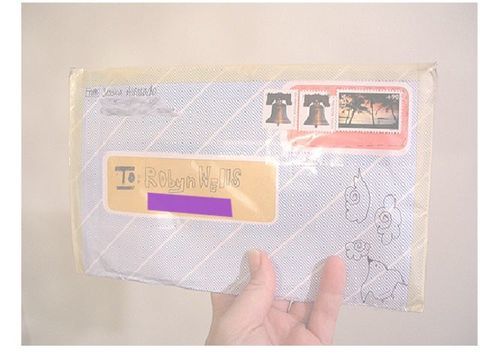} &
  \includegraphics[width=0.14\textwidth]{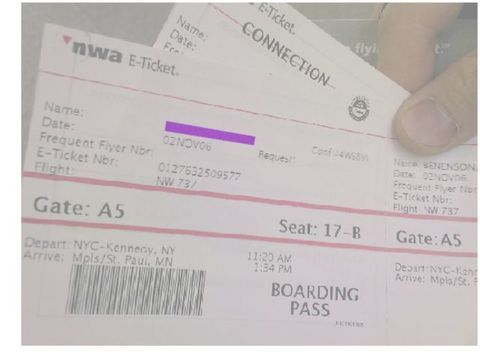} &
  \includegraphics[width=0.14\textwidth]{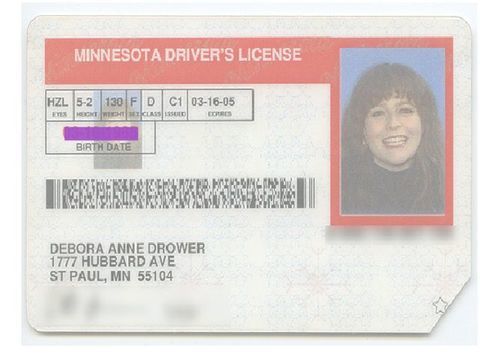} &
  \includegraphics[width=0.14\textwidth]{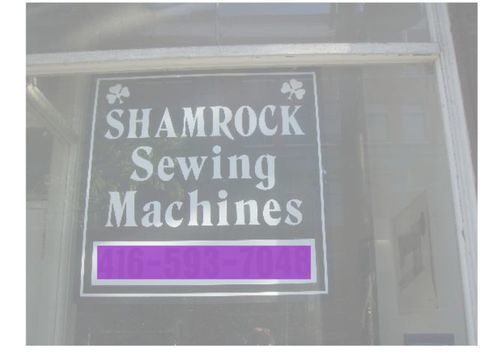} &
  \includegraphics[width=0.14\textwidth]{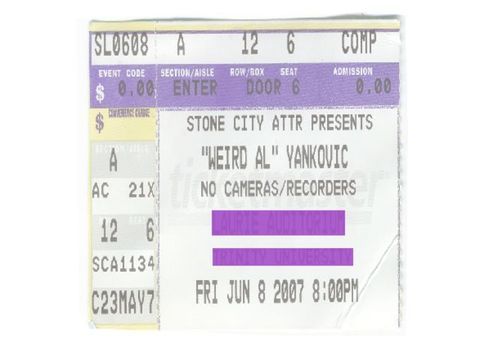} \\
  \rotatebox[origin=l]{90}{\: Predicted} &
  \includegraphics[width=0.14\textwidth]{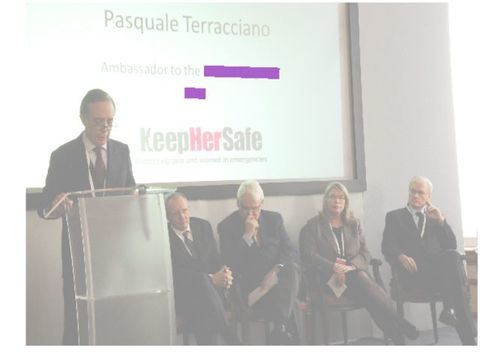} &
  \includegraphics[width=0.14\textwidth]{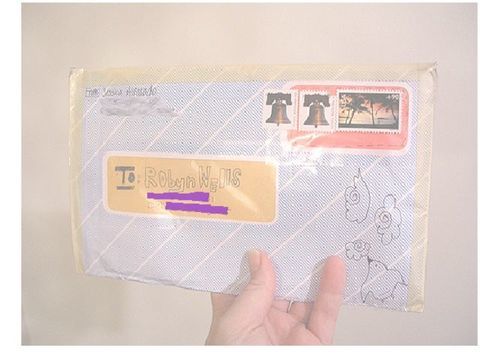} &
  \includegraphics[width=0.14\textwidth]{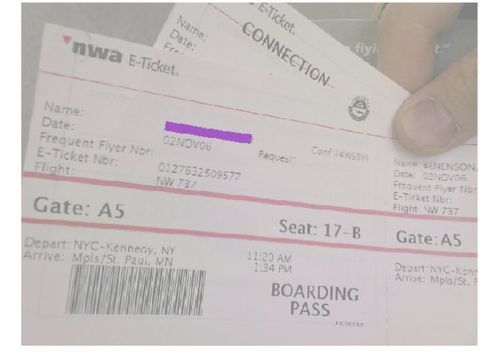} &
  \includegraphics[width=0.14\textwidth]{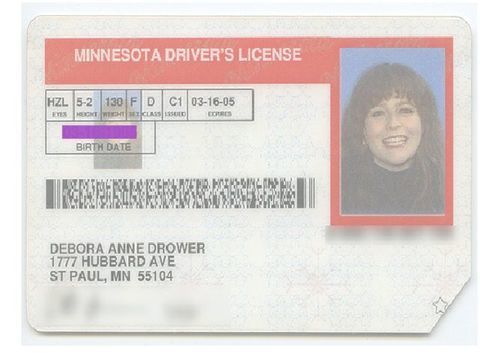} &
  \includegraphics[width=0.14\textwidth]{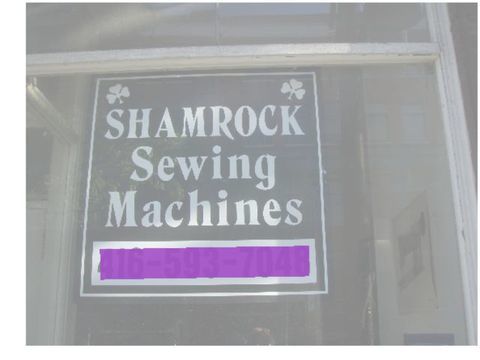} &
  \includegraphics[width=0.14\textwidth]{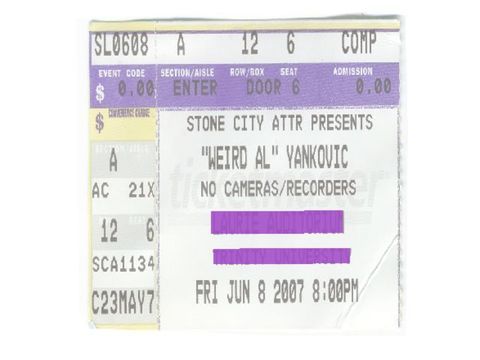} \\
  &         &            &      &           &           &          \\
  &         &            &  \multicolumn{2}{c}{mediocre (0.25 $\le$ iou $<$ 0.75)}           &           &          \\
  &         &            &      &           &           &          \\
           
  \rotatebox[origin=l]{90}{\quad \: GT} &
  \includegraphics[width=0.14\textwidth]{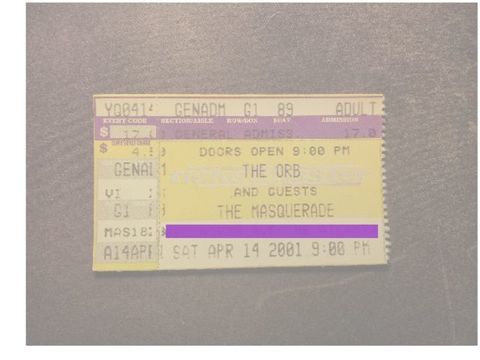} &
  \includegraphics[width=0.14\textwidth]{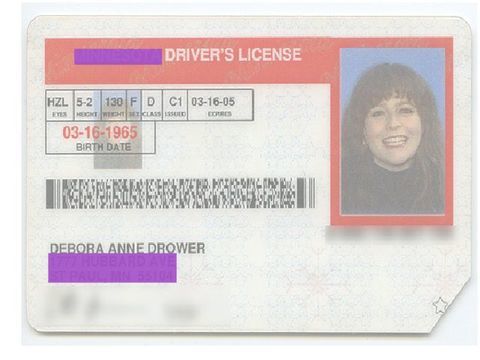} &
  \includegraphics[width=0.14\textwidth]{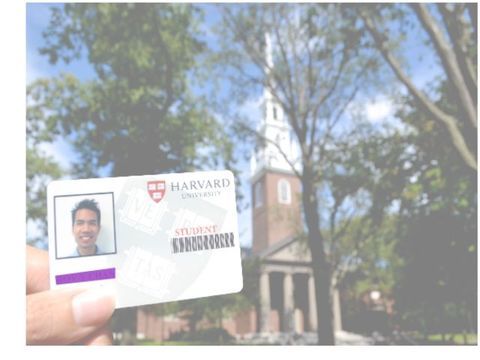} &
  \includegraphics[width=0.14\textwidth]{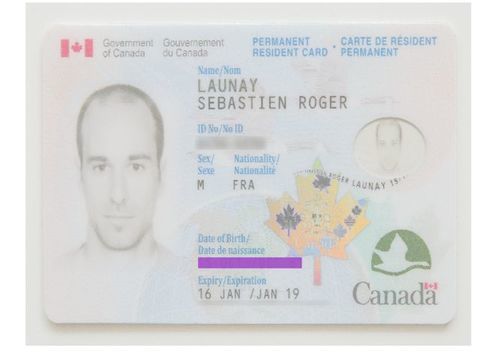} &
  \includegraphics[width=0.14\textwidth]{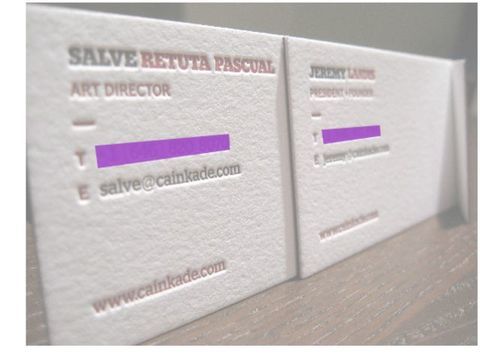} &
  \includegraphics[width=0.14\textwidth]{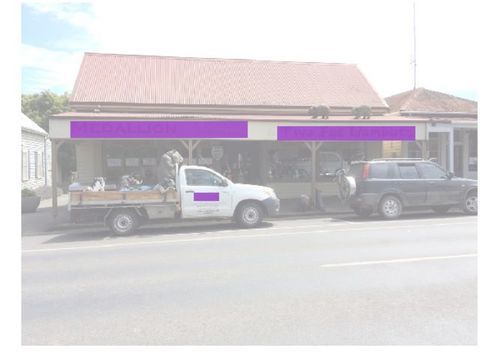} \\
  \rotatebox[origin=l]{90}{\: Predicted} &
  \includegraphics[width=0.14\textwidth]{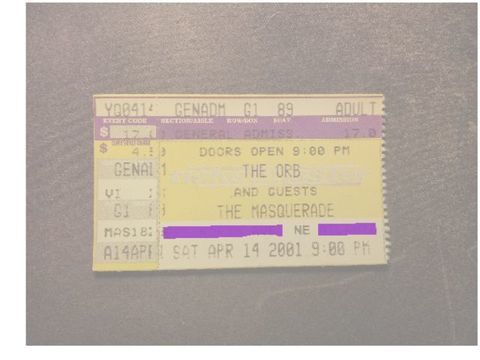} &
  \includegraphics[width=0.14\textwidth]{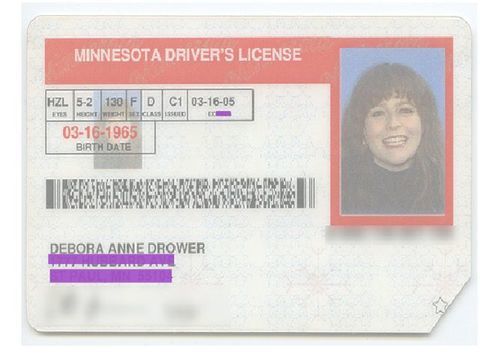} &
  \includegraphics[width=0.14\textwidth]{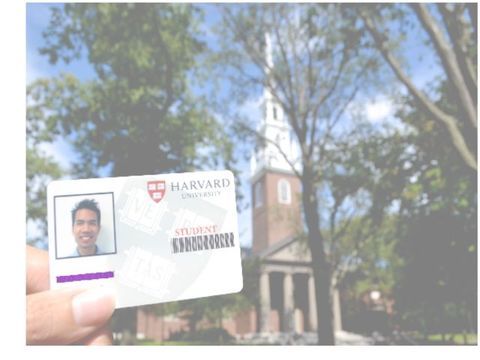} &
  \includegraphics[width=0.14\textwidth]{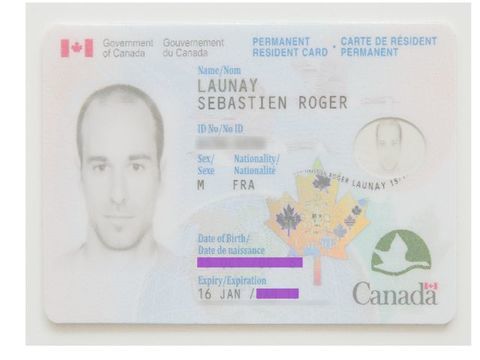} &
  \includegraphics[width=0.14\textwidth]{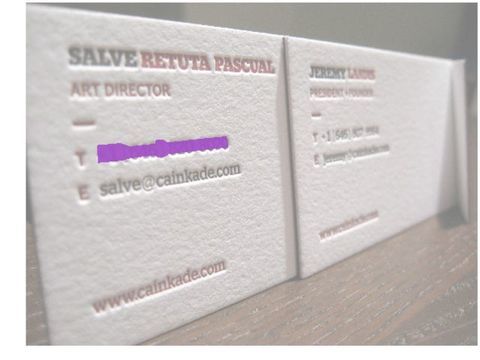} &
  \includegraphics[width=0.14\textwidth]{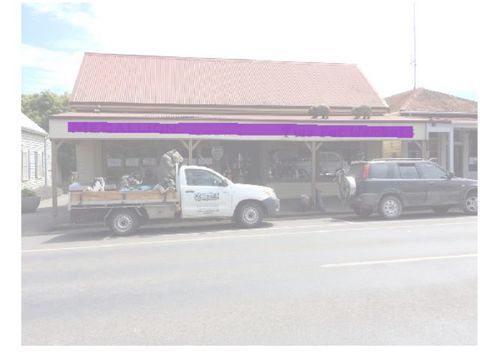} \\
  &         &            &      &           &           &          \\
           
  \rotatebox[origin=l]{90}{\quad \: GT} &
  \includegraphics[width=0.14\textwidth]{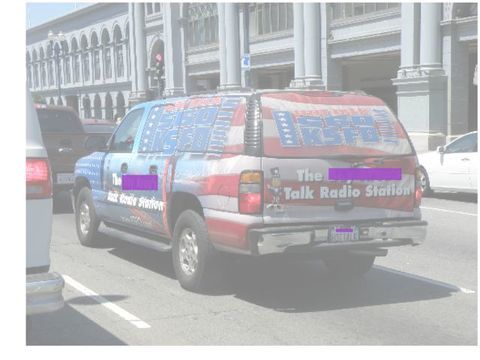} &
  \includegraphics[width=0.14\textwidth]{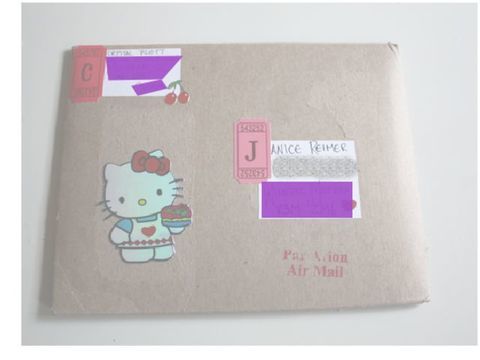} &
  \includegraphics[width=0.14\textwidth]{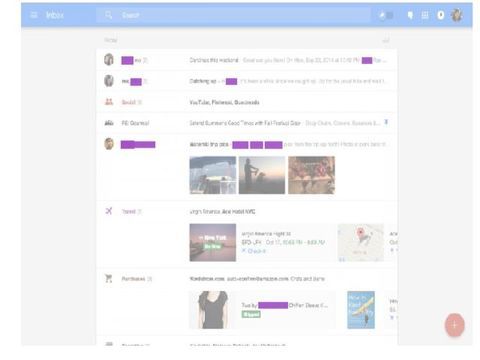} &
  \includegraphics[width=0.14\textwidth]{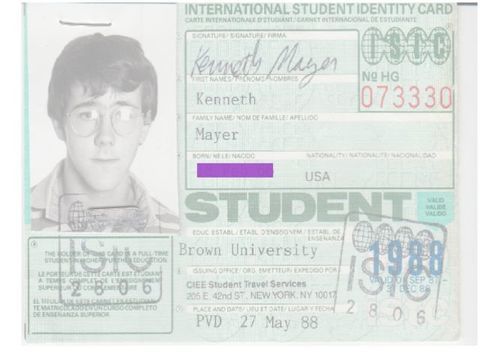} &
  \includegraphics[width=0.14\textwidth]{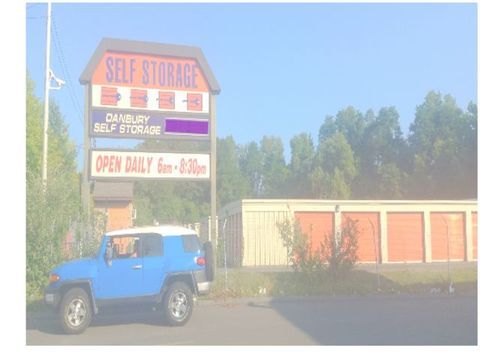} &
  \includegraphics[width=0.14\textwidth]{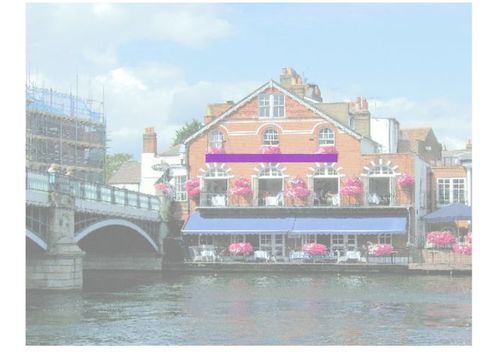} \\
  \rotatebox[origin=l]{90}{\: Predicted} &
  \includegraphics[width=0.14\textwidth]{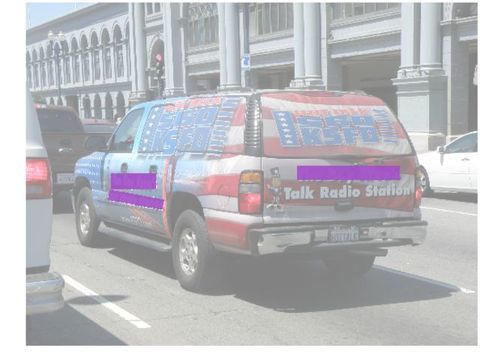} &
  \includegraphics[width=0.14\textwidth]{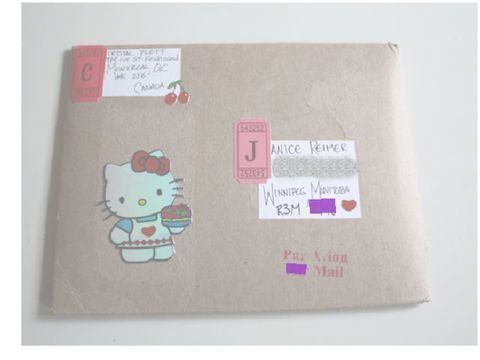} &
  \includegraphics[width=0.14\textwidth]{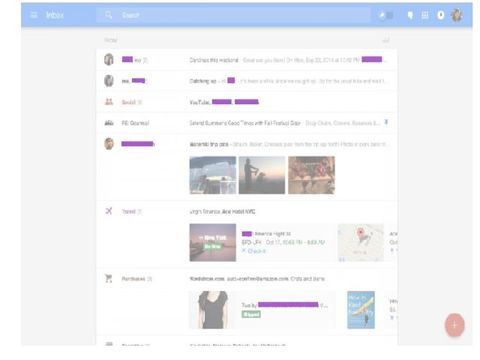} &
  \includegraphics[width=0.14\textwidth]{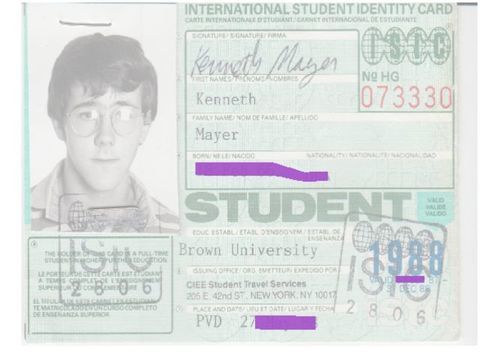} &
  \includegraphics[width=0.14\textwidth]{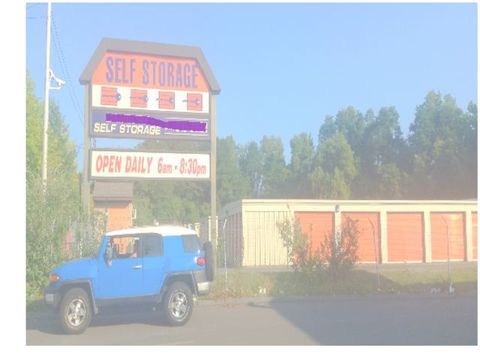} &
  \includegraphics[width=0.14\textwidth]{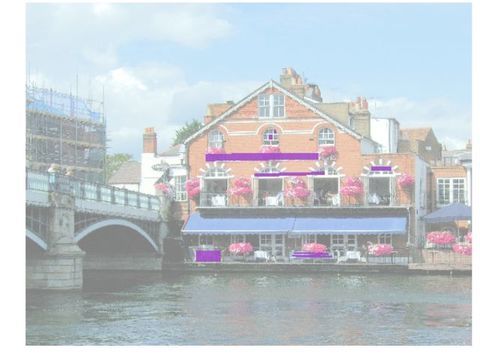} \\
  &         &            &      &           &           &          \\
  &         &            &  \multicolumn{2}{c}{failure (iou $\approx$ 0)}           &           &          \\
  &         &            &      &           &           &          \\
           
  \rotatebox[origin=l]{90}{\quad \: GT} &
  \includegraphics[width=0.14\textwidth]{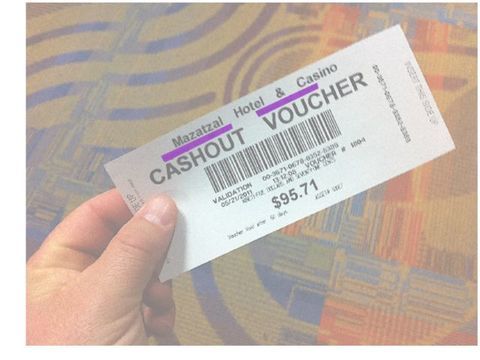} &
  \includegraphics[width=0.14\textwidth]{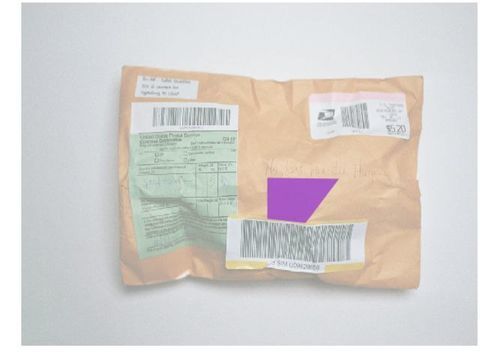} &
  \includegraphics[width=0.14\textwidth]{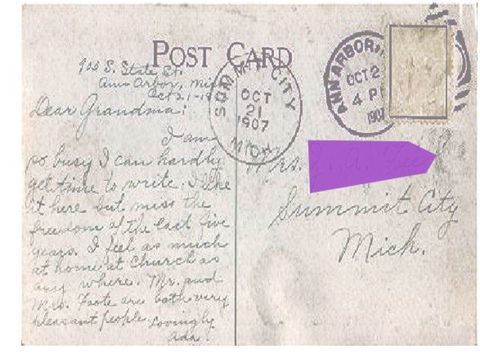} &
  \includegraphics[width=0.14\textwidth]{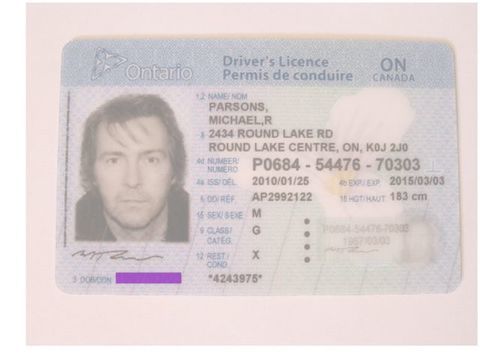} &
  \includegraphics[width=0.14\textwidth]{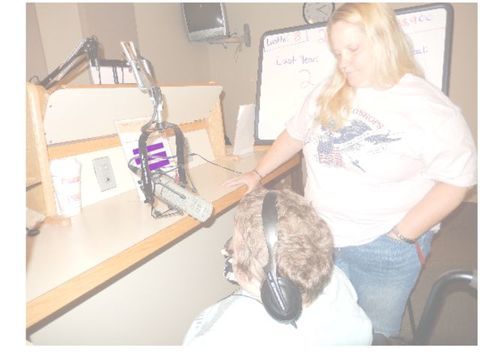} &
  \includegraphics[width=0.14\textwidth]{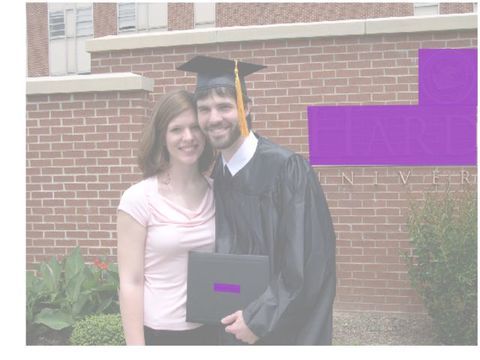} \\
  \rotatebox[origin=l]{90}{\: Predicted} &
  \includegraphics[width=0.14\textwidth]{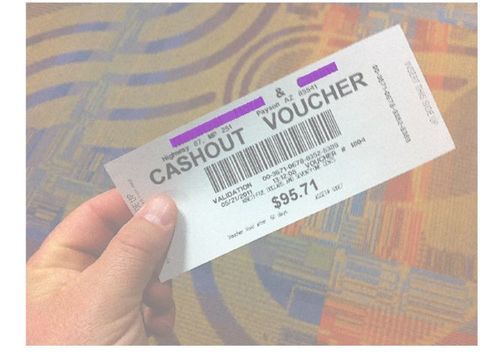} &
  \includegraphics[width=0.14\textwidth]{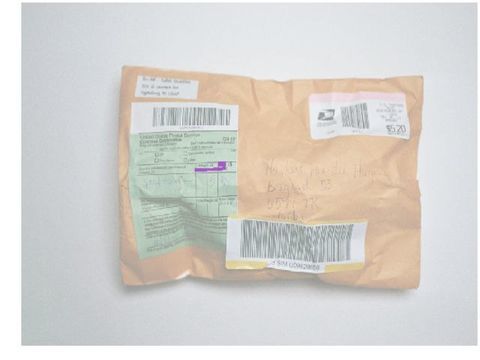} &
  \includegraphics[width=0.14\textwidth]{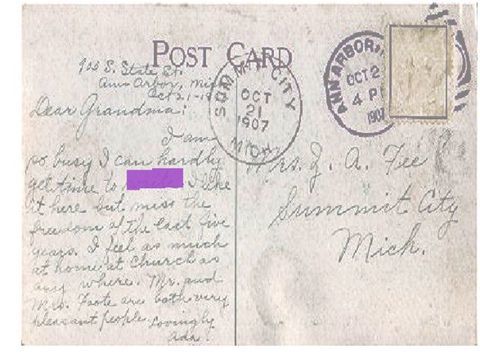} &
  \includegraphics[width=0.14\textwidth]{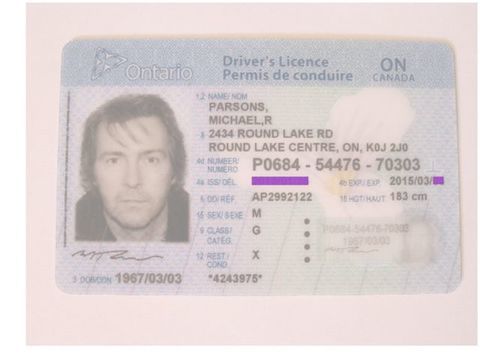} &
  \includegraphics[width=0.14\textwidth]{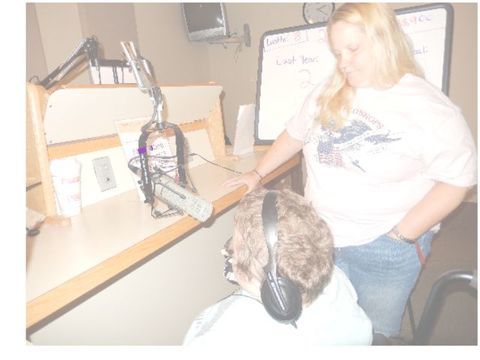} &
  \includegraphics[width=0.14\textwidth]{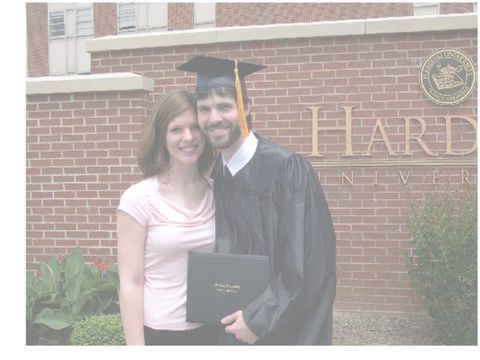}
  \end{tabular}
  \caption{Qualitative results per attribute. In each pair of images, top is ground-truth segmentation and bottom is prediction. Pairs of images in each column are sorted by IoU scores (high to low). 
  }
  \label{fig:qual1}
\end{figure*}

\begin{figure*}[tbp]
  \centering
  \begin{tabular}{p{0.01cm}cccccc}
  & \attr{datetime} & \attr{emailadd} & \attr{face} & \attr{lic\_plate} & \attr{person} & \attr{nudity} \\
  &         &            &      &           &           &          \\
  &         &            &  \multicolumn{2}{c}{good (iou $\ge$ 0.75)}           &           &          \\
  &         &            &      &           &           &          \\
  \rotatebox[origin=l]{90}{\quad \: GT} &
  \includegraphics[width=0.14\textwidth]{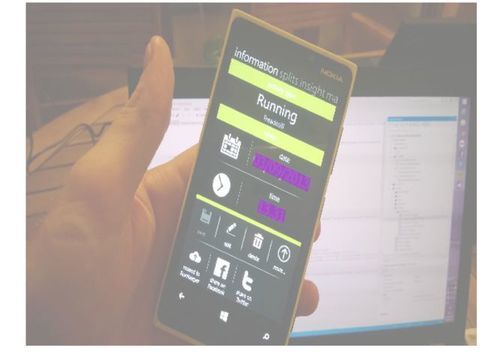} &
  \includegraphics[width=0.14\textwidth]{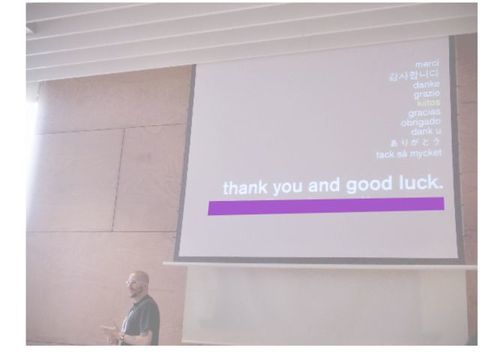} &
  \includegraphics[width=0.14\textwidth]{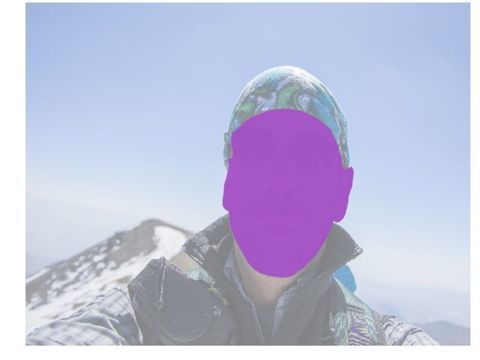} &
  \includegraphics[width=0.14\textwidth]{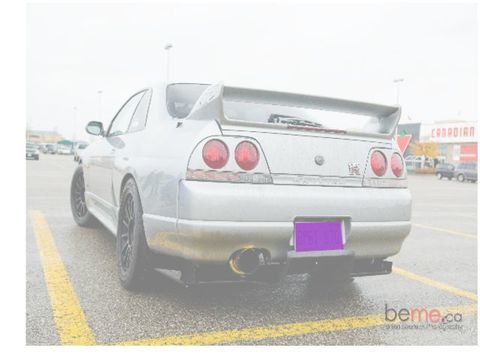} &
  \includegraphics[width=0.14\textwidth]{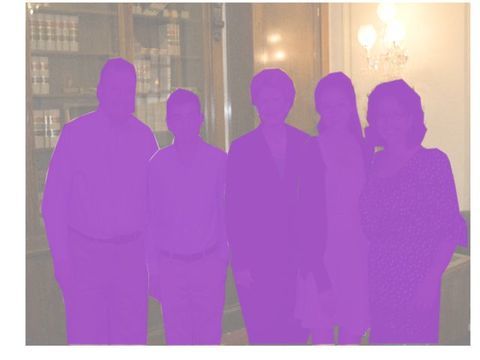} &
  \includegraphics[width=0.14\textwidth]{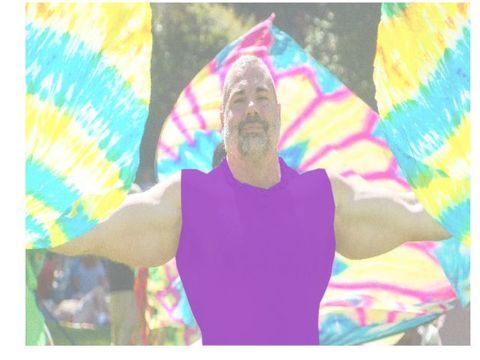} \\
  \rotatebox[origin=l]{90}{\: Predicted} &
  \includegraphics[width=0.14\textwidth]{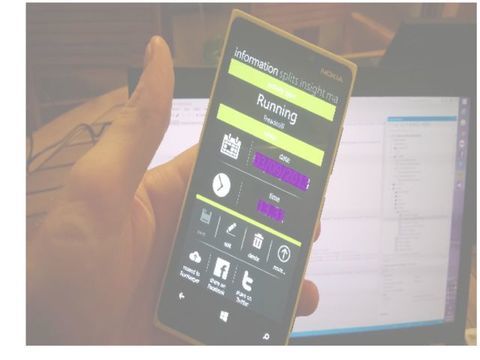} &
  \includegraphics[width=0.14\textwidth]{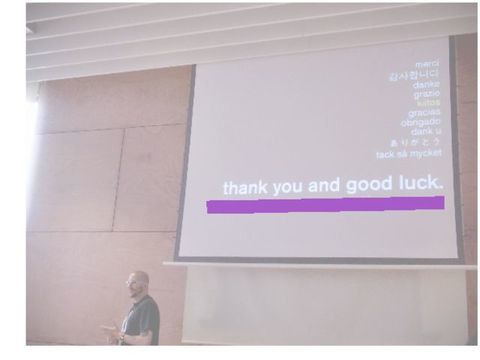} &
  \includegraphics[width=0.14\textwidth]{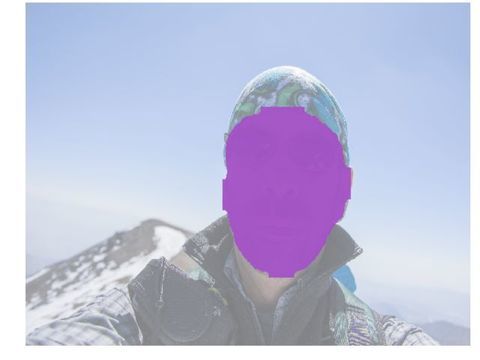} &
  \includegraphics[width=0.14\textwidth]{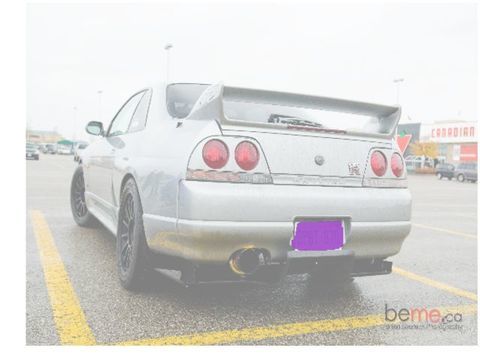} &
  \includegraphics[width=0.14\textwidth]{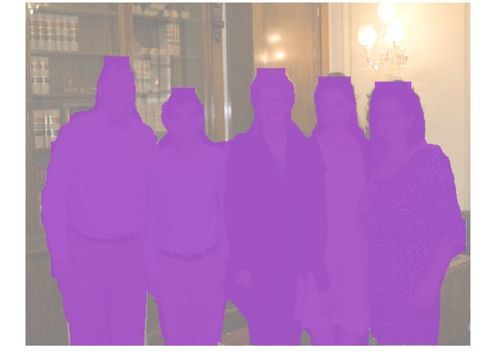} &
  \includegraphics[width=0.14\textwidth]{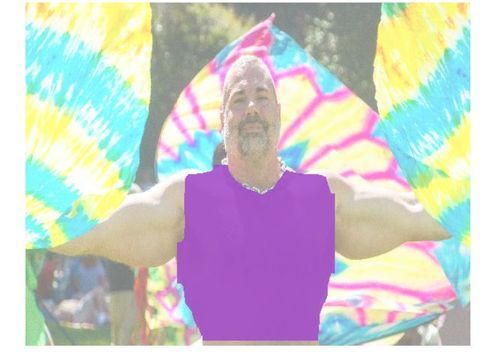} \\
  &         &            &      &           &           &          \\
  &         &            &  \multicolumn{2}{c}{mediocre (0.25 $\le$ iou $<$ 0.75)}           &           &          \\
  &         &            &      &           &           &          \\
           
  \rotatebox[origin=l]{90}{\quad \: GT} &
  \includegraphics[width=0.14\textwidth]{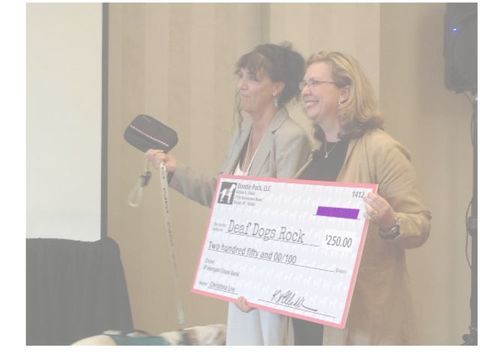} &
  \includegraphics[width=0.14\textwidth]{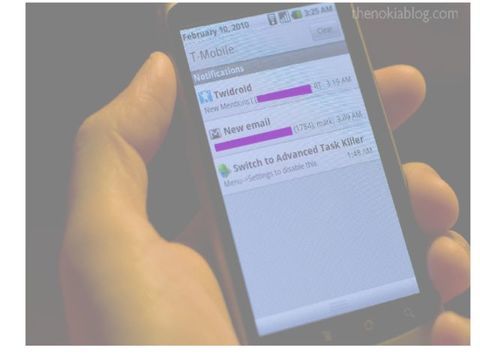} &
  \includegraphics[width=0.14\textwidth]{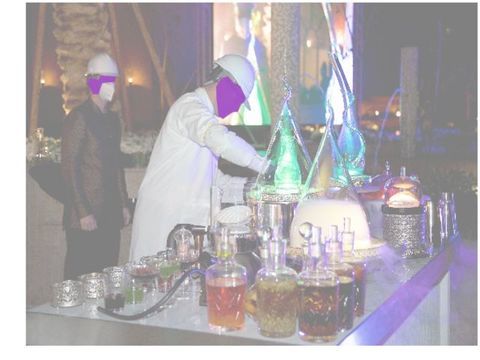} &
  \includegraphics[width=0.14\textwidth]{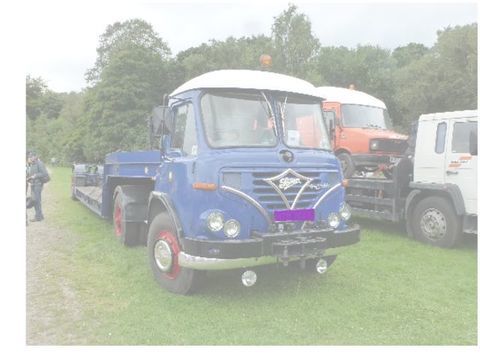} &
  \includegraphics[width=0.14\textwidth]{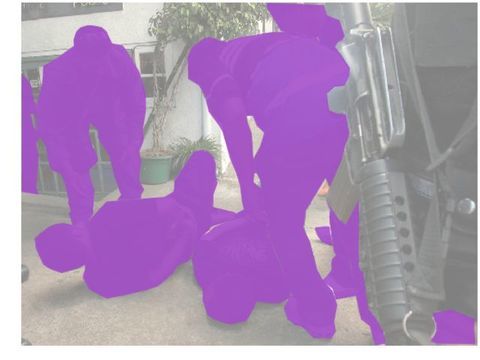} &
  \includegraphics[width=0.14\textwidth]{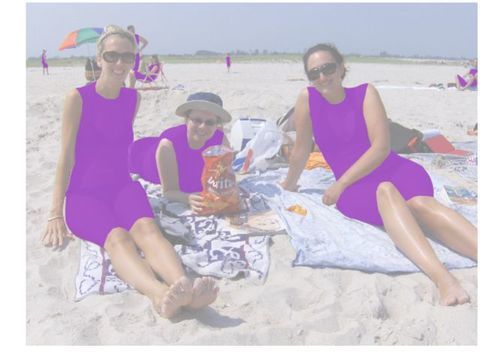} \\
  \rotatebox[origin=l]{90}{\: Predicted} &
  \includegraphics[width=0.14\textwidth]{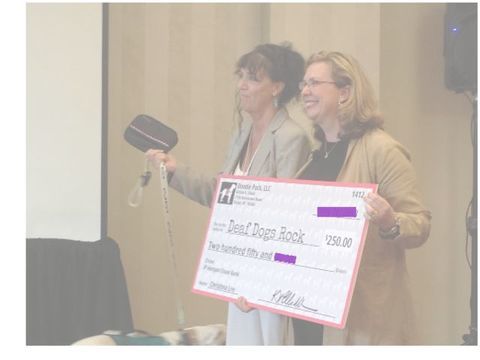} &
  \includegraphics[width=0.14\textwidth]{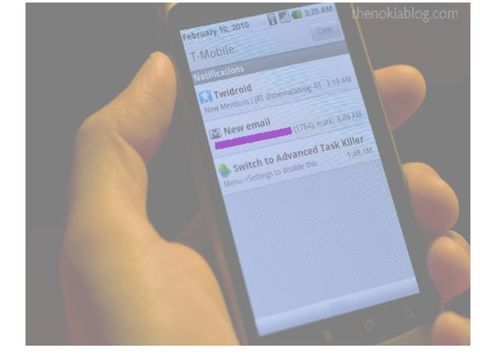} &
  \includegraphics[width=0.14\textwidth]{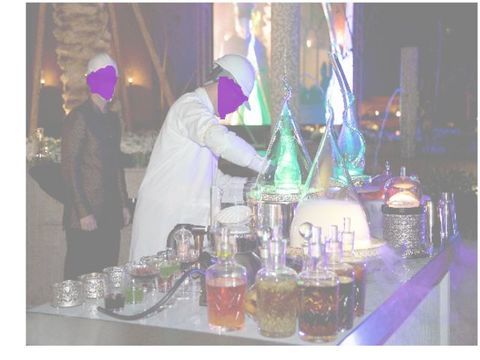} &
  \includegraphics[width=0.14\textwidth]{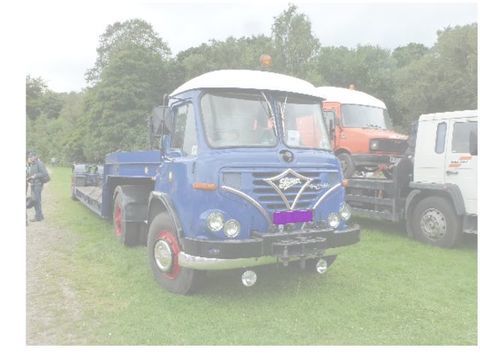} &
  \includegraphics[width=0.14\textwidth]{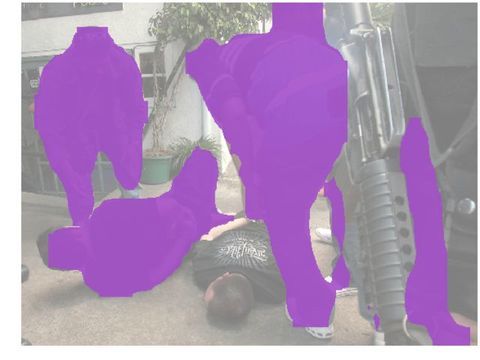} &
  \includegraphics[width=0.14\textwidth]{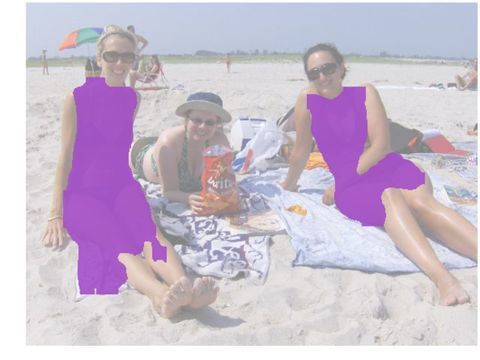} \\
  &         &            &      &           &           &          \\
           
  \rotatebox[origin=l]{90}{\quad \: GT} &
  \includegraphics[width=0.14\textwidth]{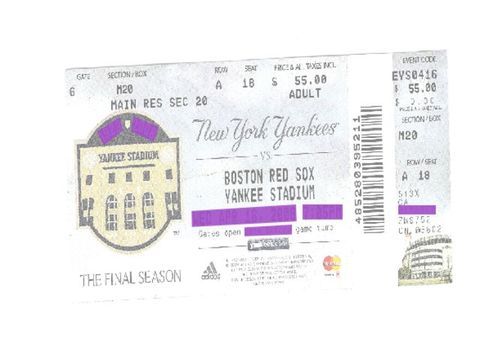} &
  \includegraphics[width=0.14\textwidth]{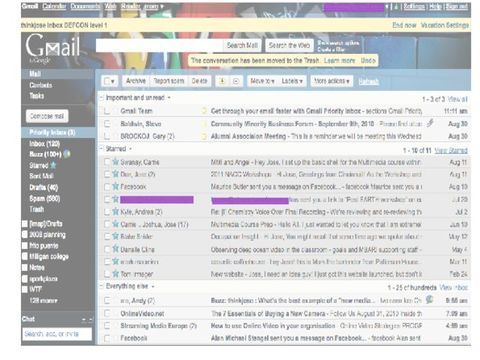} &
  \includegraphics[width=0.14\textwidth]{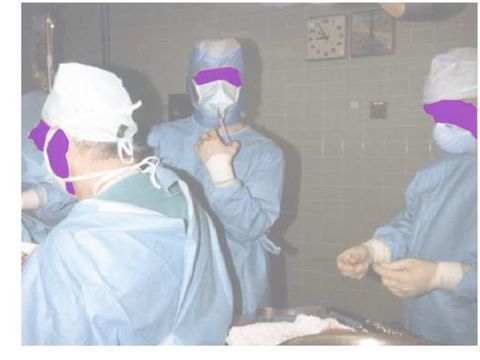} &
  \includegraphics[width=0.14\textwidth]{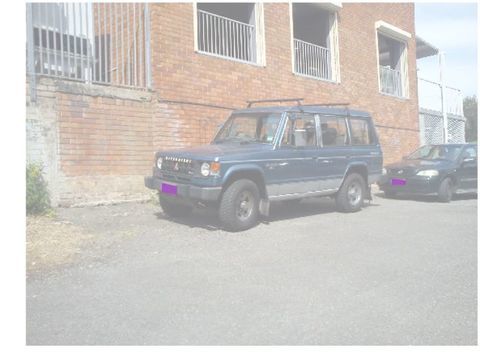} &
  \includegraphics[width=0.14\textwidth]{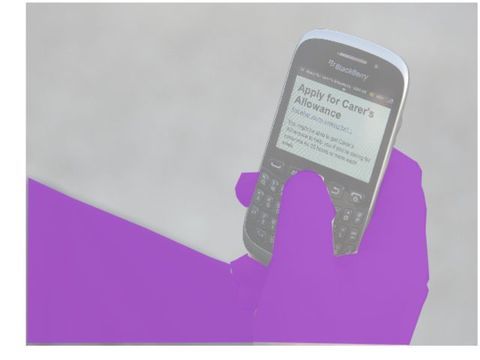} &
  \includegraphics[width=0.14\textwidth]{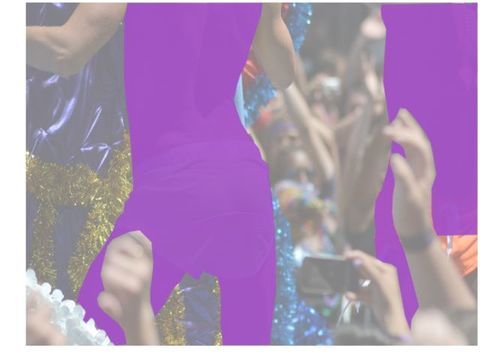} \\
  \rotatebox[origin=l]{90}{\: Predicted} &
  \includegraphics[width=0.14\textwidth]{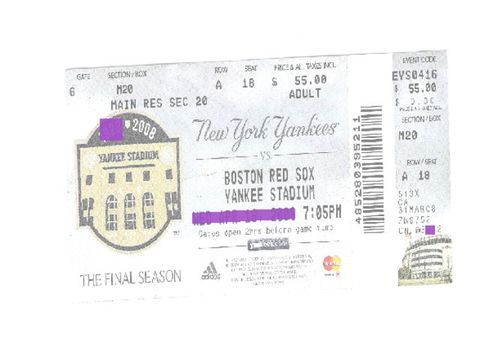} &
  \includegraphics[width=0.14\textwidth]{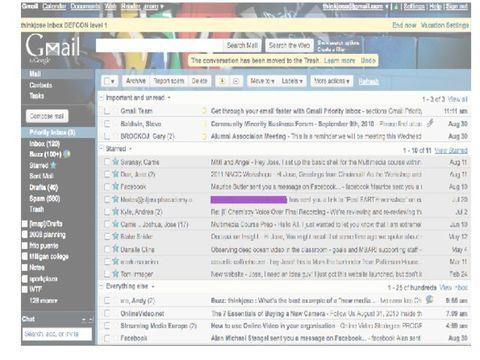} &
  \includegraphics[width=0.14\textwidth]{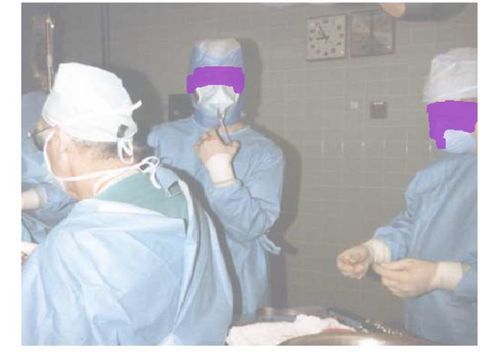} &
  \includegraphics[width=0.14\textwidth]{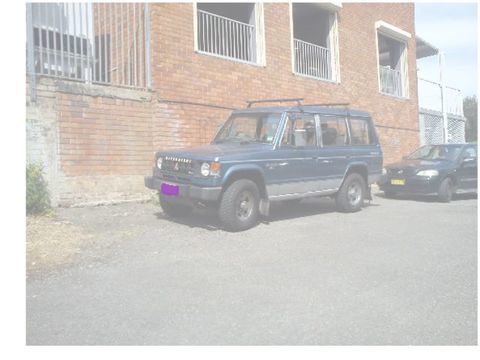} &
  \includegraphics[width=0.14\textwidth]{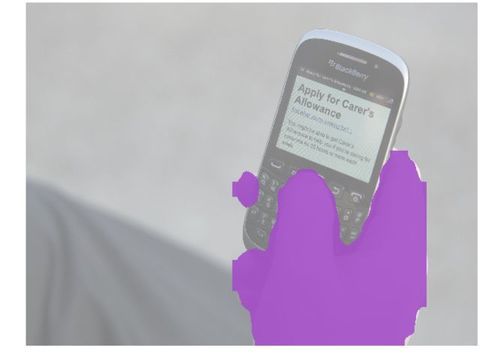} &
  \includegraphics[width=0.14\textwidth]{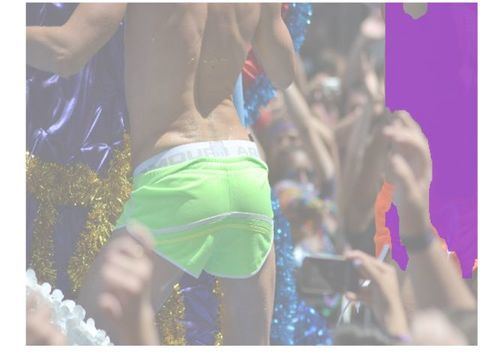} \\
  &         &            &      &           &           &          \\
  &         &            &  \multicolumn{2}{c}{failure (iou $\approx$ 0)}           &           &          \\
  &         &            &      &           &           &          \\
           
  \rotatebox[origin=l]{90}{\quad \: GT} &
  \includegraphics[width=0.14\textwidth]{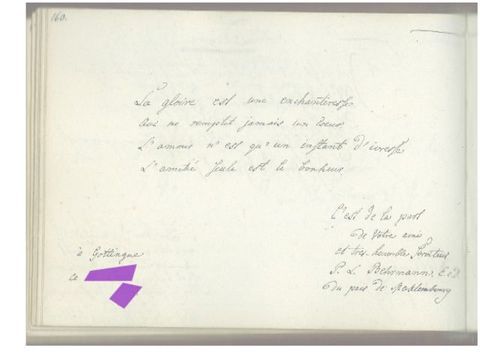} &
  \includegraphics[width=0.14\textwidth]{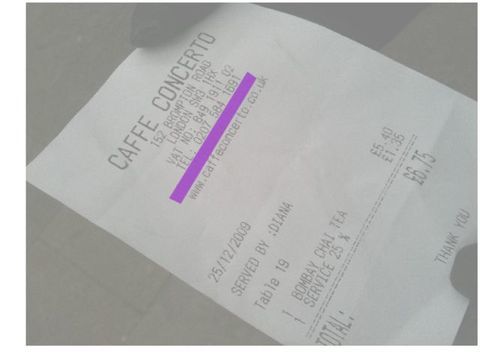} &
  \includegraphics[width=0.14\textwidth]{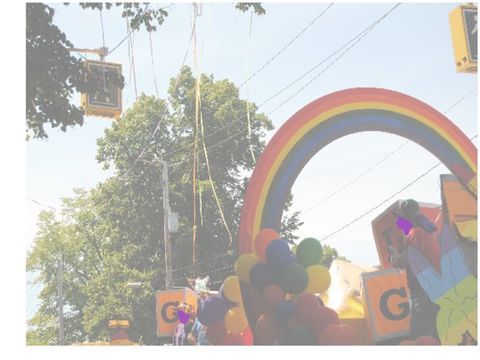} &
  \includegraphics[width=0.14\textwidth]{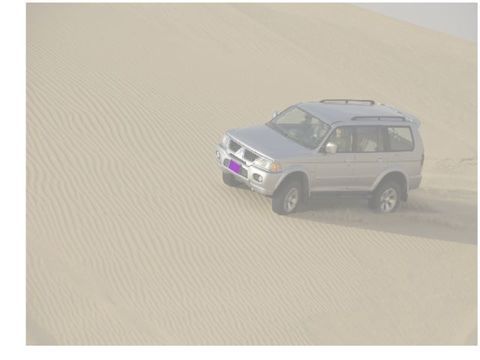} &
  \includegraphics[width=0.14\textwidth]{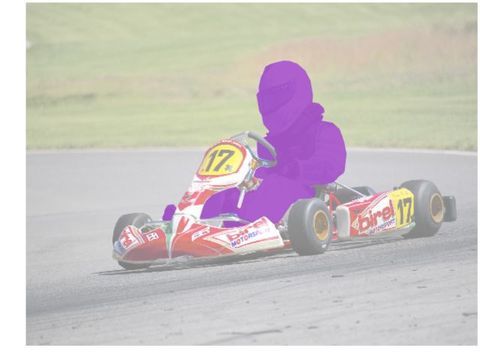} &
  \includegraphics[width=0.14\textwidth]{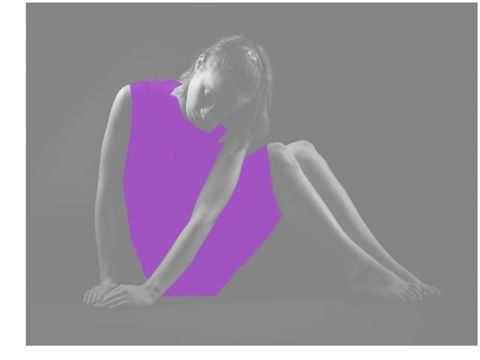} \\
  \rotatebox[origin=l]{90}{\: Predicted} &
  \includegraphics[width=0.14\textwidth]{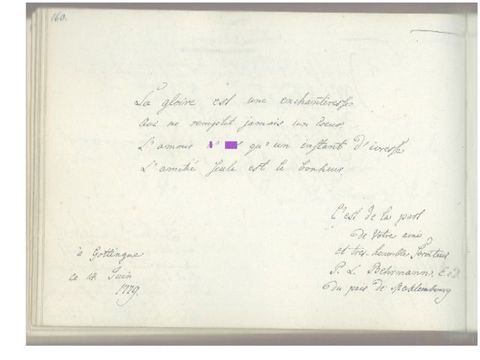} &
  \includegraphics[width=0.14\textwidth]{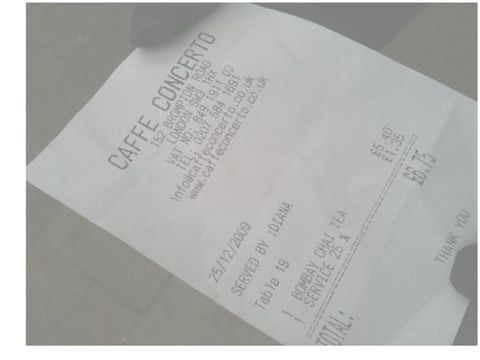} &
  \includegraphics[width=0.14\textwidth]{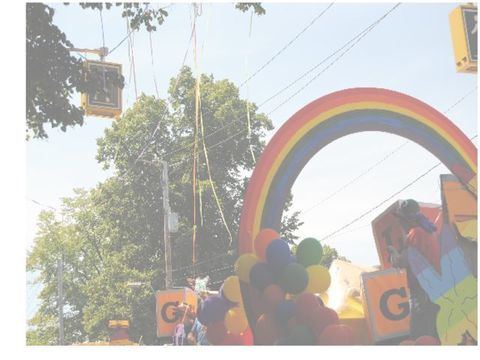} &
  \includegraphics[width=0.14\textwidth]{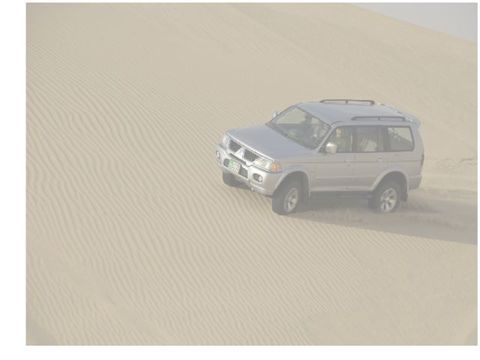} &
  \includegraphics[width=0.14\textwidth]{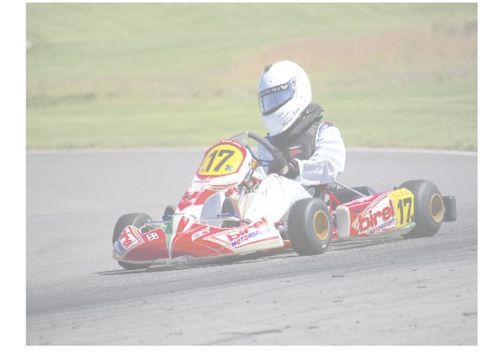} &
  \includegraphics[width=0.14\textwidth]{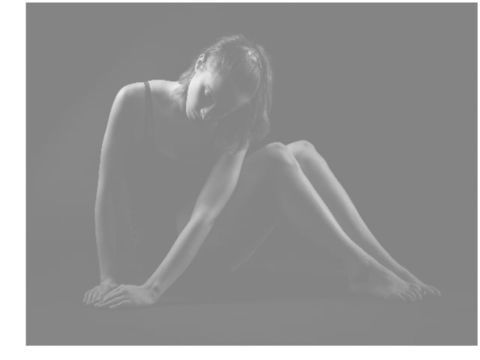}
  \end{tabular}
  \caption{Qualitative results per attribute. In each pair of images, top is ground-truth segmentation and bottom is prediction. Pairs of images in each column are sorted by IoU scores (high to low).}
  \label{fig:qual1}
\end{figure*}

\begin{figure*}[tbp]
  \centering
  \begin{tabular}{p{0.1cm}cccccc}
  & \attr{handwrit} & \attr{phy\_disb} & \attr{med\_hist} & \attr{fingerpr} & \attr{signtr} & \attr{cr\_card} \\
   &        &            &      &           &           &          \\
  &         &            &  \multicolumn{2}{c}{good (iou $\ge$ 0.75)}           &           &          \\
  &         &            &      &           &           &          \\
  \rotatebox[origin=l]{90}{\quad \: GT} &
  \includegraphics[width=0.14\textwidth]{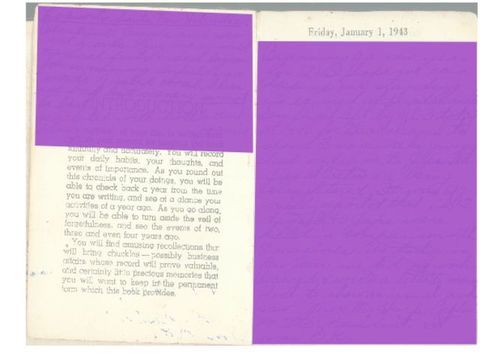} &
  \includegraphics[width=0.14\textwidth]{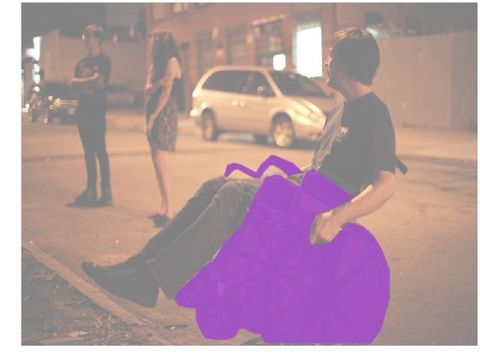} &
  \includegraphics[width=0.14\textwidth]{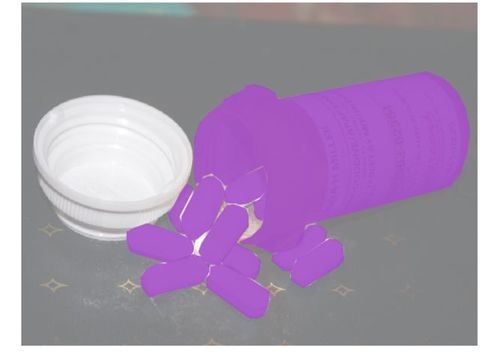} &
  \includegraphics[width=0.14\textwidth]{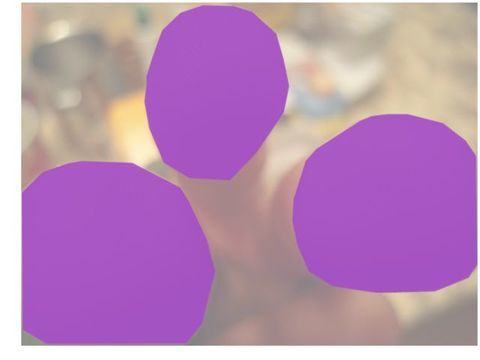} &
  \includegraphics[width=0.14\textwidth]{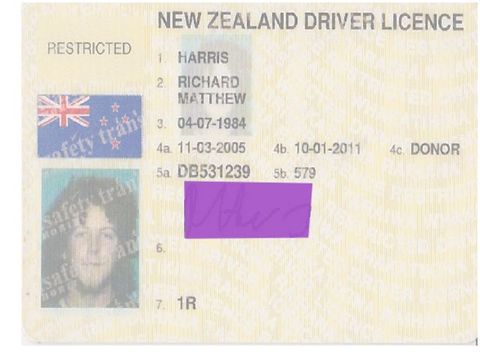} &
  \includegraphics[width=0.14\textwidth]{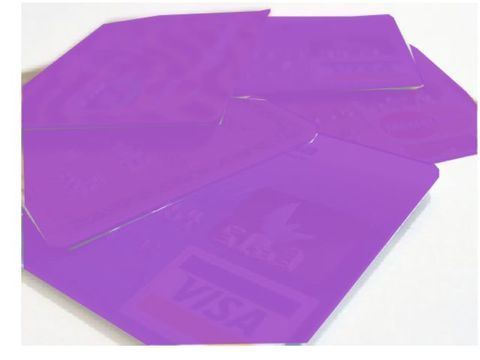} \\
  \rotatebox[origin=l]{90}{\: Predicted} &
  \includegraphics[width=0.14\textwidth]{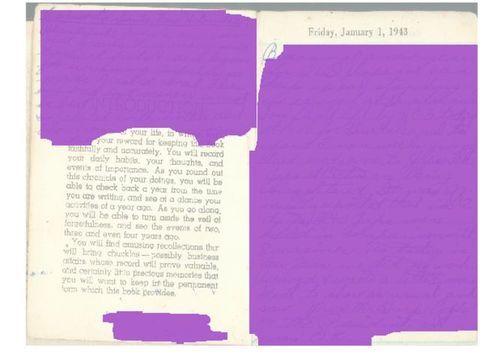} &
  \includegraphics[width=0.14\textwidth]{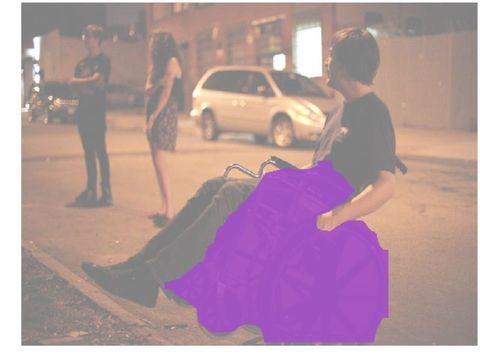} &
  \includegraphics[width=0.14\textwidth]{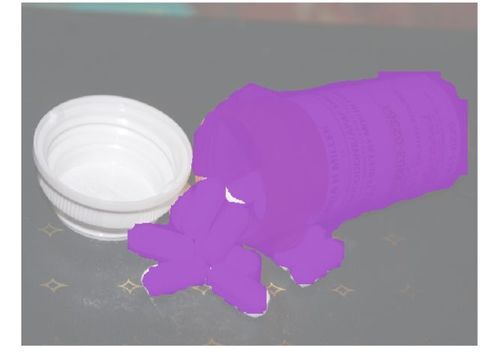} &
  \includegraphics[width=0.14\textwidth]{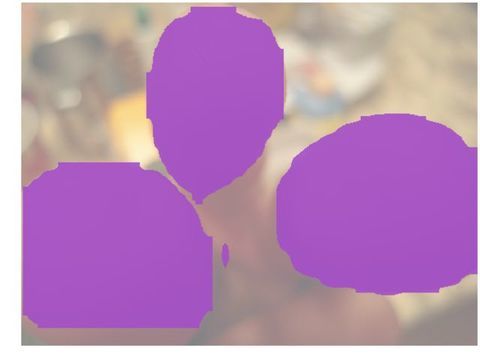} &
  \includegraphics[width=0.14\textwidth]{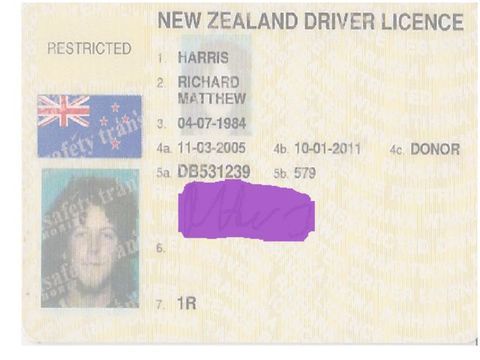} &
  \includegraphics[width=0.14\textwidth]{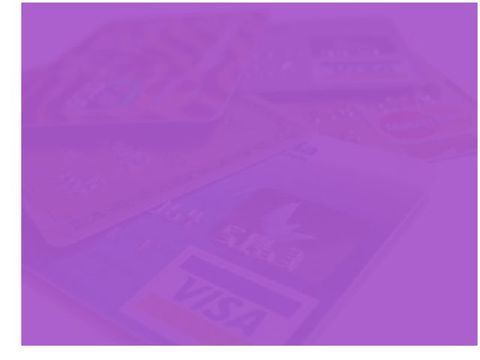} \\
   &        &            &      &           &           &          \\
   &        &            &  \multicolumn{2}{c}{mediocre (0.25 $\le$ iou $<$ 0.75)}           &           &          \\
   &        &            &      &           &           &          \\
           
  \rotatebox[origin=l]{90}{\quad \: GT} &
  \includegraphics[width=0.14\textwidth]{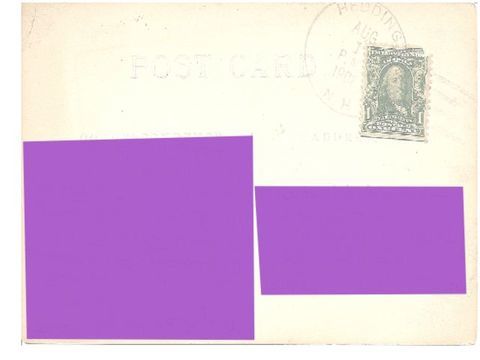} &
  \includegraphics[width=0.14\textwidth]{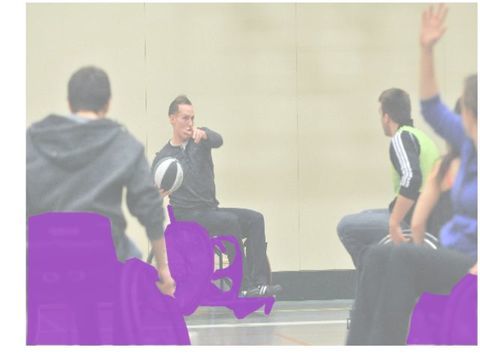} &
  \includegraphics[width=0.14\textwidth]{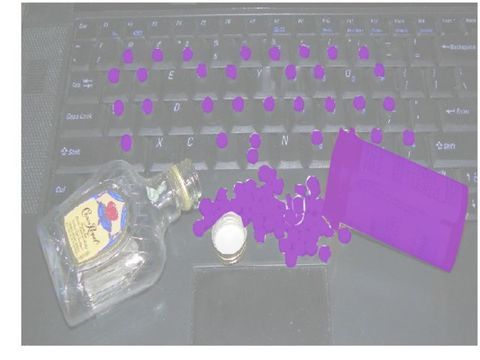} &
  \includegraphics[width=0.14\textwidth]{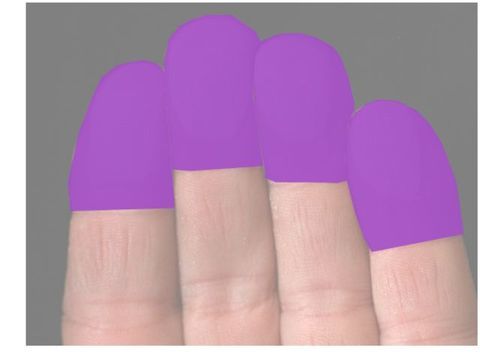} &
  \includegraphics[width=0.14\textwidth]{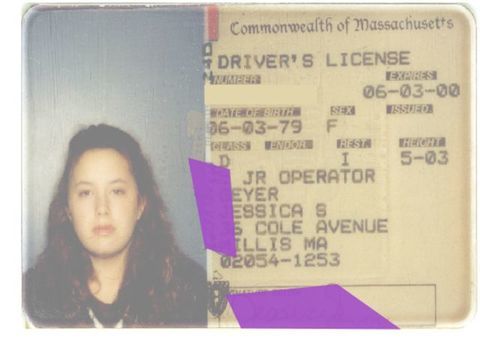} &
  \includegraphics[width=0.14\textwidth]{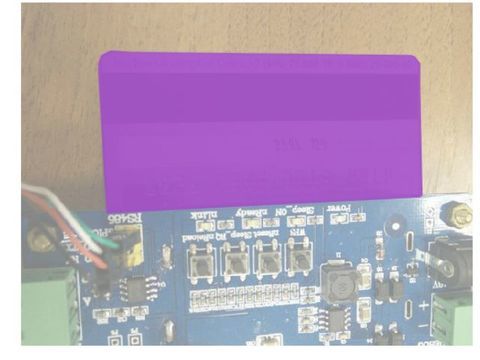} \\
  \rotatebox[origin=l]{90}{\: Predicted} &
  \includegraphics[width=0.14\textwidth]{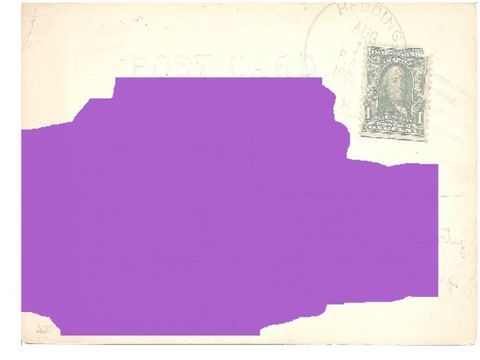} &
  \includegraphics[width=0.14\textwidth]{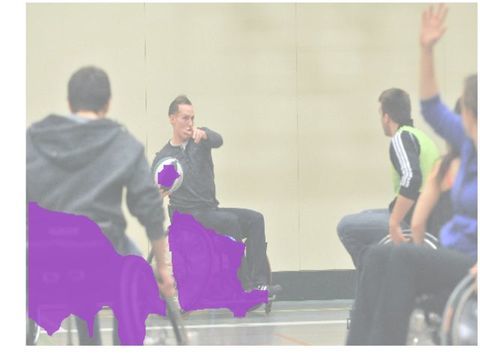} &
  \includegraphics[width=0.14\textwidth]{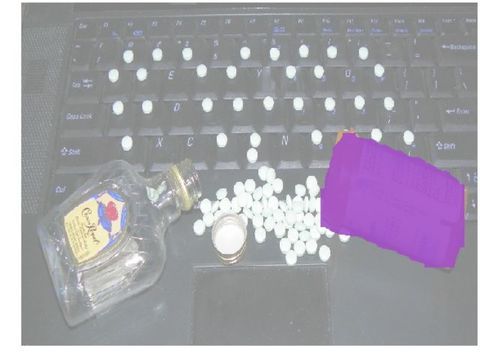} &
  \includegraphics[width=0.14\textwidth]{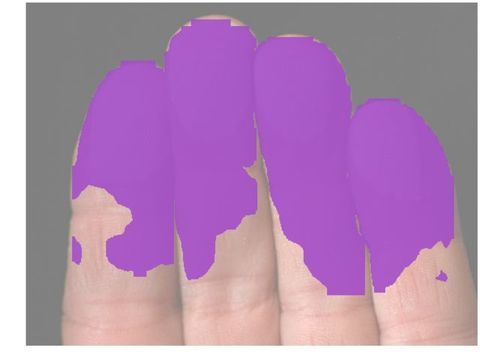} &
  \includegraphics[width=0.14\textwidth]{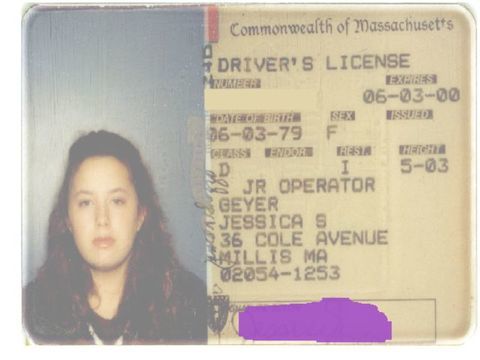} &
  \includegraphics[width=0.14\textwidth]{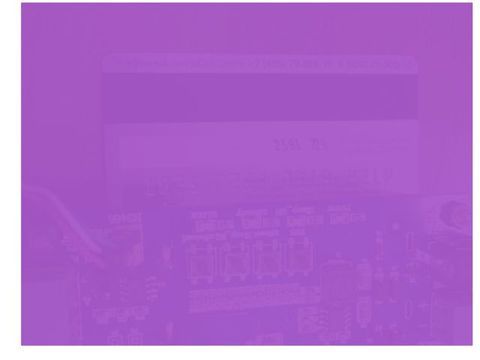} \\
   &        &            &      &           &           &          \\
           
  \rotatebox[origin=l]{90}{\quad \: GT} &
  \includegraphics[width=0.14\textwidth]{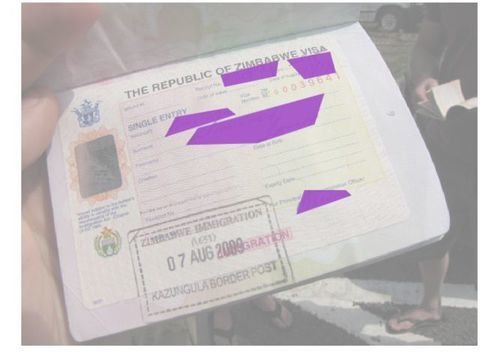} &
  \includegraphics[width=0.14\textwidth]{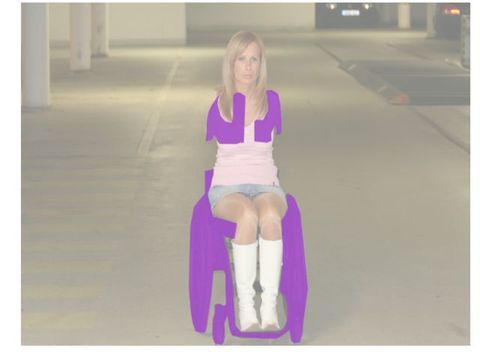} &
  \includegraphics[width=0.14\textwidth]{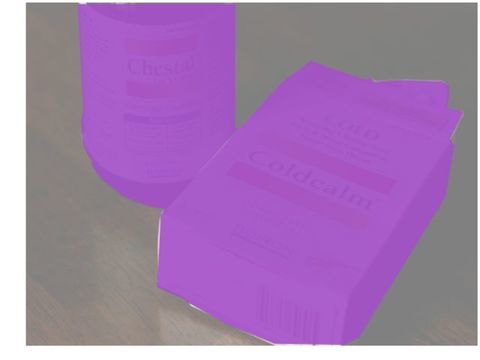} &
  \includegraphics[width=0.14\textwidth]{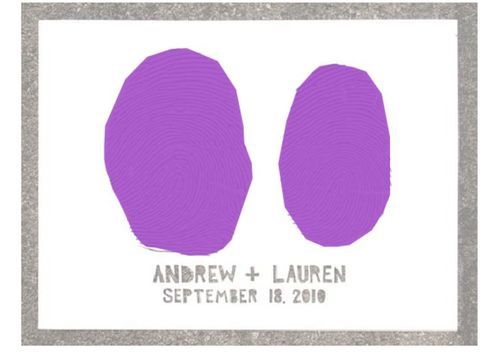} &
  \includegraphics[width=0.14\textwidth]{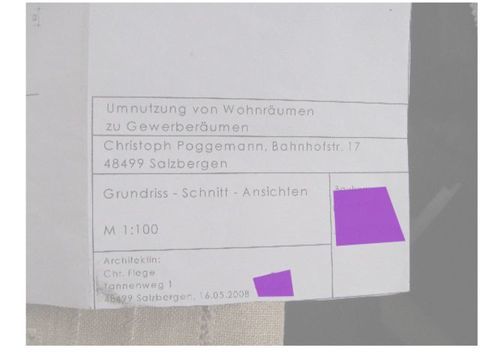} &
  \includegraphics[width=0.14\textwidth]{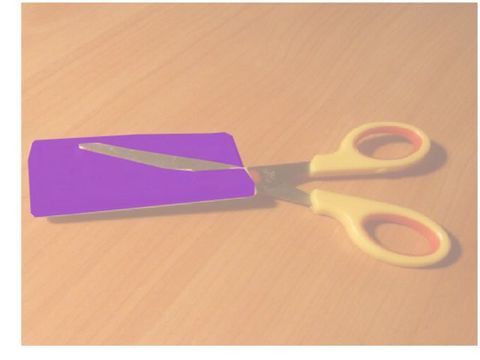} \\
  \rotatebox[origin=l]{90}{\: Predicted} &
  \includegraphics[width=0.14\textwidth]{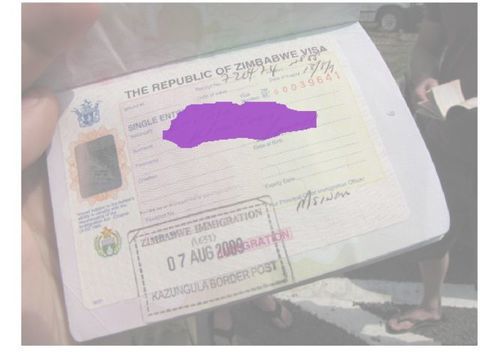} &
  \includegraphics[width=0.14\textwidth]{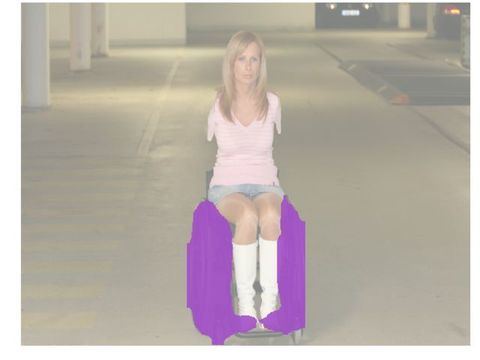} &
  \includegraphics[width=0.14\textwidth]{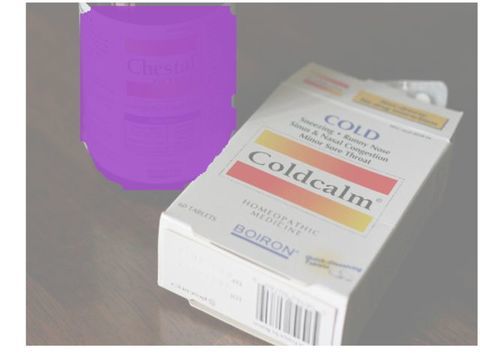} &
  \includegraphics[width=0.14\textwidth]{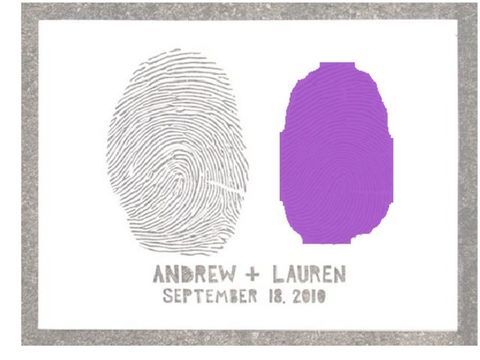} &
  \includegraphics[width=0.14\textwidth]{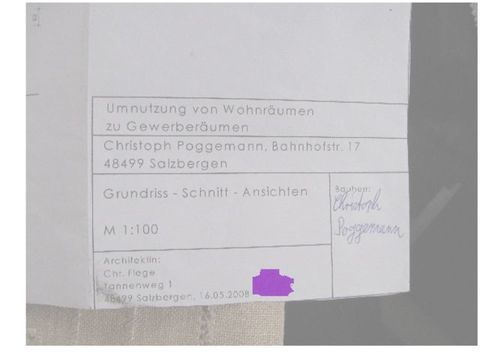} &
  \includegraphics[width=0.14\textwidth]{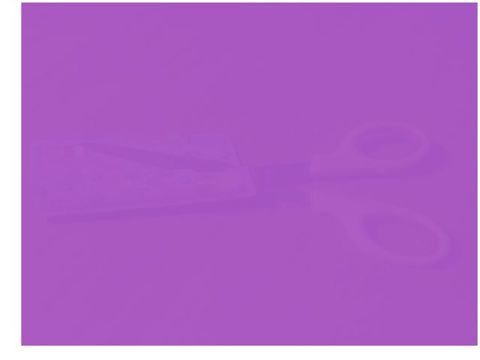} \\
   &        &            &      &           &           &          \\
   &        &            &  \multicolumn{2}{c}{failure (iou $\approx$ 0)}           &           &          \\
   &        &            &      &           &           &          \\
           
  \rotatebox[origin=l]{90}{\quad \: GT} &
  \includegraphics[width=0.14\textwidth]{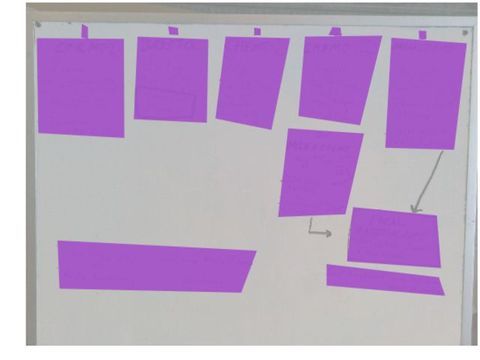} &
  \includegraphics[width=0.14\textwidth]{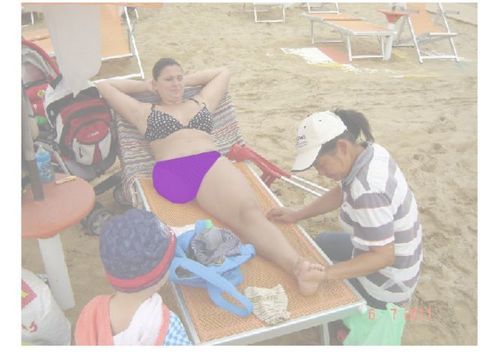} &
  \includegraphics[width=0.14\textwidth]{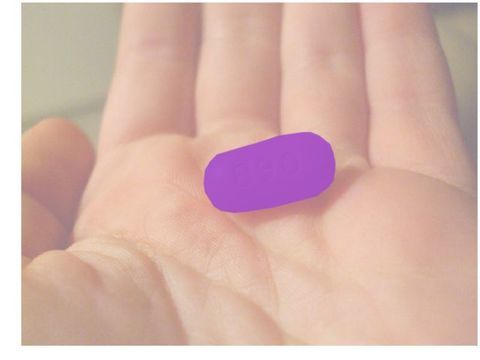} &
  \includegraphics[width=0.14\textwidth]{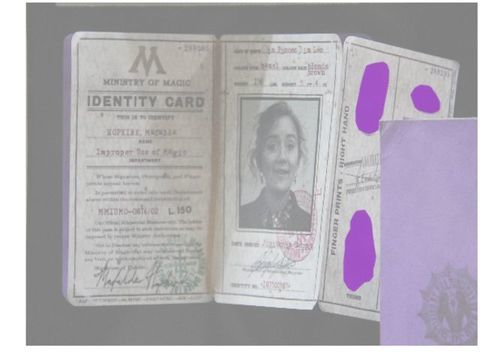} &
  \includegraphics[width=0.14\textwidth]{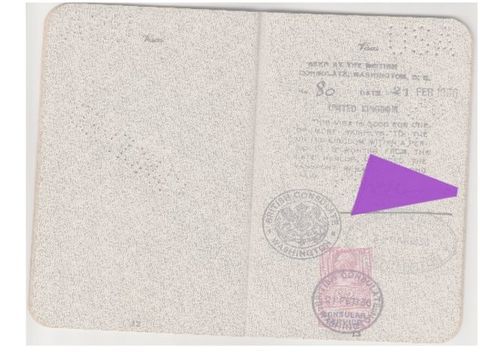} &
  \includegraphics[width=0.14\textwidth]{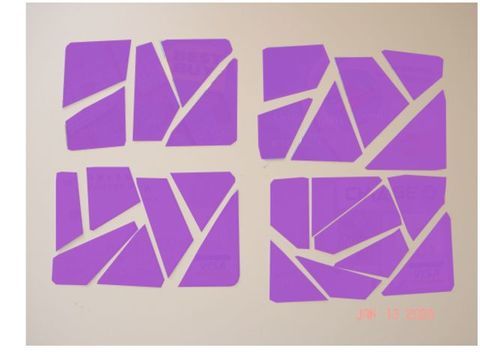} \\
  \rotatebox[origin=l]{90}{\: Predicted} &
  \includegraphics[width=0.14\textwidth]{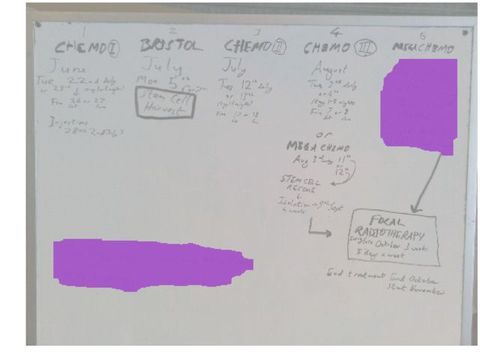} &
  \includegraphics[width=0.14\textwidth]{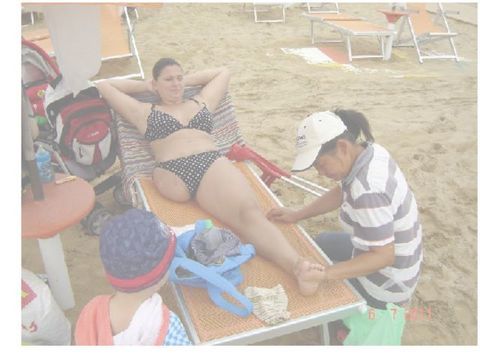} &
  \includegraphics[width=0.14\textwidth]{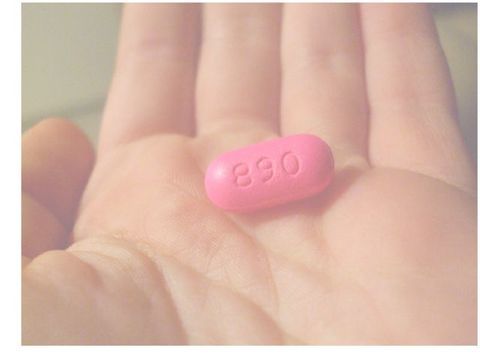} &
  \includegraphics[width=0.14\textwidth]{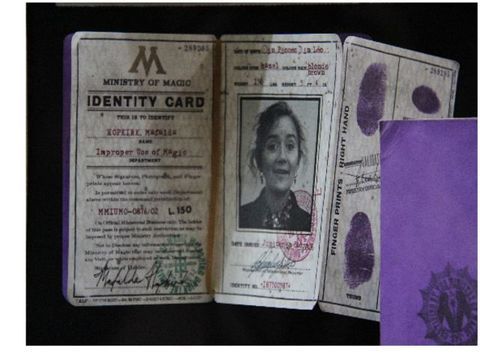} &
  \includegraphics[width=0.14\textwidth]{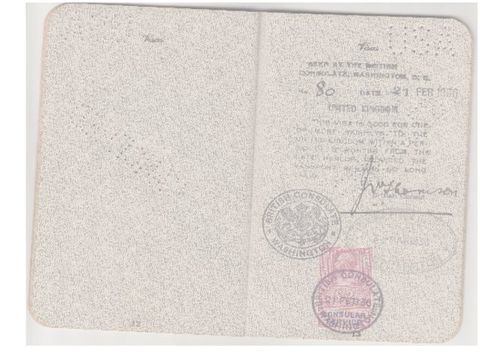} &
  \includegraphics[width=0.14\textwidth]{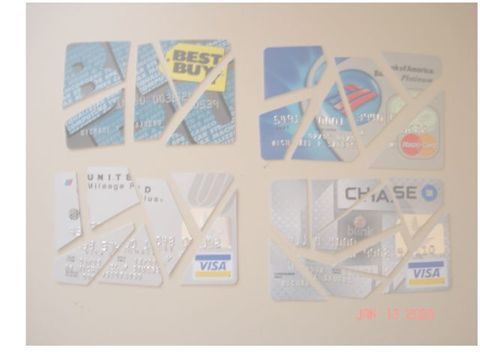}
  \end{tabular}
  \caption{Qualitative results per attribute. In each pair of images, top is ground-truth segmentation and bottom is prediction. Pairs of images in each column are sorted by IoU scores (high to low).}
  \label{fig:qual1}
\end{figure*}

\begin{figure*}[tbp]
  \centering
  \begin{tabular}{p{0.1cm}cccccc}
  & \attr{passport} & \attr{driv\_lic} & \attr{stud\_id} & \attr{mail} & \attr{receipt} & \attr{ticket} \\
   &        &            &      &           &           &          \\
   &        &            &  \multicolumn{2}{c}{good (iou $\ge$ 0.75)}           &           &          \\
   &        &            &      &           &           &          \\
  \rotatebox[origin=l]{90}{\quad \: GT} &
  \includegraphics[width=0.14\textwidth]{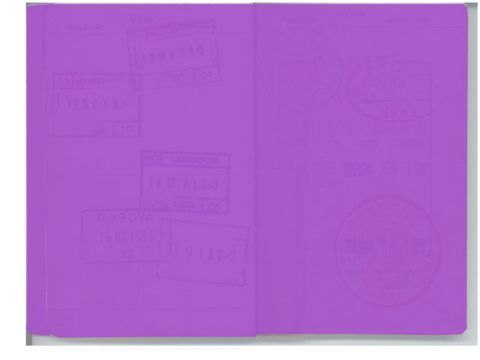} &
  \includegraphics[width=0.14\textwidth]{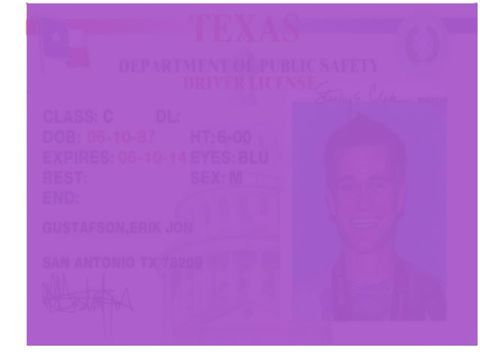} &
  \includegraphics[width=0.14\textwidth]{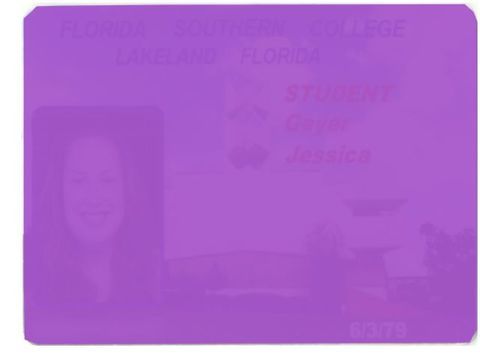} &
  \includegraphics[width=0.14\textwidth]{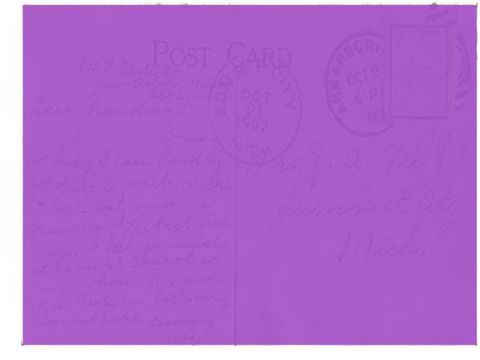} &
  \includegraphics[width=0.14\textwidth]{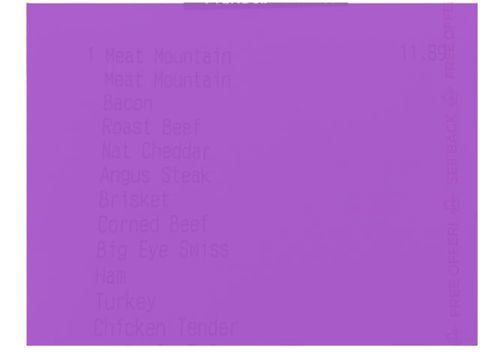} &
  \includegraphics[width=0.14\textwidth]{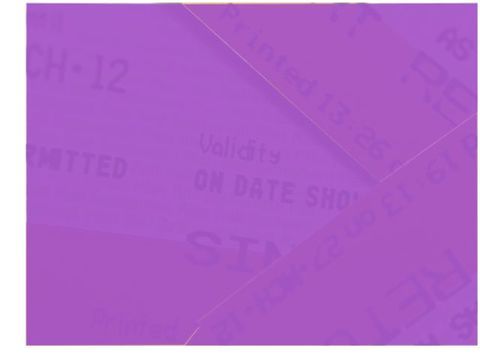} \\
  \rotatebox[origin=l]{90}{\: Predicted} &
  \includegraphics[width=0.14\textwidth]{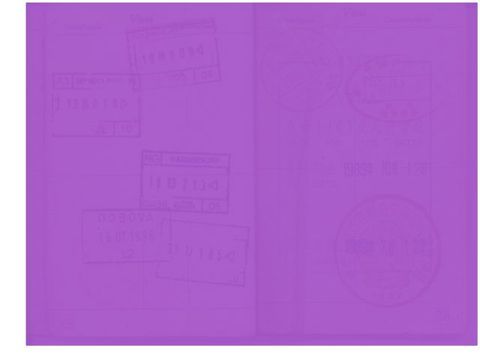} &
  \includegraphics[width=0.14\textwidth]{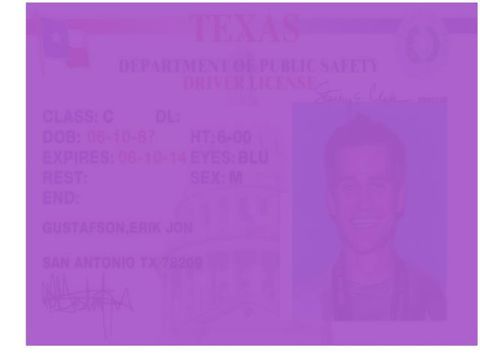} &
  \includegraphics[width=0.14\textwidth]{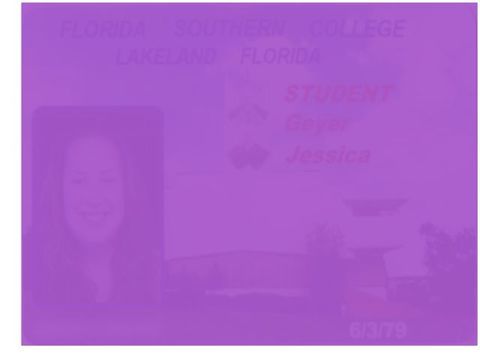} &
  \includegraphics[width=0.14\textwidth]{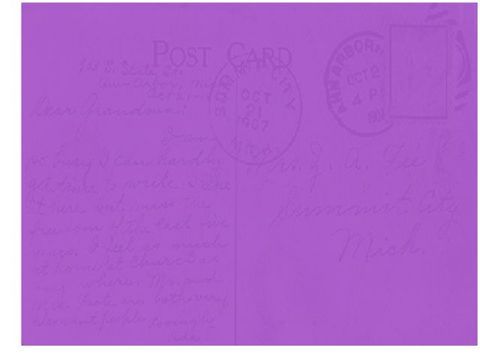} &
  \includegraphics[width=0.14\textwidth]{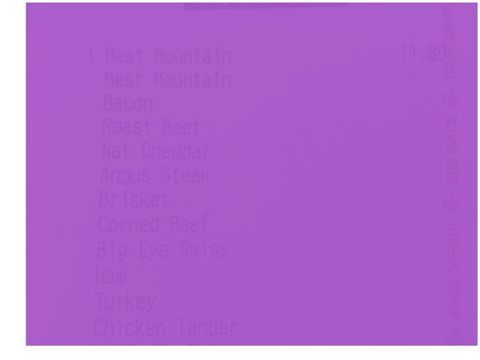} &
  \includegraphics[width=0.14\textwidth]{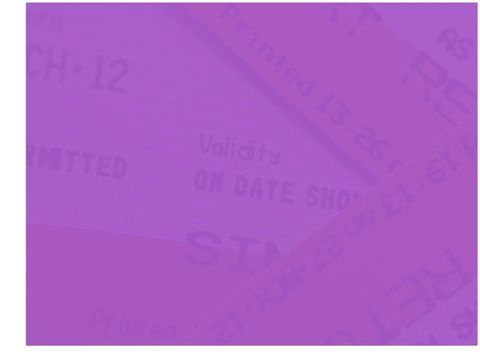} \\
   &        &            &      &           &           &          \\
   &        &            &  \multicolumn{2}{c}{mediocre (0.25 $\le$ iou $<$ 0.75)}           &           &          \\
   &        &            &      &           &           &          \\
           
  \rotatebox[origin=l]{90}{\quad \: GT} &
  \includegraphics[width=0.14\textwidth]{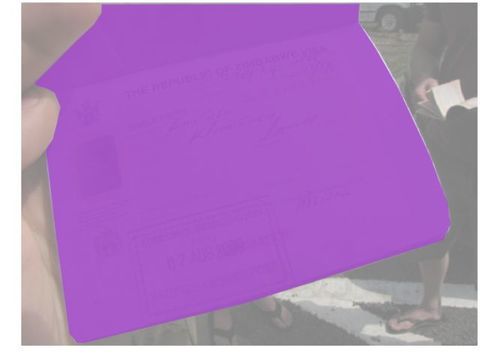} &
  \includegraphics[width=0.14\textwidth]{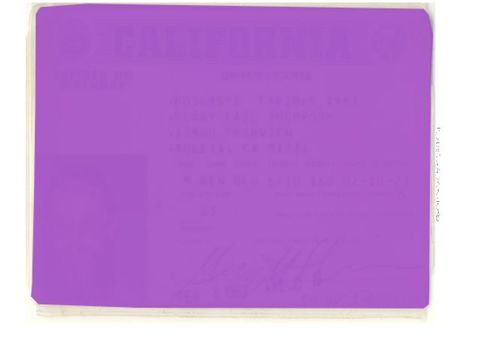} &
  \includegraphics[width=0.14\textwidth]{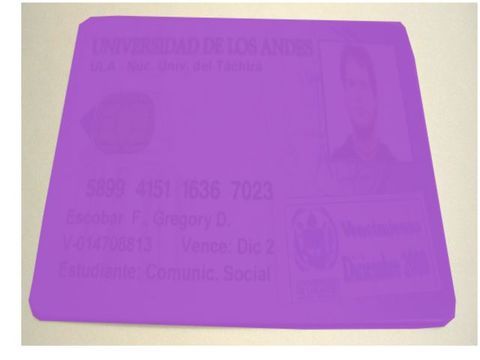} &
  \includegraphics[width=0.14\textwidth]{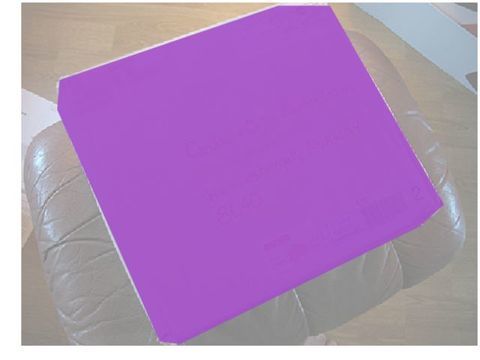} &
  \includegraphics[width=0.14\textwidth]{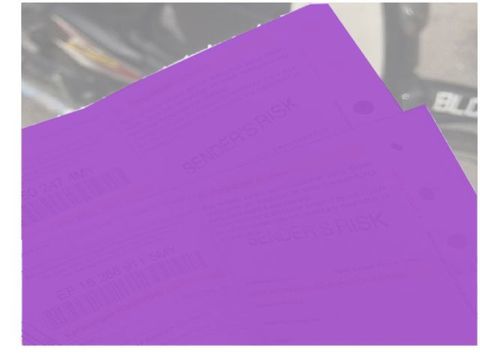} &
  \includegraphics[width=0.14\textwidth]{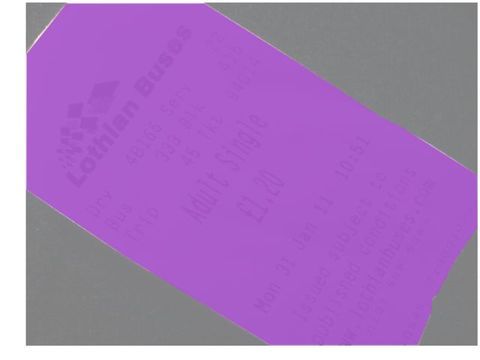} \\
  \rotatebox[origin=l]{90}{\: Predicted} &
  \includegraphics[width=0.14\textwidth]{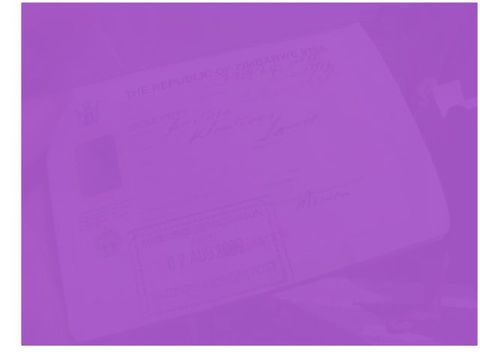} &
  \includegraphics[width=0.14\textwidth]{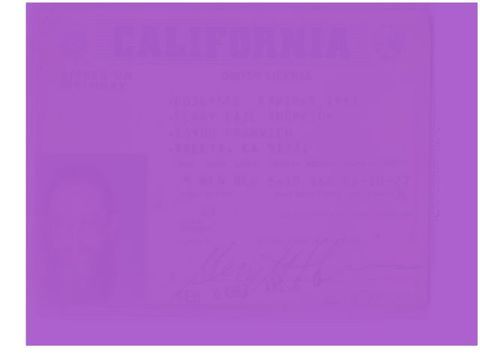} &
  \includegraphics[width=0.14\textwidth]{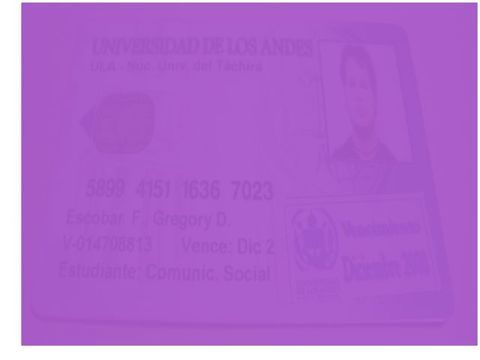} &
  \includegraphics[width=0.14\textwidth]{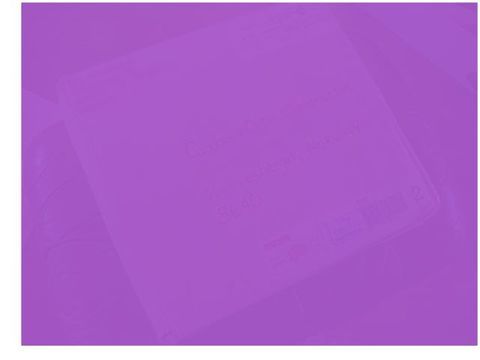} &
  \includegraphics[width=0.14\textwidth]{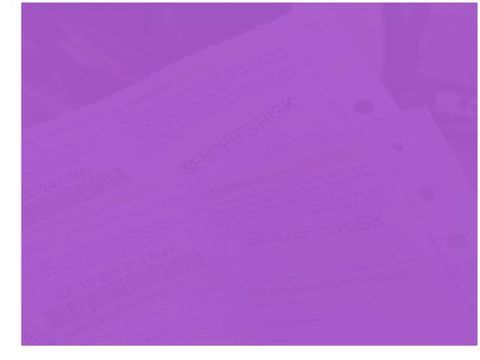} &
  \includegraphics[width=0.14\textwidth]{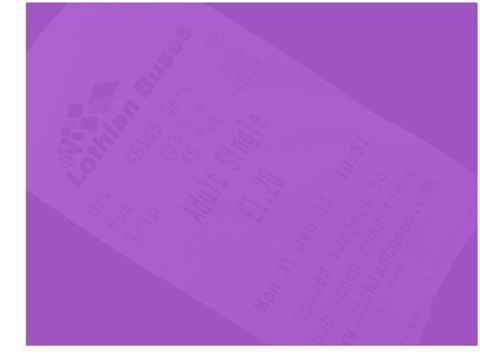} \\
   &        &            &      &           &           &          \\
           
  \rotatebox[origin=l]{90}{\quad \: GT} &
  \includegraphics[width=0.14\textwidth]{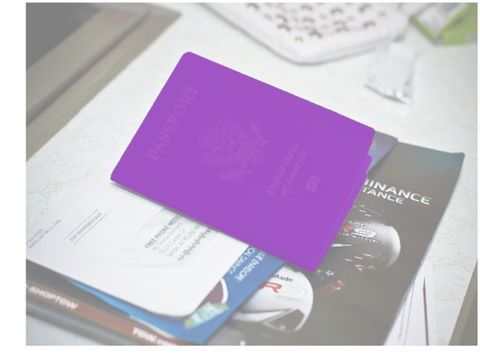} &
  \includegraphics[width=0.14\textwidth]{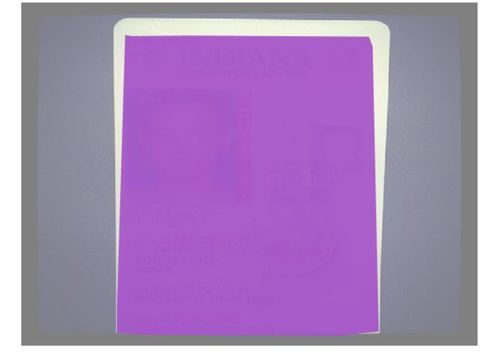} &
  \includegraphics[width=0.14\textwidth]{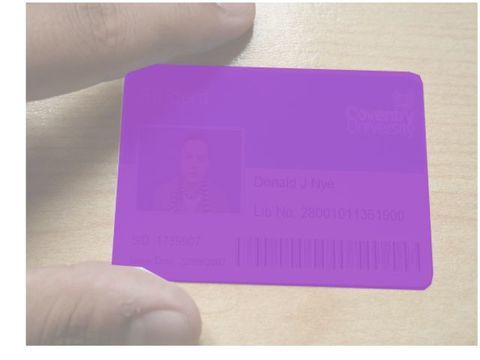} &
  \includegraphics[width=0.14\textwidth]{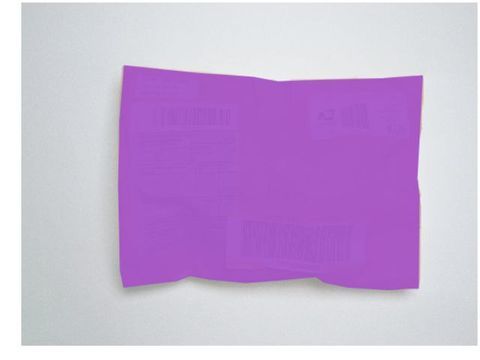} &
  \includegraphics[width=0.14\textwidth]{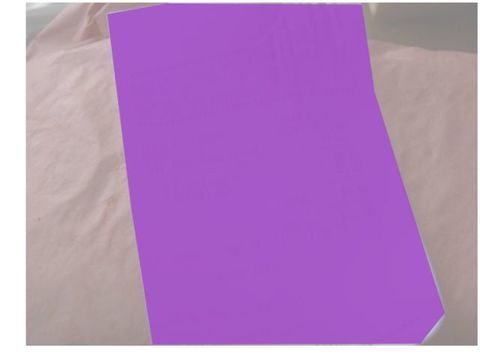} &
  \includegraphics[width=0.14\textwidth]{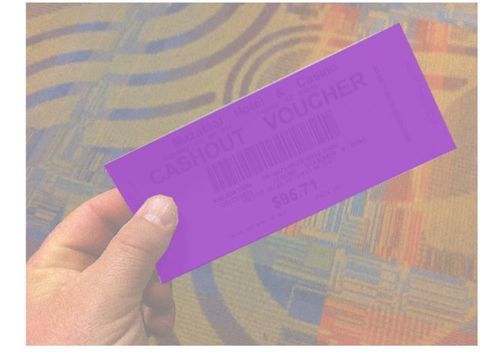} \\
  \rotatebox[origin=l]{90}{\: Predicted} &
  \includegraphics[width=0.14\textwidth]{fig/qual/resized/passport/2017_37610154_pred_segmentation.jpg} &
  \includegraphics[width=0.14\textwidth]{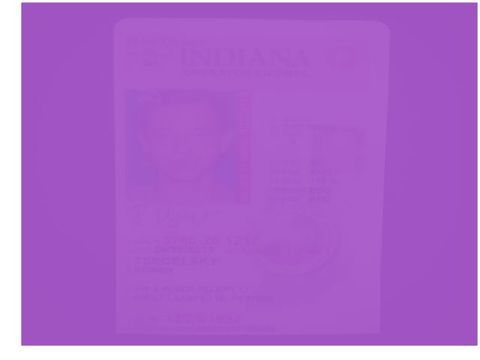} &
  \includegraphics[width=0.14\textwidth]{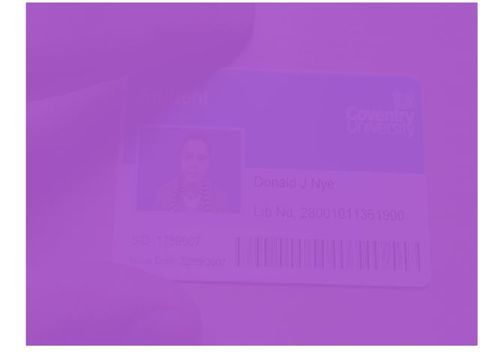} &
  \includegraphics[width=0.14\textwidth]{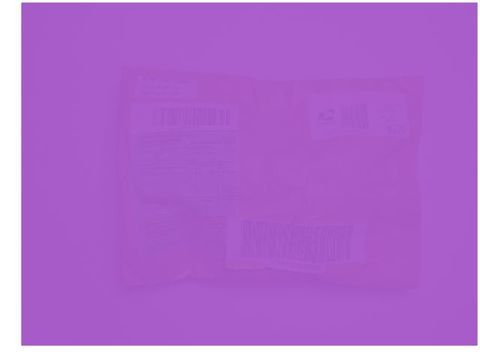} &
  \includegraphics[width=0.14\textwidth]{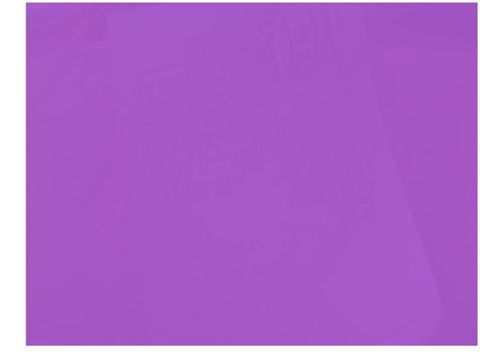} &
  \includegraphics[width=0.14\textwidth]{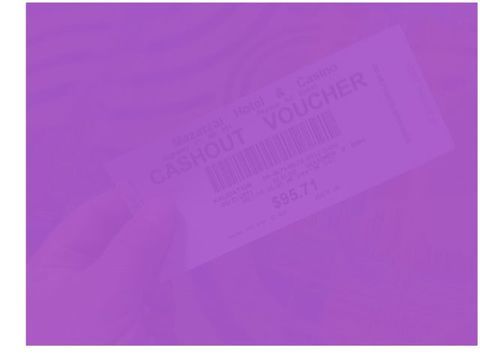} \\
  &         &            &      &           &           &          \\
  &         &            &  \multicolumn{2}{c}{failure (iou $\approx$ 0)}           &           &          \\
  &         &            &      &           &           &          \\
           
  \rotatebox[origin=l]{90}{\quad \: GT} &
  \includegraphics[width=0.14\textwidth]{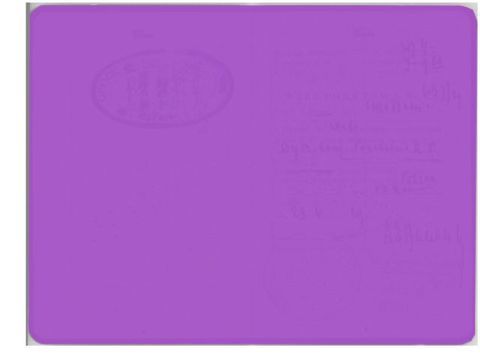} &
  \includegraphics[width=0.14\textwidth]{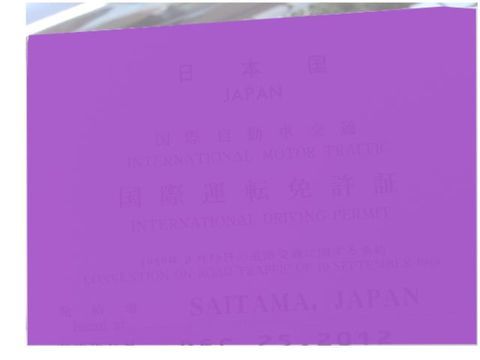} &
  \includegraphics[width=0.14\textwidth]{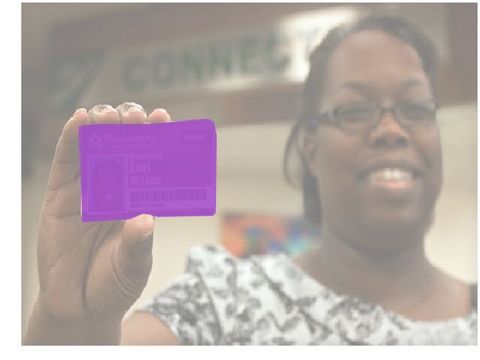} &
  \includegraphics[width=0.14\textwidth]{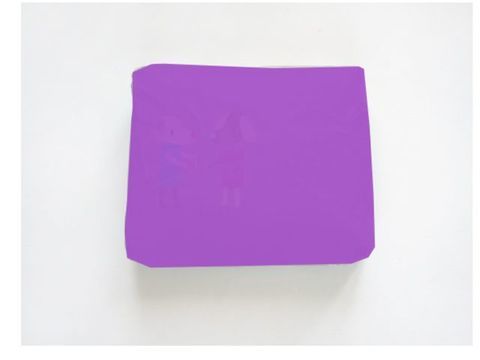} &
  \includegraphics[width=0.14\textwidth]{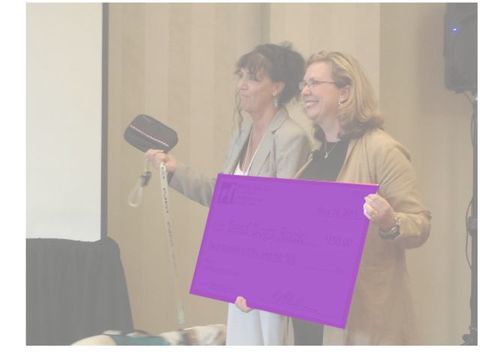} &
  \includegraphics[width=0.14\textwidth]{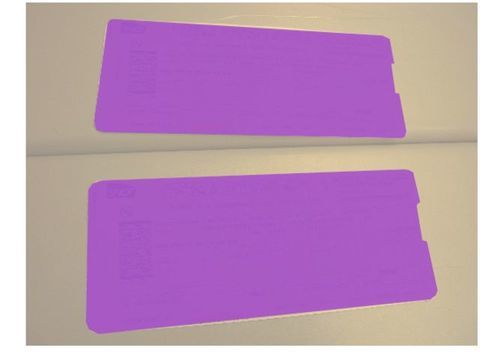} \\
  \rotatebox[origin=l]{90}{\: Predicted} &
  \includegraphics[width=0.14\textwidth]{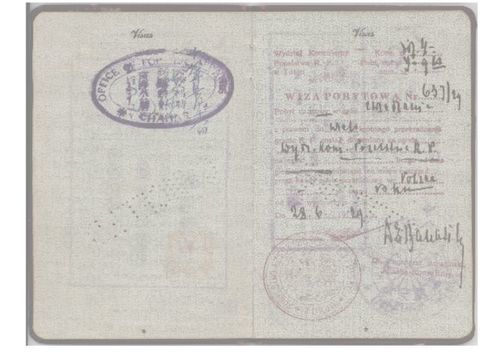} &
  \includegraphics[width=0.14\textwidth]{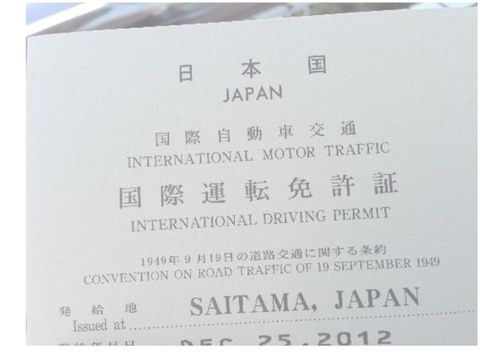} &
  \includegraphics[width=0.14\textwidth]{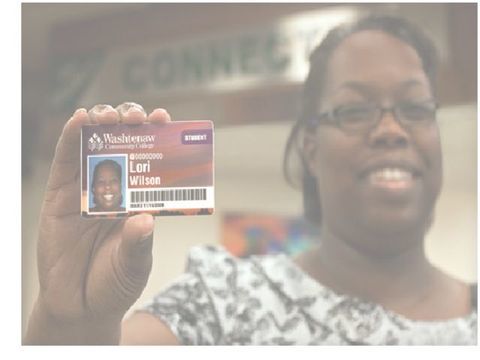} &
  \includegraphics[width=0.14\textwidth]{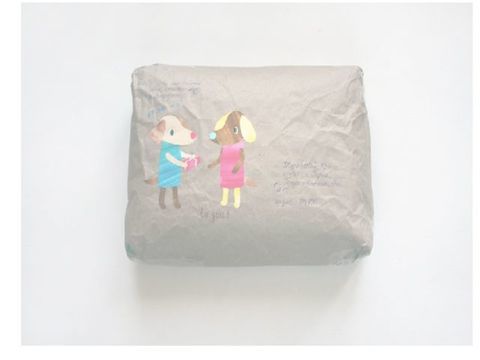} &
  \includegraphics[width=0.14\textwidth]{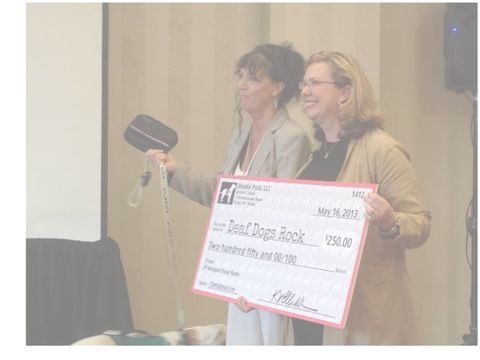} &
  \includegraphics[width=0.14\textwidth]{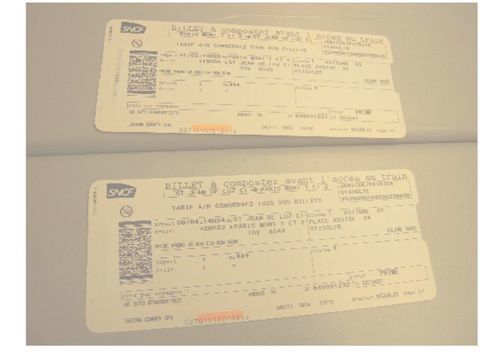}
  \end{tabular}
  \caption{Qualitative results per attribute. In each pair of images, top is ground-truth segmentation and bottom is prediction. Pairs of images in each column are sorted by IoU scores (high to low).}
  \label{fig:qual1}
\end{figure*}

\begin{figure*}[]
  \centering
  \begin{tabular}{p{0.1cm}ccccc}
 						&              &              & \textit{Success Modes}             &              &             \\
                        &              &              &              &              &             \\
                        & 0.25         & 0.5          & 1.0          & 2.0          & 4.0          \\
  						&              &              &              &              &             \\
                        &              &              &   \textsc{Textual} (\attr{location})     &              &             \\
                        &              &              &              &              &             \\
				       & $P$ = \red{0}, $U$ = \green{100} 
					   & $P$ = \red{0}, $U$ = \green{100} 
                       & $P$ = \green{80}, $U$ = \green{100} 
                       & $P$ = \green{100}, $U$ = \green{100} 
                       & $P$ = \green{100}, $U$ = \green{80} \\
 \rotatebox[origin=l]{90}{\: GT-based} & \includegraphics[width=0.15\textwidth]{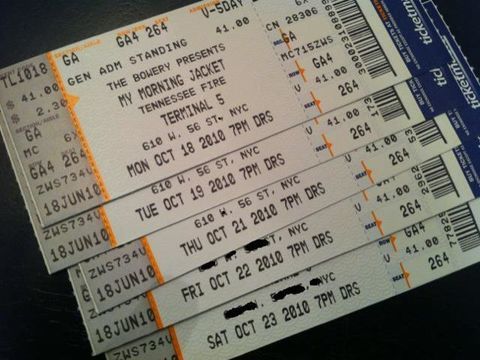} 
                        & \includegraphics[width=0.15\textwidth]{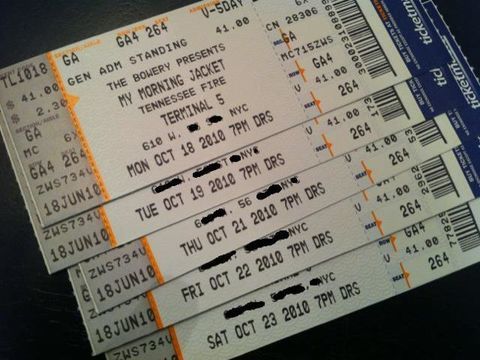} 
                        & \includegraphics[width=0.15\textwidth]{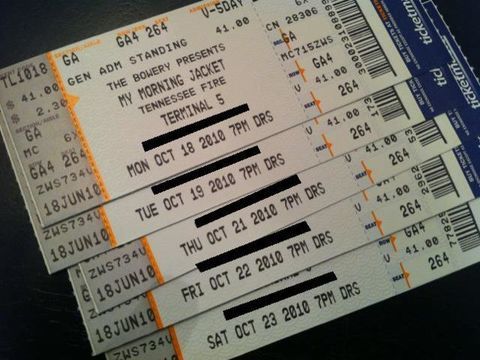} 
                        & \includegraphics[width=0.15\textwidth]{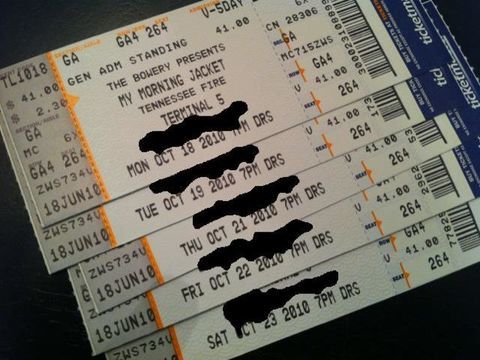} 
                        & \includegraphics[width=0.15\textwidth]{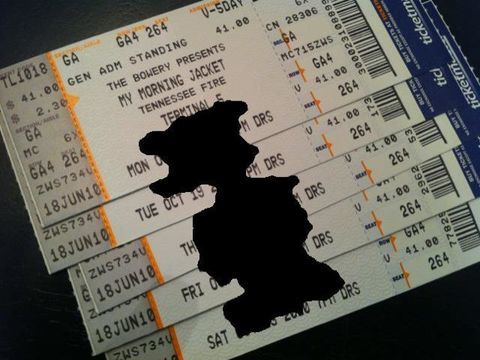} \\
                      & $P$ = \red{20}, $U$ = \green{100} 
					  & $P$ = \red{20}, $U$ = \green{100} 
                      & $P$ = \red{0}, $U$ = \green{100} 
                      & $P$ = \red{20}, $U$ = \green{100} 
                      & $P$ = \green{80}, $U$ = \green{100} \\
 \rotatebox[origin=l]{90}{\: Predicted}                       & \includegraphics[width=0.15\textwidth]{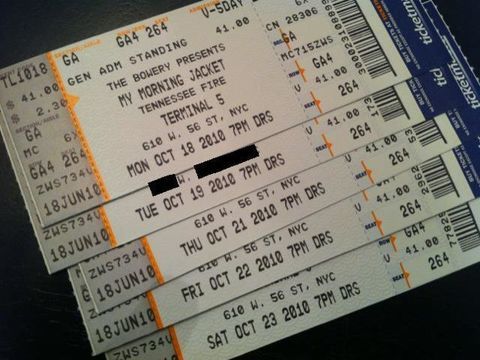} 
                        & \includegraphics[width=0.15\textwidth]{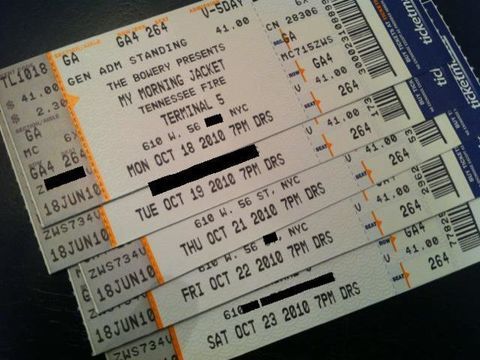} 
                        & \includegraphics[width=0.15\textwidth]{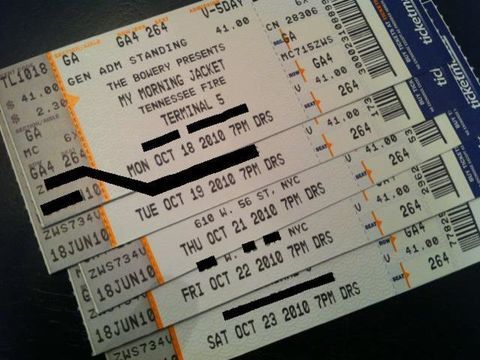} 
                        & \includegraphics[width=0.15\textwidth]{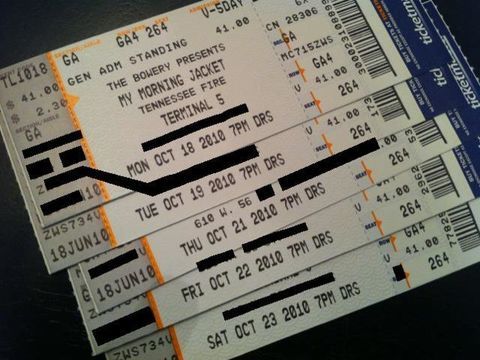} 
                        & \includegraphics[width=0.15\textwidth]{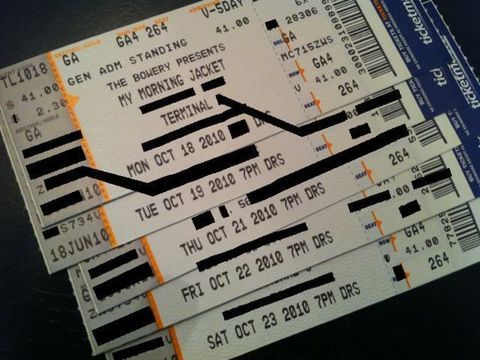} \\
                        &              &              &              &              &             \\
                        &              &              &   \textsc{Visual} (\attr{face})      &              &             \\
                        &              &              &              &              &             \\
                      & $P$ = \green{60}, $U$ = \green{80} 
					  & $P$ = \green{80}, $U$ = \green{80} 
                      & $P$ = \green{100}, $U$ = \green{100} 
                      & $P$ = \green{100}, $U$ = \green{80} 
                      & $P$ = \green{100}, $U$ = \green{80} \\
 \rotatebox[origin=l]{90}{\: GT-based}                       & \includegraphics[width=0.15\textwidth]{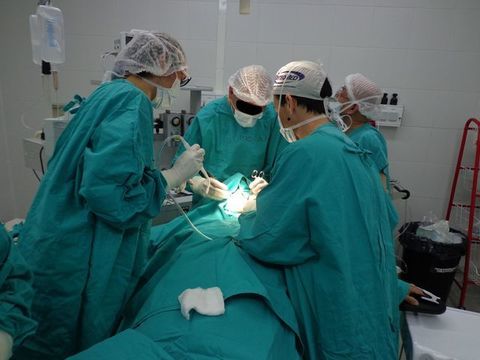} 
                        & \includegraphics[width=0.15\textwidth]{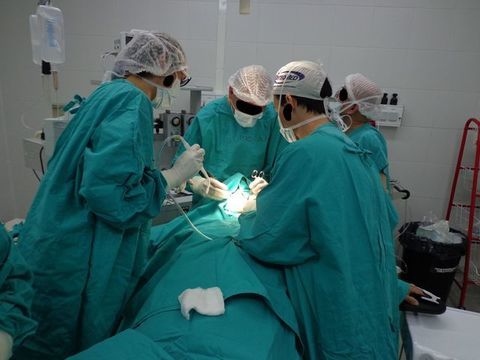} 
                        & \includegraphics[width=0.15\textwidth]{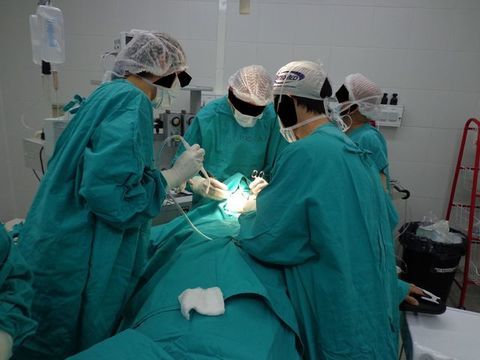} 
                        & \includegraphics[width=0.15\textwidth]{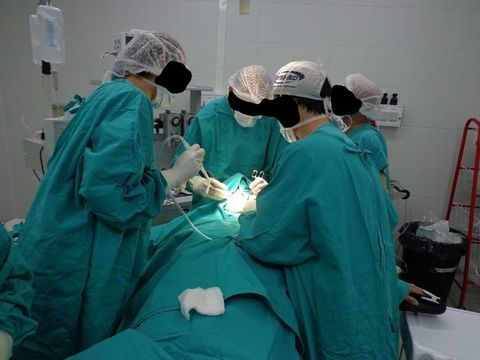} 
                        & \includegraphics[width=0.15\textwidth]{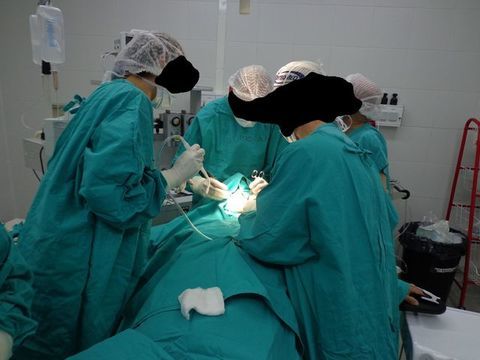} \\
   & $P$ = \red{0}, $U$ = \green{100} 
					   & $P$ = \red{0}, $U$ = \green{100} 
                       & $P$ = \green{60}, $U$ = \green{100} 
                       & $P$ = \green{100}, $U$ = \green{100} 
                       & $P$ = \green{100}, $U$ = \green{100} \\
 \rotatebox[origin=l]{90}{\: Predicted}                       & \includegraphics[width=0.15\textwidth]{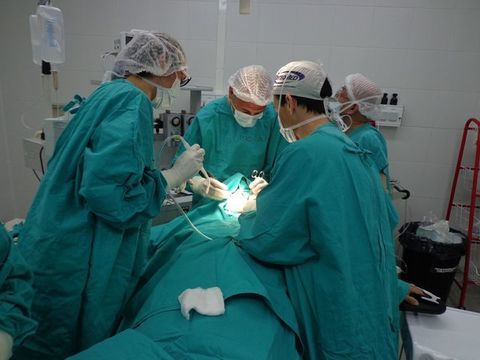} 
                        & \includegraphics[width=0.15\textwidth]{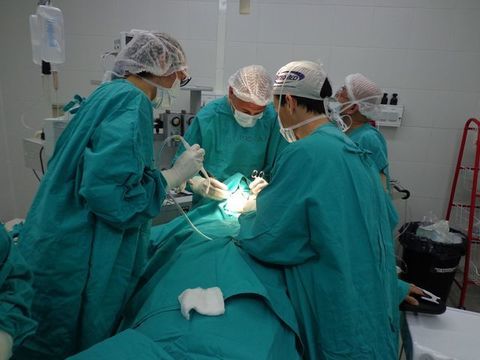} 
                        & \includegraphics[width=0.15\textwidth]{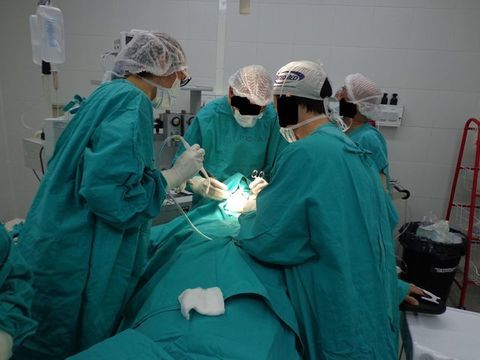} 
                        & \includegraphics[width=0.15\textwidth]{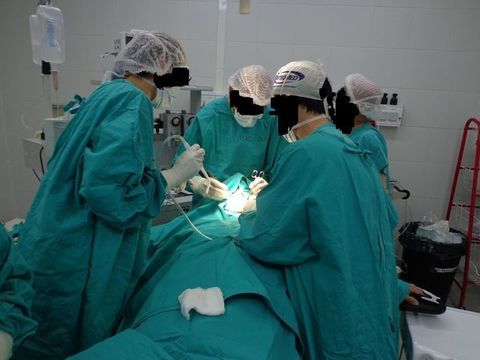} 
                        & \includegraphics[width=0.15\textwidth]{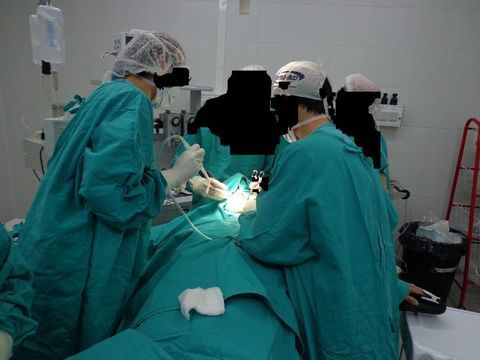} \\
                        &              &              &              &              &             \\
                        &              &              &   \textsc{Multimodal} (\attr{mail})      &              &             \\
                        &              &              &              &              &             \\
   & $P$ = \red{0}, $U$ = \green{100} 
					  & $P$ = \red{0}, $U$ = \green{80} 
                      & $P$ = \green{100}, $U$ = \red{20} 
                      & $P$ = \green{100}, $U$ = \red{0} 
                      & $P$ = \green{100}, $U$ = \red{0} \\
\rotatebox[origin=l]{90}{\: GT-based}                        & \includegraphics[width=0.15\textwidth]{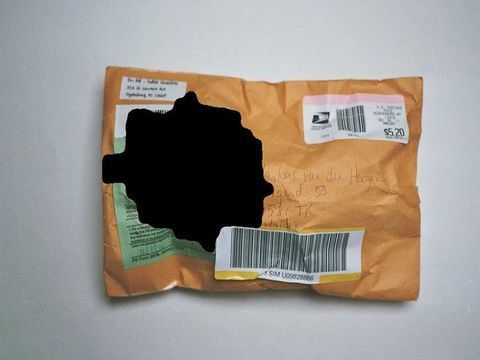} 
                        & \includegraphics[width=0.15\textwidth]{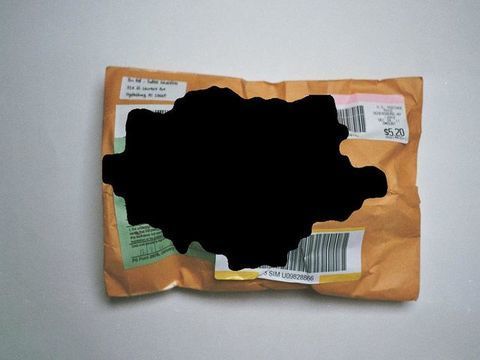} 
                        & \includegraphics[width=0.15\textwidth]{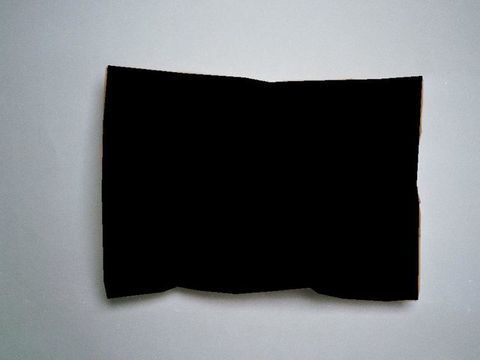} 
                        & \includegraphics[width=0.15\textwidth]{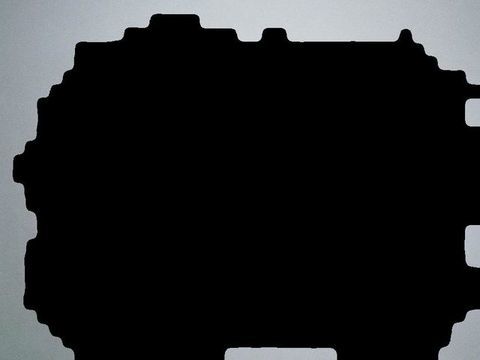} 
                        & \includegraphics[width=0.15\textwidth]{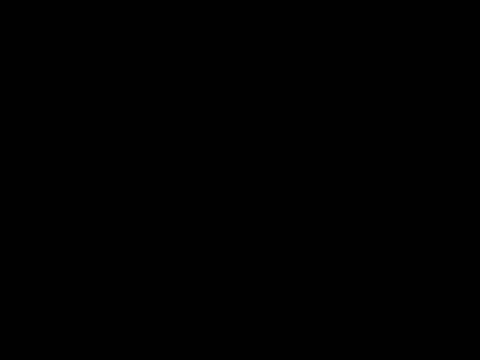} \\
   & $P$ = \red{0}, $U$ = \green{100} 
					   & $P$ = \red{0}, $U$ = \green{100} 
                       & $P$ = \green{100}, $U$ = \red{0} 
                       & $P$ = \green{100}, $U$ = \red{0} 
                       & $P$ = \green{100}, $U$ = \red{0} \\
\rotatebox[origin=l]{90}{\: Predicted}                        & \includegraphics[width=0.15\textwidth]{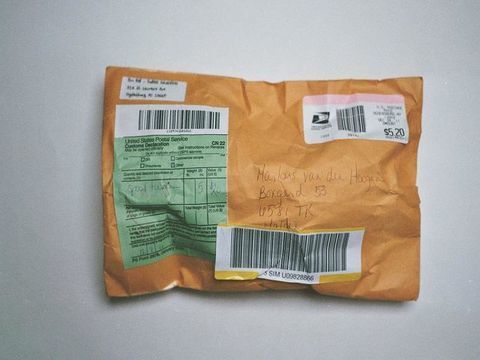} 
                        & \includegraphics[width=0.15\textwidth]{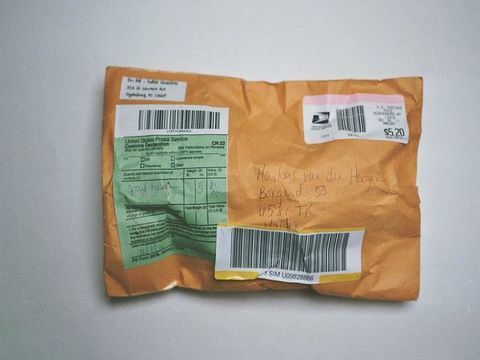} 
                        & \includegraphics[width=0.15\textwidth]{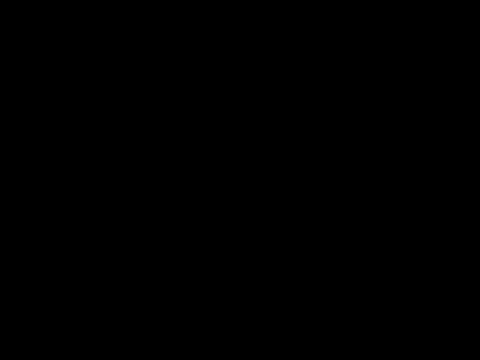} 
                        & \includegraphics[width=0.15\textwidth]{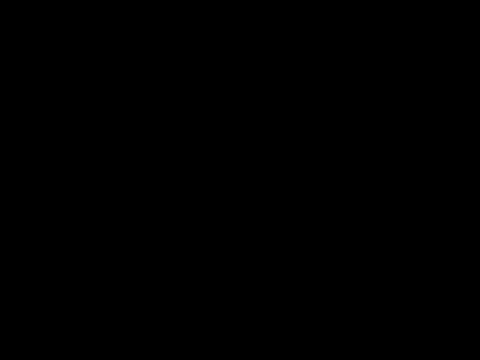} 
                        & \includegraphics[width=0.15\textwidth]{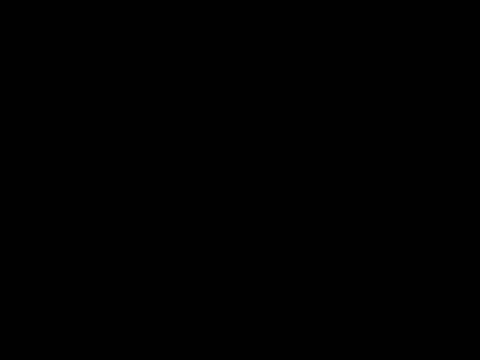} \\
  \end{tabular}
  \caption{\textit{Common Success Modes of Automatic Redactions}. GT-based are ground-truth regions scaled and redacted as discussed previously. Predicted are automatic redactions generated by method ENSEMBLE. $P$ indicates privacy score and $U$ indicates utility score. In both cases, higher is better. Scores are indicated in \green{green} in case of majority agreement and \red{red} otherwise.}
  \label{fig:redactions_success}
\end{figure*}

\begin{figure*}[]
  \centering
  \begin{tabular}{lccccc}
 						&              &              & \textit{Failure Modes}             &              &             \\
                        &              &              &              &              &             \\
                        & 0.25         & 0.5          & 1.0          & 2.0          & 4.0          \\
  						&              &              &              &              &             \\
                        &              &              &   \textsc{Textual} (\attr{home\_addr})      &              &             \\
                        &              &              &              &              &             \\
   & $P$ = \red{0}, $U$ = \green{100}
					  & $P$ = \green{100}, $U$ = \green{80} 
                      & $P$ = \green{100}, $U$ = \red{20} 
                      & $P$ = \green{100}, $U$ = \green{60} 
                      & $P$ = \green{100}, $U$ = \red{0} \\
\rotatebox[origin=l]{90}{\: GT-based}                        & \includegraphics[width=0.15\textwidth]{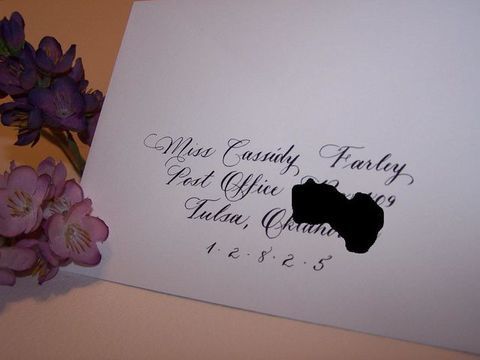} 
                        & \includegraphics[width=0.15\textwidth]{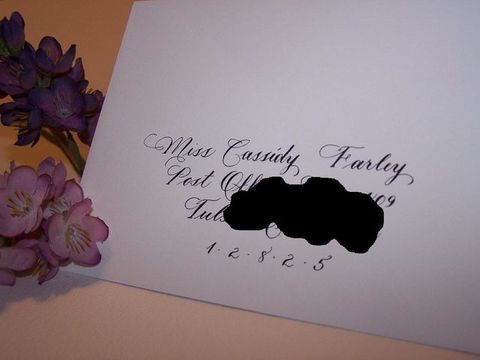} 
                        & \includegraphics[width=0.15\textwidth]{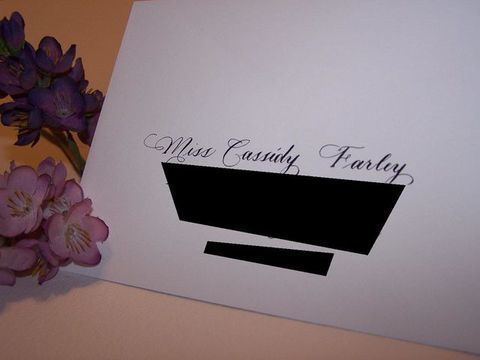} 
                        & \includegraphics[width=0.15\textwidth]{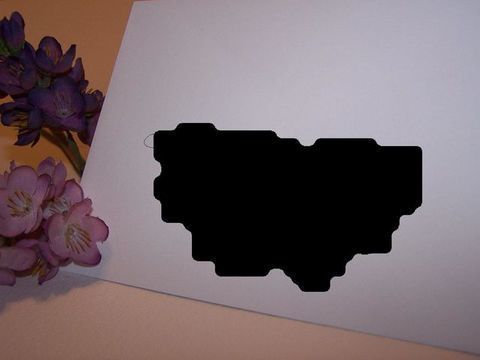} 
                        & \includegraphics[width=0.15\textwidth]{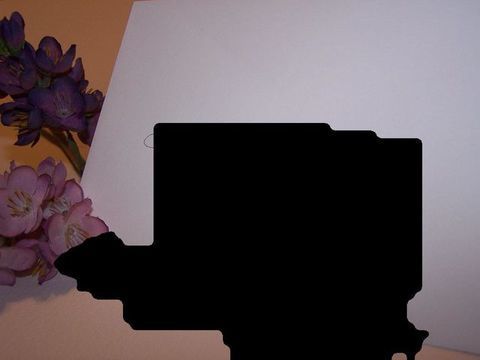} \\
   & $P$ = \red{0}, $U$ = \green{100} 
					  & $P$ = \red{0}, $U$ = \green{100} 
                      & $P$ = \red{0}, $U$ = \green{100} 
                      & $P$ = \red{0}, $U$ = \green{100} 
                      & $P$ = \red{0}, $U$ = \green{100} \\
\rotatebox[origin=l]{90}{\: Predicted}                        & \includegraphics[width=0.15\textwidth]{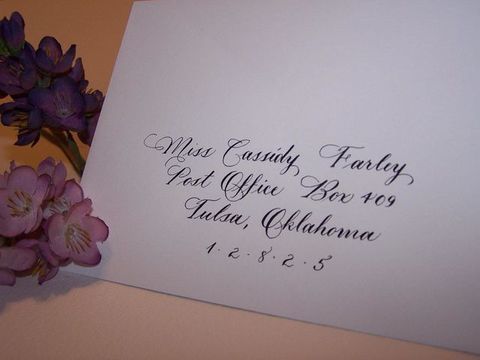} 
                        & \includegraphics[width=0.15\textwidth]{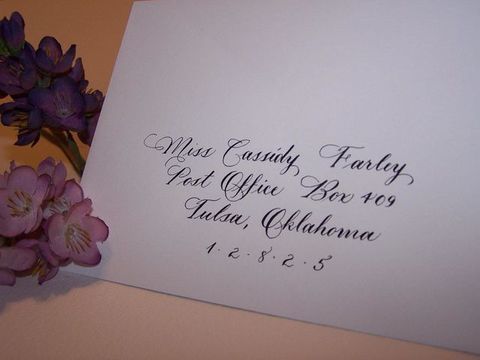} 
                        & \includegraphics[width=0.15\textwidth]{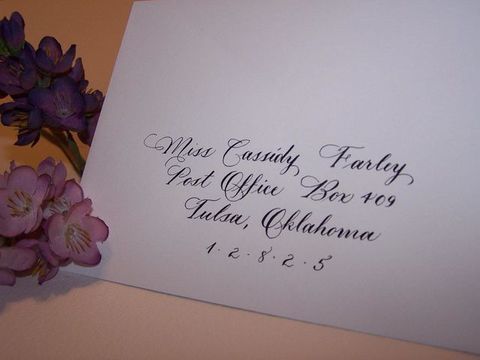} 
                        & \includegraphics[width=0.15\textwidth]{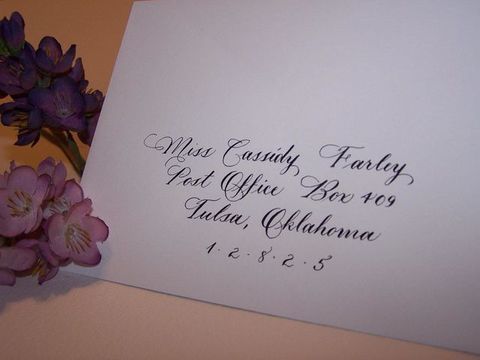} 
                        & \includegraphics[width=0.15\textwidth]{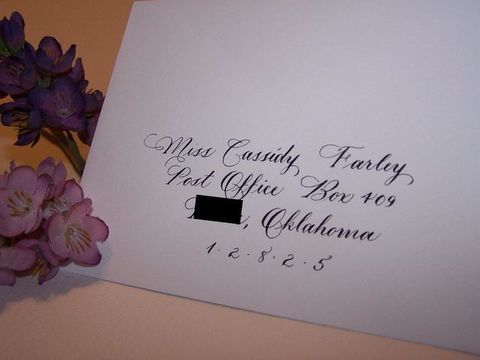} \\
                        &              &              &              &              &             \\
                        &              &              &  \textsc{Visual} (\attr{fingerpr})      &              &             \\
                        &              &              &              &              &             \\
   & $P$ = \red{10}, $U$ = \green{90}
					  & $P$ = \red{0}, $U$ = \green{100} 
                      & $P$ = \red{30}, $U$ = \green{100} 
                      & $P$ = \green{100}, $U$ = \red{40} 
                      & $P$ = \green{100}, $U$ = \red{0} \\
\rotatebox[origin=l]{90}{\: GT-based}                        & \includegraphics[width=0.15\textwidth]{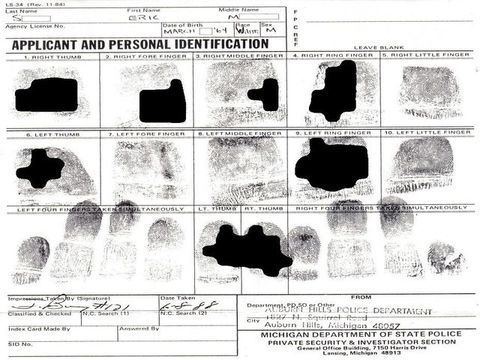} 
                        & \includegraphics[width=0.15\textwidth]{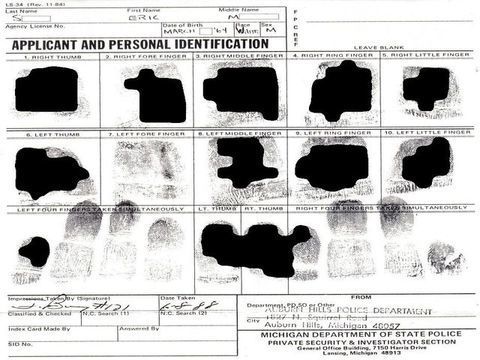} 
                        & \includegraphics[width=0.15\textwidth]{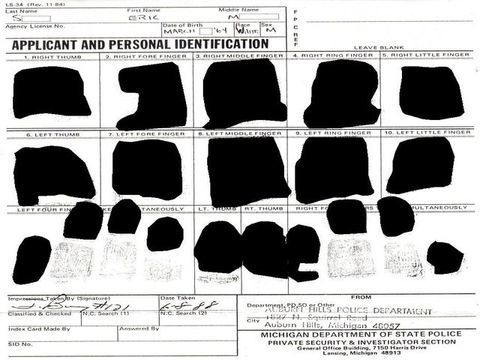} 
                        & \includegraphics[width=0.15\textwidth]{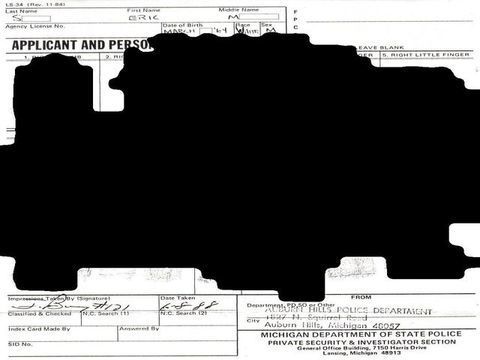} 
                        & \includegraphics[width=0.15\textwidth]{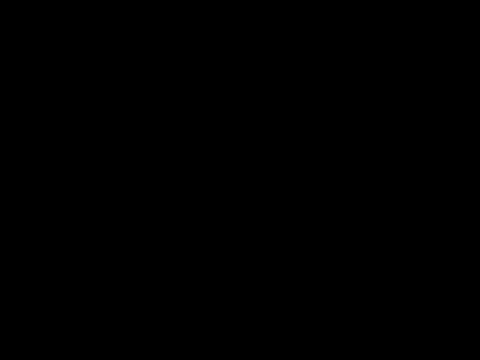} \\
   & $P$ = \red{0}, $U$ = \green{100} 
					  & $P$ = \red{0}, $U$ = \green{100} 
                      & $P$ = \red{0}, $U$ = \green{100} 
                      & $P$ = \red{0}, $U$ = \green{100} 
                      & $P$ = \red{0}, $U$ = \green{100} \\
\rotatebox[origin=l]{90}{\: Predicted}                        & \includegraphics[width=0.15\textwidth]{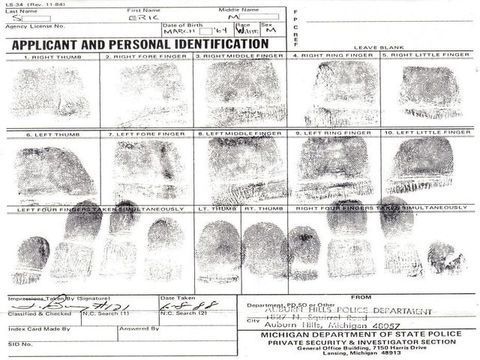} 
                        & \includegraphics[width=0.15\textwidth]{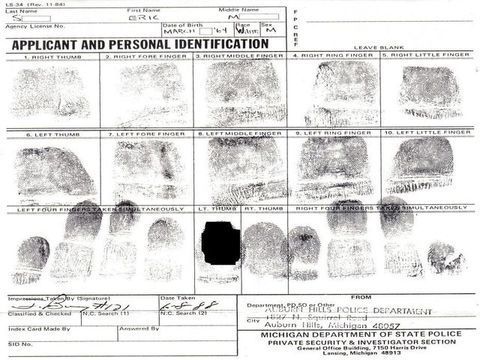} 
                        & \includegraphics[width=0.15\textwidth]{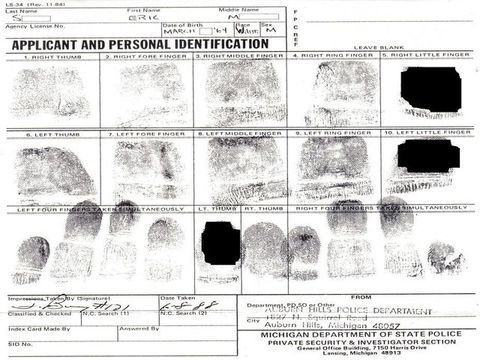} 
                        & \includegraphics[width=0.15\textwidth]{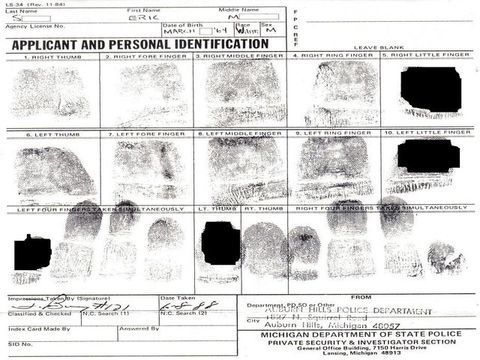} 
                        & \includegraphics[width=0.15\textwidth]{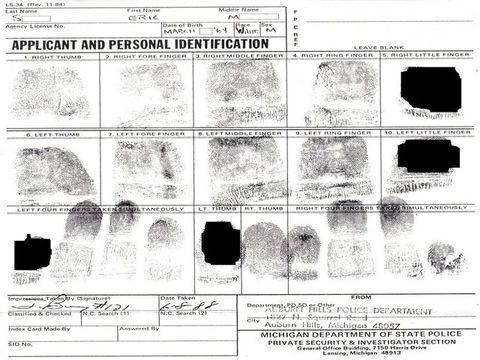} \\
                        &              &              &              &              &             \\
                        &              &              &  \textsc{Multimodal} (\attr{stud\_id})      &              &             \\
                        &              &              &              &              &             \\
   & $P$ = \red{0}, $U$ = \green{100} 
					  & $P$ = \red{0}, $U$ = \green{100} 
                      & $P$ = \green{100}, $U$ = \green{100}
                      & $P$ = \green{100}, $U$ = \green{60} 
                      & $P$ = \green{80}, $U$ = \red{40} \\
 \rotatebox[origin=l]{90}{\: GT-based}                       & \includegraphics[width=0.15\textwidth]{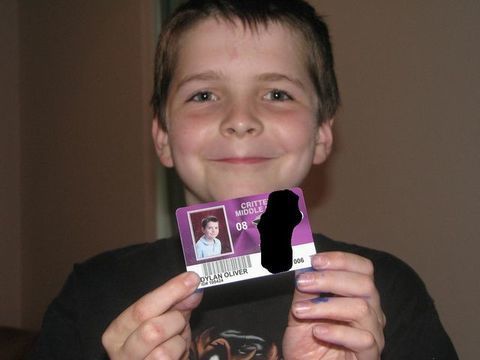} 
                        & \includegraphics[width=0.15\textwidth]{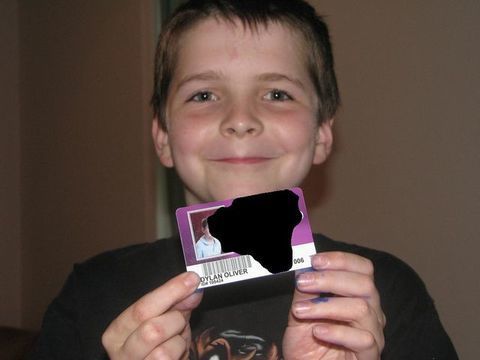} 
                        & \includegraphics[width=0.15\textwidth]{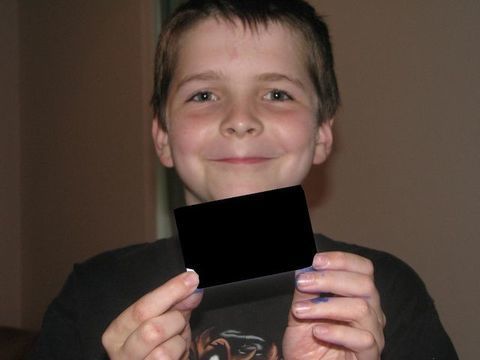} 
                        & \includegraphics[width=0.15\textwidth]{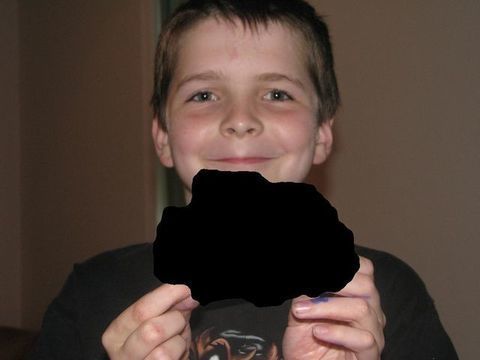} 
                        & \includegraphics[width=0.15\textwidth]{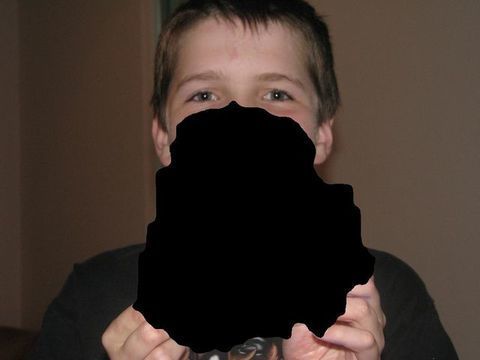} \\
   & $P$ = \red{0}, $U$ = \green{100} 
					  & $P$ = \red{0}, $U$ = \green{100} 
                      & $P$ = \red{0}, $U$ = \green{100} 
                      & $P$ = \red{0}, $U$ = \green{100} 
                      & $P$ = \green{100}, $U$ = \red{0} \\
\rotatebox[origin=l]{90}{\: Predicted}                        & \includegraphics[width=0.15\textwidth]{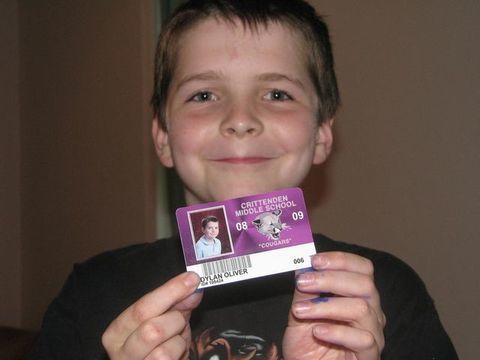} 
                        & \includegraphics[width=0.15\textwidth]{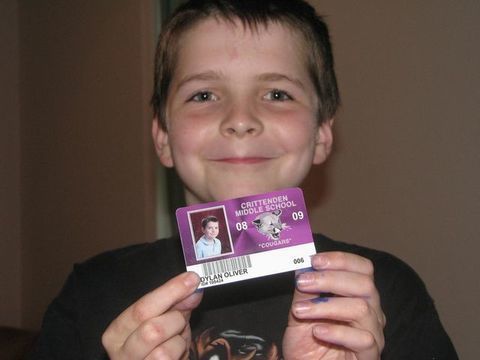} 
                        & \includegraphics[width=0.15\textwidth]{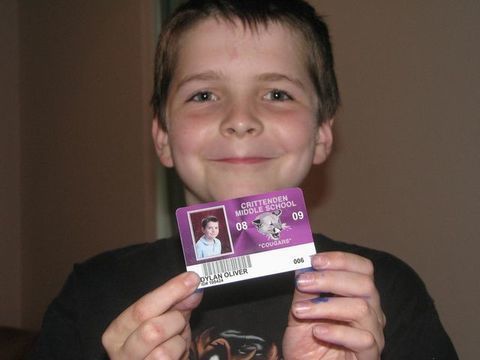} 
                        & \includegraphics[width=0.15\textwidth]{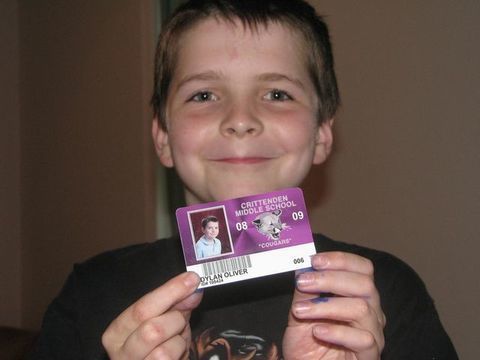} 
                        & \includegraphics[width=0.15\textwidth]{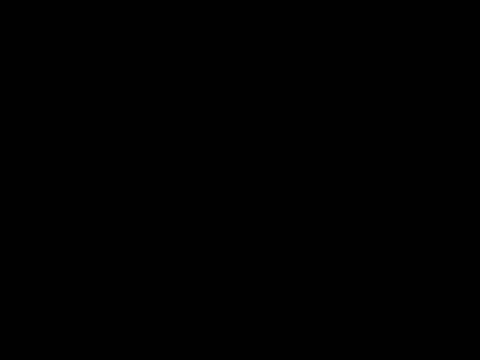} \\
  \end{tabular}
  \caption{\textit{Common Failure Modes of Automatic Redactions}. GT-based are ground-truth regions scaled and redacted as discussed previously. Predicted are automatic redactions generated by method ENSEMBLE. $P$ indicates privacy score and $U$ indicates utility score. In both cases, higher is better. Scores are indicated in \green{green} in case of majority agreement and \red{red} otherwise.%
  }
  \label{fig:redactions_failure}
\end{figure*}

\end{document}